\documentclass{article}
\PassOptionsToPackage{numbers, compress}{natbib}
\usepackage[final]{neurips_2020}

\usepackage[utf8]{inputenc} %
\usepackage[T1]{fontenc}    %
\usepackage{hyperref}       %
\hypersetup{
    colorlinks=true,
    linkcolor=blue,
    filecolor=magenta,      
    urlcolor=cyan,
}
\usepackage{url}            %
\usepackage{booktabs}       %
\usepackage{rotating}
\usepackage{amsfonts}       %
\usepackage{nicefrac}       %
\usepackage{microtype}      %
\usepackage{graphicx}
\usepackage{amsmath,amssymb} %
\usepackage{cleveref}
\usepackage{array}
\usepackage{multirow}
\usepackage{bbding} %
\usepackage{paralist}
\usepackage{xspace}
\usepackage{subcaption}
\usepackage{xcolor} %
\usepackage{mathtools}
\usepackage{comment}

\definecolor{cadmiumgreen}{rgb}{0.0, 0.42, 0.24}
\definecolor{amethyst}{rgb}{0.6, 0.4, 0.8}
\definecolor{magentagoal}{RGB}{255, 0, 255}
\definecolor{cyangoal}{RGB}{60, 255, 255}

\DeclareMathOperator*{\concat}{concat}

\newcommand{\task}{multiON\xspace}
\newcommand{\mon}[1]{$#1$-ON}
\newcommand{\found}{\textsc{found}\xspace}

\newcolumntype{L}[1]{>{\raggedright\let\newline\\\arraybackslash\hspace{0pt}}m{#1}}
\newcolumntype{C}[1]{>{\centering\let\newline\\\arraybackslash\hspace{0pt}}m{#1}}
\newcolumntype{R}[1]{>{\raggedleft\let\newline\\\arraybackslash\hspace{0pt}}m{#1}}
\newcolumntype{H}[1]{>{\raggedright\let\newline\\\arraybackslash\hspace{0pt}}m{#1}}
\newcommand{\ra}[1]{\renewcommand{\arraystretch}{#1}}
\newcommand{\xhdr}[1]{\vspace{2pt}\noindent\textbf{#1}}
\newcommand{\greencheck}{{\textcolor{green}\Checkmark}}
\newcommand{\redx}{{\textcolor{red}\XSolidBrush}}
\DeclareMathSymbol{@}{\mathord}{letters}{"3B}

\newcommand{\Success}{\textsc{Success}\xspace}
\newcommand{\Progress}{\textsc{Progress}\xspace}
\newcommand{\SPL}{\textsc{SPL}\xspace}
\newcommand{\PPL}{\textsc{PPL}\xspace}
\newcommand{\Rand}{\texttt{Rand}\xspace}
\newcommand{\RandOracleFound}{\texttt{Rand+OracleFound}\xspace}

\newcommand{\NoMap}{\texttt{NoMap(RNN)}\xspace}
\newcommand{\NoMapAll}{the NoMap baselines (\texttt{RNN}, \texttt{SMT}, \texttt{FRMQN})\xspace}
\newcommand{\OracleMap}{\texttt{OracleMap}\xspace}
\newcommand{\OracleEgoMap}{\texttt{OracleEgoMap}\xspace}
\newcommand{\ProjNeuralMap}{\texttt{ProjNeuralMap}\xspace}
\newcommand{\ObjRecogMap}{\texttt{ObjRecogMap}\xspace}
\newcommand{\DyMap}{\texttt{DynamicOracleMap}\xspace}
\newcommand{\OnlyMap}{\texttt{OnlyOracleMap}\xspace}
\newcommand{\NoObject}{\texttt{OracleMapHiddenObjects}\xspace}
\newcommand{\FRMQN}{\texttt{FRMQN}\xspace}
\newcommand{\SMT}{\texttt{SMT}\xspace}
\newcommand{\EgoMap}{\texttt{EgoMap}\xspace}

\newcommand\filledcirc[1]{{\color{#1}\bullet}\mathllap{\color{#1}\circ}}
\newcommand\filledcircb[2]{{\color{#1}\bullet}\mathllap{\color{#2}\circ}}
\newcommand\goal[1]{{\large$\filledcirc{#1}$}}
\newcommand\goalb[2]{{\large$\filledcircb{#1}{#2}$}}
\newcommand{\startsq}{%
  \setlength{\fboxsep}{0pt}%
  \setlength{\fboxrule}{1.2pt}%
  \raisebox{0.6pt}[0pt][0pt]{\fcolorbox{orange}{white}{\color{white}{o}}}
}
\newcommand{\vizc}[2]{\footnotesize{P=$#1$, PPL=$#2$}}

\newcommand*\rot{\rotatebox[origin=c]{90}}
\newcommand{\STAB}[1]{\begin{tabular}{@{}c@{}}#1\end{tabular}}

\title{MultiON: Benchmarking Semantic Map Memory using Multi-Object Navigation}

\author{%
  Saim Wani$^{1}$\thanks{denotes equal contribution} \quad
  Shivansh Patel$^{1,3}$\footnotemark[1] \quad
  Unnat Jain$^{2}$\footnotemark[1] \quad
  Angel X. Chang$^{3}$\quad
  Manolis Savva$^{3}$\\[5pt]
  $^{1}$IIT Kanpur\qquad$^{2}$UIUC\qquad$^{3}$Simon Fraser University\\[5pt]
  \url{https://shivanshpatel35.github.io/multi-ON/}
}

\begin{document}

\maketitle

\begin{abstract}
Navigation tasks in photorealistic 3D environments are challenging because they require perception and effective planning under partial observability. Recent work shows that map-like memory is useful for long-horizon navigation tasks. However, a focused investigation of the impact of maps on navigation tasks of varying complexity has not yet been performed. We propose the \emph{\task} task, which requires \emph{navigation to an episode-specific sequence of objects in a realistic environment}. MultiON generalizes the ObjectGoal navigation task~\cite{anderson2018evaluation,ZhuARXIV2017} and explicitly tests the ability of navigation agents to locate previously observed goal objects. We perform a set of \task experiments to examine how a variety of agent models perform across a spectrum of navigation task complexities. Our experiments show that: i) navigation performance degrades dramatically with escalating task complexity; ii) a simple semantic map agent performs surprisingly well relative to more complex neural image feature map agents; and iii) even oracle map agents achieve relatively low performance, indicating the potential for future work in training embodied navigation agents using maps. Video summary: \url{https://youtu.be/yqTlHNIcgnY}
\end{abstract}

\section{Introduction}

Recent work on embodied AI agents has made tremendous progress on tasks such as visual navigation~\cite{gupta2017cognitive,habitat19iccv,ZhuARXIV2017}, embodied question answering~\cite{das2018eqa,Wijmans2019EQAPhoto}, and natural language instruction following~\cite{anderson2018vision}.
This progress has been enabled by the availability of realistic 3D environments and software platforms that simulate navigation tasks within such data~\cite{ai2thor,SavvaARXIV2017Minos,habitat19iccv,xia2018gibson}.
Large-scale training has led to near-perfect agent performance for basic visual navigation tasks under certain assumptions~\cite{wijmans2020dd}.

At the same time, memory has emerged as a key bottleneck for further progress in longer-horizon tasks.
Embodied navigation agents are surprisingly brittle, with complete failure being common when the path to be navigated and number of locations to be visited is modestly increased~\cite{anderson2018vision,habitat19iccv,wijmans2020dd}.
In response, much prior work has incorporated some form of map-like memory to aid performance in longer-horizon tasks~\cite{bhatti2016playing,chaplot2020neural,GordonCVPR2018,gupta2017cognitive,gupta2017unifying,henriques2018mapnet,parisotto2018neural,savinov2018semi,zhang2017neural}.
This prior work has mostly focused on proposing approaches for constructing the maps from partial observations of the environment, and architectures for leveraging the map representation.
Though map-like memory structures need not be optimal for learning-based agents, they bring the advantage of a widely-used spatial abstraction and human interpretability.
They also impose inductive bias tied to the structure of interiors that have been shown to outperform implicit memory architectures in a variety of navigation tasks~\cite{beeching2020egomap,chaplot2020object,chaplot2020learning,chaplot2020neural}.

We focus on studying \emph{what} information is useful in maps, under perfect agent localization and map aggregation.
This is in contrast with prior work which focuses on \emph{how} to aggregate information under imperfect localization~\cite{chaplot2020neural,henriques2018mapnet}.
Perfect localization is a simplifying assumption that decouples the question of what is useful in a map representation from dealing with noisy sensing and actuation.

We propose \emph{\task}, a task framework that involves \emph{navigation to an ordered sequence of objects} placed within realistic 3D interiors.
This task framework generalizes earlier visual navigation tasks such as the PointGoal and ObjectGoal tasks~\cite{anderson2018evaluation}, by defining an arbitrary sequence of semantically distinct objects as navigation goals.
Using a set of \task experiments, we benchmark agents with egocentric mapping, agents using oracle maps, and agents possessing no map memory.

The \task framework allows for direct control of task difficulty by generating navigation episodes with multiple goal objects.
At the same time, the base navigation task is simple, allowing us to focus our investigation on map utilization in visual navigation.
Though there is much prior work involving higher-level understanding, planning and reasoning (e.g., natural language understanding~\cite{anderson2018vision,ChaplotARXIV2017,hill2017understanding}, embodied question answering~\cite{das2018eqa,yu2019multi}, interaction with the environment~\cite{shridhar:cvpr20}, leveraging world priors~\cite{yang2018visualsemantic}) we avoid introducing these elements to limit confounding factors that influence the impact of map memory on visual navigation.
In summary, we make the following contributions:
\begin{compactitem}
    \item Propose the \task task framework to allow for controlled, systematic investigation of visual navigation in 3D environments.
    \item Benchmark the impact of egocentrically-constructed map representations and oracle maps on navigation performance across a breadth of navigation task complexities.
    \item Show that agent navigation performance drops dramatically with task complexity, that a simple semantic map construction approach outperforms a more complex neural image feature map agent, and that even oracle map agents achieve relatively low performance, indicating the potential for future work in this direction.
\end{compactitem}

\section{Related Work}

\xhdr{Embodied AI agents.}
There has been much interest in studying AI agents in simulated 3D environments~\cite{ammirato2016avd,BrodeurARXIV2017,chang2017matterport,ai2thor,xia2018gibson,stanford2d3d,habitat19iccv,xia2019interactive,weihs2020allenact}.
Learning how to tackle a variety of tasks from egocentric perception is a common theme in this area.
Embodied navigation is a family of closely related tasks where the goal is to navigate to specific points, objects or areas, respectively PointGoal, ObjectGoal, and AreaGoal~\cite{anderson2018evaluation}.
In PointGoal navigation~\cite{habitat19iccv,chaplot2020learning,gordon2019splitnet}, the agent has access to a displacement vector to the goal at each time step, largely obviating the need for long-horizon planning and map-like memory.
PointGoal has been extensively studied and recently `solved'~\cite{wijmans2020dd}.
In contrast, ObjectGoal has not been well-studied despite being introduced in early work by \citet{ZhuARXIV2016} and explored in natural language grounding settings~\cite{ChaplotARXIV2017,hill2017understanding}.  

We generalize the ObjectGoal navigation task to allow systematic investigation of agents using spatial and semantic map representations for grounding egocentrically-acquired visual information.

\xhdr{Long-horizon embodied agent tasks.}
There has been relatively little work on training embodied agents for long-horizon tasks.
\citet{mirowski2018learningcity} propose an LSTM-based architecture for a \textit{courier task}, where an agent navigates to a series of random locations in a city.
Multi-target Embodied QA~\cite{yu2019multi} extends single object EQA~\cite{DasCVPR2018} to answer questions such as ``Does the table in the bedroom have same color as the sink in the bathroom?'', which require visually navigating to exactly two objects.
\citet{fang2019scene} propose a transformer-based architecture for a variety of tasks (collision avoidance, scene exploration, and object search), of which object search is most related to our work.
In object search the agent navigates to televisions, refrigerators, bookshelves, tables, sofas, and beds in the scene with no specific order.
Note that the position and number of the objects within the scene is not controlled, and thus the complexity of the task is fixed and a property of the dataset.
\citet{beeching2020egomap} use spatial memory for several tasks including an `ordered k-item' task~\cite{beeching2019deep} within a gridworld-like game environment~\cite{kempka2016vizdoom}.
The item list is kept fixed across all experiments.

Our \task task requires navigation to an episode-specific ordered list of objects within a challenging photorealistic 3D environment.
The episode-specificity requires grounding object class labels to their visual appearance.
The ordered aspect requires storing information on previously encountered objects that need to be retrieved later (e.g., seeing the third object on the way to the first one).
In contrast, `ordered k-item'~\cite{beeching2019deep} is not episode-specific and `object search'~\cite{fang2019scene} is neither episode-specific nor ordered (see \Cref{tab:object-nav-tasks}).
To the best of our knowledge, \task is the first task designed to benchmark long-horizon embodied navigation in photo-realistic environments.

\xhdr{Maps.} 
Maps which encode semantic information about the environment broadly fall into two categories: spatial~\cite{gupta2017cognitive,parisotto2018neural} and topological~\cite{chaplot2020neural,savinov2018semi}.
\textit{Spatial maps} are tensors where two dimensions align with an environment's top-down layout.
Other dimensions contain features encoding semantic information for a particular location of the environment.
Spatial maps built with Simultaneous Localization and Mapping (SLAM) have been used for tasks such as exploration~\cite{zhang2017neural}, and playing FPS games~\cite{bhatti2016playing}.
Work on deep RL for embodied navigation has leveraged egocentric maps~\cite{gupta2017cognitive}, and a combination of egocentric and allocentric maps~\cite{gupta2017unifying}.
\citet{GordonCVPR2018} store object detection probabilities in spatial maps to aid recognition for question answering.
More recent work investigates learning and use of semantic grid-based maps for navigating to a single ObjectGoal~\cite{chaplot2020object,cartillier2020semantic}.
\citet{chaplot2020object} perform first person view semantic segmentation and project the segmentation onto the top-down map using a differentiable mapping module.
\citet{cartillier2020semantic} study whether it is better to segment and then project, or to project and then segment directly on the top-down map.
These works assume perfect knowledge of agent egomotion (i.e. perfect localization).
MapNet~\cite{henriques2018mapnet} does not make this assumption and integrates information over time for use in an end-to-end differentiable architecture.
In contrast, \textit{topological maps} such as those proposed by \citet{chaplot2020neural,savinov2018semi} do not align visual information to the environment's top-down layout.
Instead, they store landmarks (e.g., specific input frame) as nodes and their connectivity as edges.
Additionally, prior knowledge of scene layout is used as a knowledge graph in various works~\citep{yang2018visualsemantic,wu2019bayesian}.

Despite this rich body of work using maps for embodied AI, there has been no systematic investigation of what map information benefits the core task of navigation, and how much value egocentrically acquired maps add relative to oracle maps or no map representations.
We focus on these questions in this paper.

\section{Task}
\label{sec:task}

\begin{table}
    \centering
    \caption{Comparison of \task to navigation tasks with multiple object goals from prior work. Note that prior work does not adopt a \found action, so incidental navigation to a goal is treated as success. Moreover, the set of objects is held fixed, with no episode-specifity for the goal objects.}
    \label{tab:object-nav-tasks}
    \resizebox{\linewidth}{!}{
    \begin{tabular}{@{}lccccc@{}}
        \toprule
        Task & Environment & Position control & \# Objects & Goal & Evaluation \\
        \midrule
        Search~\cite{fang2019scene} & Synthetic & \redx & $6$ & all in any order & reward, classes found \\
        Ordered $k$-item~\cite{beeching2019deep} & VizDoom~\cite{kempka2016vizdoom} & \greencheck & $4$ or $6$ & fixed set, in order & reward \\
        MultiON (ours) & Habitat~\cite{habitat19iccv} & \greencheck & arbitrary & episode-specific ordered set & success \& efficiency metrics \\
        \bottomrule
    \end{tabular}
    }
\end{table}

Here we define the \task task in detail.
In an episode of \task, the agent must navigate to an ordered sequence of objects $G$ placed within the environment, where $G_n$ is the $n$-th object in the sequence.
The number of objects $m$ allows us to vary the complexity of the navigation episode and create more `long-term' navigation scenarios.
We use \mon{m} to refer to an episode with $m$ ordered goal objects.
The $m$ objects are selected from a set of $k$ available objects where $k \geq m$.

During the episode, the agent receives at each step $t$: i) egocentric RGB and depth sensor images $o_t$; and ii) current goal object specified as a $k$-dimensional one-hot vector $g_t$.
The agent must navigate to the vicinity of the current goal object $G_n$ and declare the object has been found using a \found action.
The goal object specification is then updated, and the agent must continue to locate the next object $G_{n+1}$ in the sequence $G$ until all objects have been located.
If the \found action is called while not in the vicinity of the current goal object, or if the agent has taken a threshold maximum number of actions the episode is declared a failure.
Otherwise, the episode is successful.

The \task task is a generalization of the ObjectGoal navigation task proposed by \citet{ZhuARXIV2016} and \citet{anderson2018evaluation} allowing more controlled complexity through the selection of multiple object goals.
The flexibility of \task allows us to systematically study the benefit of map memory.
To summarize, some of the key differentiating characteristics of our task are:
\begin{compactitem}
    \item A different goal sequence is specified to the agent per episode necessitating grounding of object identity to visual inputs, in contrast to \citet{fang2019scene,beeching2019deep} where the goals are fixed and predetermined.
    \item The goal objects are inserted into realistic 3D environments at controlled locations. Programmatic placement and category selection of the objects allows us to precisely control the geodesic distances between target objects and the resulting episode complexity (in contrast to relying on pre-existing objects~\cite{fang2019scene,chaplot2020object}).
    \item The agent needs to explicitly declare when the goal object has been \found, in line with the recommendation of \citet{anderson2018evaluation}.
    \item Task evaluation metrics are defined independent of reward formulations, allowing future work to explore different reward structures and other methods such as imitation learning.
\end{compactitem}

In this paper, we adopt the following specific instantiation of \task.
The agent is embodied with a cylindrical body of radius $0.1\text{m}$ and height $1.5\text{m}$.
The agent action space has four actions: \{\textsc{forward}, \textsc{turn-left}, \textsc{turn-right}, \textsc{found}\} where turns are by $30^{\circ}$ and \textsc{forward} moves forward by $0.25\text{m}$.
An \mon{m} episode specifies the agent start position and orientation, and the positions of $m$ goal objects.
The objects are randomly sampled without replacement from a set of $8$ cylinder objects with height $0.75\text{m}$ and radius $0.05\text{m}$ and different colors: \{red, green, blue, cyan, magenta, yellow, black, white\}.
As such, there are $\binom{8}{m} \cdot m!$ possible goal sequences $G$.

\section{Agent Models}

We focus on examining the benefit of incorporating spatial maps for embodied navigation agents.
We first introduce a general network architecture that is shared by all the agent models we study.
Then, we specify variations on this base architecture for each model, comparing agent models with no map representation to ones with access to a ground truth map, as well as agents with learned maps.

\begin{figure}
\centering
\includegraphics[width=\textwidth]{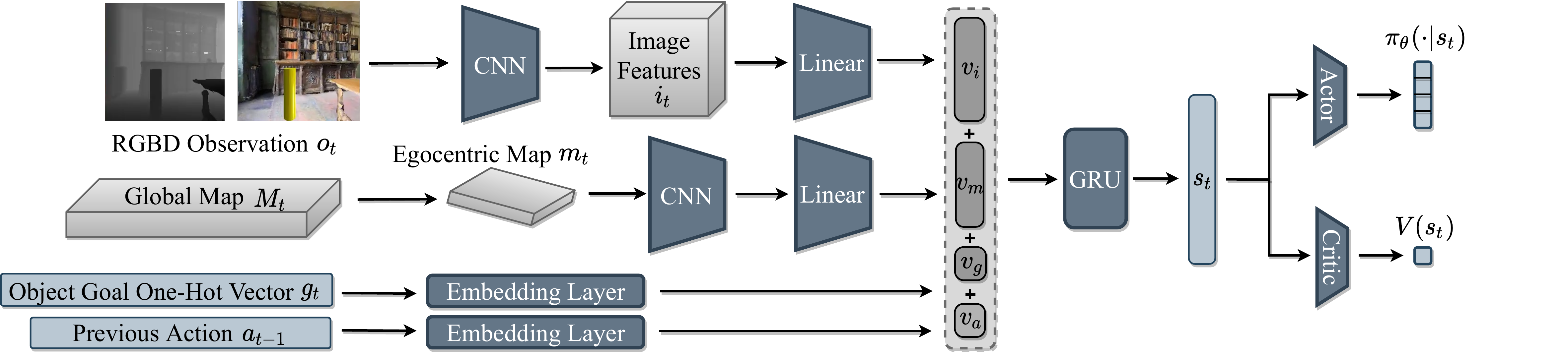}
\caption{
\textbf{Base agent architecture.} At each time step $t$, RGBD frames $o_t$, an egocentric map $m_t$, the object goal $g_t$, and the previous action $a_{t-1}$, are encoded into a concatenated embedding $[v_t, v_m, v_g, v_a]$ passed to a GRU to construct state $s_t$ for an actor-critic pair used by the agent.
}
\label{fig:architecture}
\end{figure}

\subsection{Base Agent Architecture}

\xhdr{Inputs.}
At each time step the agent receives a number of sensory inputs.
These include a visual observation $c_t$ (RGB image of $256\times256\times3$), a corresponding $256\times256\times1$ depth observation $d_t$, and the object goal specified as a one-hot vector $g_t$.
The agent also has access to the action at the previous step $a_{t-1}$, a common strategy for augmenting purely reactive agents~\cite{SavvaARXIV2017Minos,das2018eqa, wang2018look}.
We denote the agent's sensory observations as $o_t = (c_t, d_t)$.
All agents use either an explicit spatial grid map representation, or have implicit memory: \NoMapAll.
We refer to the map of the entire environment as the `global map' and denote it with $M$.
The agent models operate with an egocentric map $m_t$, representing a rotated and cropped view of the global map centered and oriented at the agent's location and orientation at step $t$.
\Cref{fig:architecture} illustrates this agent architecture.

\xhdr{Network components.}
The RGBD sensory input $o_t$ is transformed into image features $i_t$ and linearly embedded into $v_i$ using a CNN block and a linear layer.
Similarly, the egocentric map $m_t$ is embedded into $v_m$.
These linear embeddings are then concatenated with $16$-dimensional embeddings $v_g$ and $v_a$ for the current goal definition $g_t$ and previous action $a_{t-1}$.
The concatenated representation $v_t=\concat(v_i, v_m, v_g, v_a)$ is transformed using a Gated Recurrent Unit (GRU)~\cite{cho2014learning} to give the final state representation $s_t$ at step $t$.
The GRU provides an implicit memory over the embedded inputs.
Let $\theta$ represent the parameters of this end-to-end trainable neural architecture.
The output of the GRU is used to predict the action policy $\pi_{\theta}(\cdot|s_t)$ and the approximate value function $V(s_t)$.
We next describe the critical difference between agent models which lies in how the global map $M$ is constructed and utilized.

\subsection{Agents}

We investigate the performance of different agents on the \task task.
We select representative agents with implicit memory (\NoMap, \FRMQN\cite{oh2016control}, \SMT\cite{fang2019scene}) and compare them against agents that use explicit grid map representations, with both oracle maps and learned maps.
In addition, we consider two random action baselines, with one variant selecting the \found action randomly, and the other using oracle \found.

\xhdr{Implicit memory baselines}

\xhdr{\NoMap}:
An agent that does not use explicit map information.
This agent establishes a comparison point for the performance of our base architecture without maps as trained through PPO~\cite{schulman2017proximal}.

\xhdr{\FRMQN}:
This agent is based on \citet{oh2016control}'s architecture which stores observation embeddings of the previous episode steps in an LSTM-like memory as key-value pairs, and reads from this memory using soft attention.

\xhdr{\SMT}:
This agent is based on \citet{fang2019scene}.
Here, an embedding of each observation is again stored in memory, and an encoder-decoder network is used to retrieve relevant information from the memory.
We use the same CNN as in our other agent models to produce the observation embeddings.

\xhdr{Oracle baselines}

\xhdr{\OracleMap}:
This agent gets the entire global ground truth (i.e. oracle) map $M_\text{\tiny GT}$ for the current environment.
The global map is a $300\times300$ cell grid with $0.8\text{m}\times0.8\text{m}$ cells.
Each cell contains two channels: i) occupancy information; and ii) goal object category.
Concretely, the occupancy channel stores a $16$-dimensional learnable neural embedding of whether a cell is navigable, non-navigable or outside the scene.
The goal object category channel also stores a $16$-dimensional embedding corresponding to the category of the object occupying that cell~\cite{rodriguez2018beyond, guo2016entity}.
There are $k+1$ categories: $k$ goal categories, and a `no goal' category.
This oracle map agent provides an upper bound for the benefit of the spatial map to agent navigation performance in \task.

\xhdr{\OracleEgoMap}:
Same as \OracleMap, except that the global map is progressively `revealed' from an egocentric perspective as the agent explores the environment.
Thus, at step $t$ the agent has access to a partially revealed global map $M_t$.
Note that visibility is accounted for by ray tracing and checking against the agent's current depth frame $d_t$ so that the agent cannot see behind obstacles or outside its field of view.
Please refer to \Cref{sec:details} for details.
This agent represents an idealized egocentric mapping agent (i.e. perfect sensing of the environment within a limited field of view)

\xhdr{Learned map agents}

\xhdr{\ObjRecogMap}:
This agent does not use any oracle map information.
It progressively constructs a global map storing predicted categories of goal objects visible in the sensory inputs.
An auxiliary $(k+1)-$way classification network is supervised via the ground truth object category at train time to predict which goal object is in view.
If no goal is within the field-of-view ($79^{\circ}$) and $2.5\text{m}$ of the agent, a `no object' label is assigned.
The embedding of this predicted category is used to populate the global map $M_t$ at the grid cell corresponding to the agent's position.
See \Cref{sec:details} for details about this auxiliary classification task.
This agent is representative of approaches that embed egocentrically observed object information into a spatial map.
Surprisingly, to our knowledge there is no prior work employing this straightforward approach.
Our scheme is similar to the one employed by \citet{GordonCVPR2018}, where object detection probabilities are encoded into a map (in addition to occupancy, coverage, and navigation intent).
However, we use a simpler object classifier rather than an object detector (i.e. the output is a label rather than a detection localizing the object).

\begin{figure}
\centering
\includegraphics[width=\textwidth]{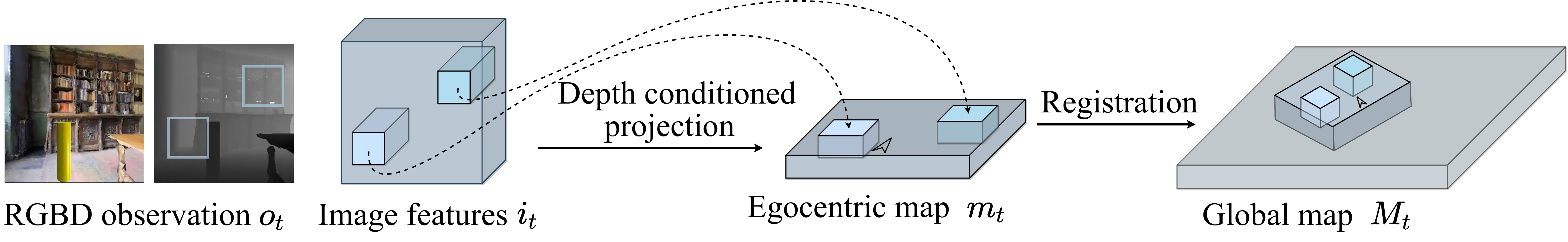}
\caption{
\textbf{Projection and map registration module.}
Input depth $d_t$ and color $c_t$ frames are encoded into image features $i_t$ which are then projected into an egocentric map $m_t$.
This map is then registered onto the global map $M$.
}
\label{fig:egocentric_projection}
\end{figure}

\xhdr{\ProjNeuralMap}:
This agent neurally projects image features to the map using the projection module of the MapNet architecture~\cite{henriques2018mapnet}.
\Cref{fig:egocentric_projection} illustrates the approach: i) depth-conditioned projection of the image features $i_t$ to generate egocentric map $m_t$; and ii) registering $m_t$ to a global map $M_t$ accumulated over time.
Concretely, we use the depth buffer $d_t$ to appropriately project each feature $i_t(i,j,\cdot)$ onto the egocentric grid $m_t$.
Note the agent is always at the mid-bottom of $m_t$.
This egocentric $m_t$ is integrated into $M_t$ via a registration function $\mathsf{R}(m_t, M_t|p_t)$, where $p_t$ is the agent's position and orientation.
We integrate neural features into cells that are already filled using element-wise max-pooling.
Refer to \Cref{sec:details} for more details.
This agent is a representative of approaches storing neural image features in the map.

\xhdr{\EgoMap}:
This agent is adapted from \citet{beeching2020egomap}, who use projection architecture based on MapNet~\cite{henriques2018mapnet} to construct a semantic map.
It differs from \ProjNeuralMap in using an attention-based read mechanism to read from the semantic map.

\xhdr{Random baseline agents}

\xhdr{\Rand}:
This agent chooses an action randomly from the action space: \{\textsc{forward}, \textsc{turn-left}, \textsc{turn-right}, \textsc{found}\}.
It establishes a navigation performance lower bound and reveals the difficulty of variations of the \task task with time-limited random search.

\xhdr{\RandOracleFound}:
Chooses an action randomly from: \{\textsc{forward}, \textsc{turn-left}, \textsc{turn-right}\}.
When the agent is within a threshold distance of the current goal, an `oracle' calls the \found action.
This implies the episode terminates either on success, or if the agent takes the maximum allowed steps ($2@500$ steps for all experiments here).

\subsection{Training}

\xhdr{Reward structure.}
The agent receives the following reward $r_t$ at each timestep $t$:
\[
r_t = 1_{[\text{reached-goal}]} \cdot r_{\text{goal}} + r_{\text{closer}} + r_{\text{time-penalty}}
=  \begin{cases}
 r_{\text{goal}} + r_{\text{closer}} + \alpha & \text{if goal is found }\\
 r_{\text{closer}} + \alpha & \text{otherwise}
\end{cases},
\]
where $1_{[\text{reached-goal}]}$ is an indicator variable equal to $1$ if a \found action was called within the threshold distance of the current target goal in the current time step, $r_{\text{closer}} = (d_{t-1} - d_t)$ is the decrease in geodesic distance to the current goal at timestep $t$ (in meters), $r_{\text{goal}}$ is the reward for discovery of a goal, and $\alpha$ is a negative slack reward that encourages the agent to take shorter paths.
We use $r_{\text{goal}} = 3.0$ and $\alpha = -0.01$ for all experiments.

\xhdr{Training setup.}
The agent is trained using the reward structure defined above using proximal policy optimization (PPO) \cite{schulman2017proximal}.
All the agent models are trained for $40$ million frames using $16$ parallel threads.
We use $4$ mini-batches and do $2$ epochs in each PPO update.
We do not tweak any other hyperparameters.
The object recognition network in \ObjRecogMap is supervised through a cross-entropy loss on the ground truth goal object category.

\section{Experiments}

Here we describe a series of experiments with the various agent models on the \task task.
We seek to answer the following questions:
(i) how much does an occupancy-based map help over implicit memory?
(ii) does incorporating semantics into the map benefit agent navigation performance?
(iii) how does our agent performance compare with prior work on similar navigation tasks?

\subsection{Datasets}

We generate datasets for the \mon{1}, \mon{2}, and \mon{3} tasks on the Matterport3D~\cite{chang2017matterport} scenes using the standard train/val/test split.
As recommended by \citet{anderson2018evaluation} there is no overlap between train, val and test scenes.
For each \mon{m} dataset the train split consists of $50@000$ episodes per scene, and the validation and test splits each contain $12@500$ episodes per scene.

The agent starting position and goal locations are randomly sampled from navigable points on the same floor within each environment between which there exists a navigable path.
The geodesic distance from the agent starting position to the first goal and between successive goals is constrained to be between $2\text{m}$ and $20\text{m}$.
This ensures that `trivial' episodes are not generated.

\subsection{Metrics}

We evaluate agent performance with several metrics.
These are based on guidelines by \citet{anderson2018evaluation} and followed in other work~\cite{gupta2017cognitive,habitat19iccv,jain2019CVPRTBONE,JainWeihs2020CordialSync,chen2019audio,chaplot2020learning}, but extended to account for multiple object goals, as well as to give partial credit for navigating to a subset of goals.

\xhdr{\Success:}
Binary indicator of episode success.
Episode is successful if the agent navigates to all goals in correct order, calling \found at each goal, within the maximum number of steps allowed.
An incorrectly called \found terminates the episode immediately.

\xhdr{\Progress:}
The fraction of object goals that are successfully \found.
Equal to success for \mon{1}.

\xhdr{\SPL:}
Extension of `Success weighted by Path Length'~\cite{anderson2018evaluation} to the \task task.
Concretely, $\text{SPL} = s \cdot d/\max(p,d)$ where $s$ is the binary success indicator, $p$ is the total distance traveled by the agent, and $d = \sum_{i=1}^{n}d_{i-1,i}$ is the total geodesic shortest path distance from the agent's starting point through each goal position in order with $d_{i-1,i}$ indicating geodesic distance from goal $i-1$ to goal $i$, and goal $0$ being the starting point.

\xhdr{\PPL:}
A version of \SPL based on progress: $\text{PPL} = \bar{s} \cdot \bar{d} / \max(p, \bar{d})$
where $\bar{s}$ is the \Progress value, $\bar{d} = \sum_{i=1}^{l}d_{i-1,i}$ with $l$ being the number of objects found, and $p$ and $d_{i-1, i}$ are defined as before.
The overall path length is weighted by progress, rather than weighing individual subgoal discoveries by their respective path lengths so as to not assign disproportionately high weights to shorter goal-to-goal trajectories within the episode.
\PPL is equal to \SPL for \mon{1}.

\begin{figure}
\resizebox{\linewidth}{!}{
\newcolumntype{C}{>{\centering\arraybackslash} m{2.7cm} }
\begin{tabular}{@{}CCCCC@{}}
\NoMap & \OracleMap & \OracleEgoMap & \ProjNeuralMap & \ObjRecogMap \\
    \includegraphics[width=1\linewidth]{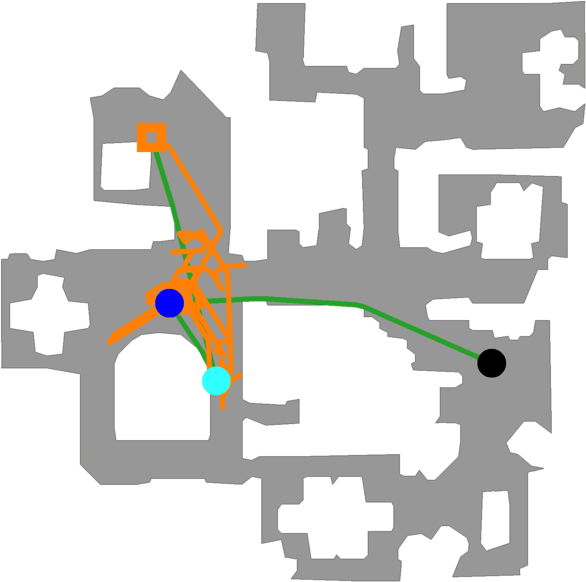} \newline  \vizc{0.66}{0.15}
    &  
    \includegraphics[width=1\linewidth]{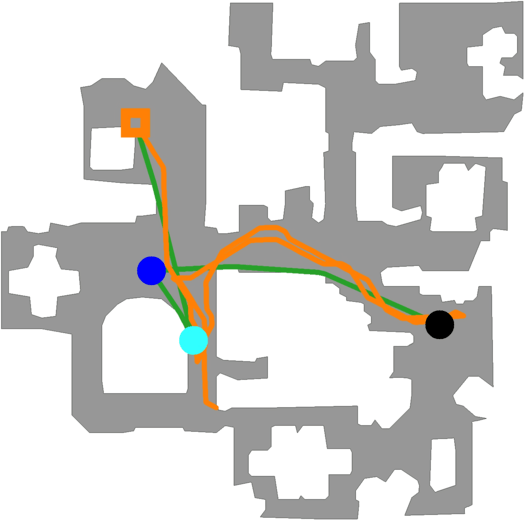} \newline \vizc{0.66}{0.58}  & 
    \includegraphics[width=1\linewidth]{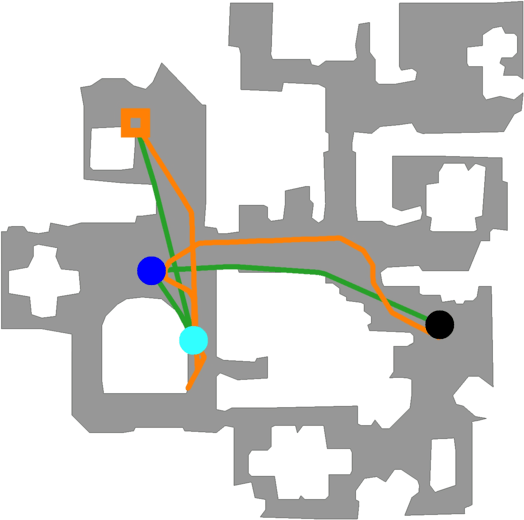} \newline \vizc{1}{0.79}        &   
    \includegraphics[width=1\linewidth]{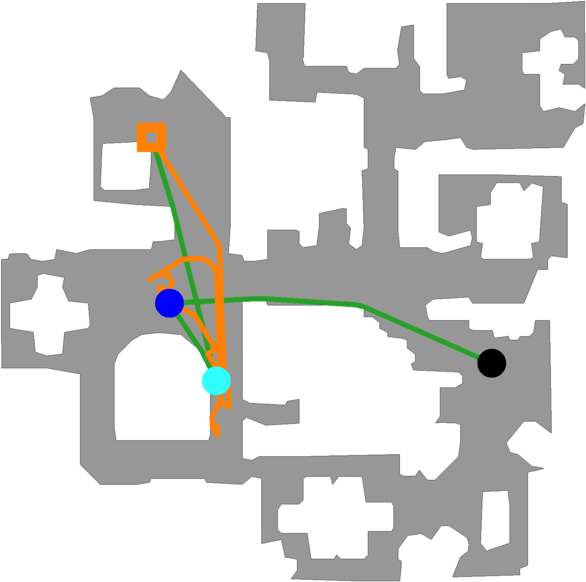}  \newline \vizc{0.66}{0.49}  &   
    \includegraphics[width=1\linewidth]{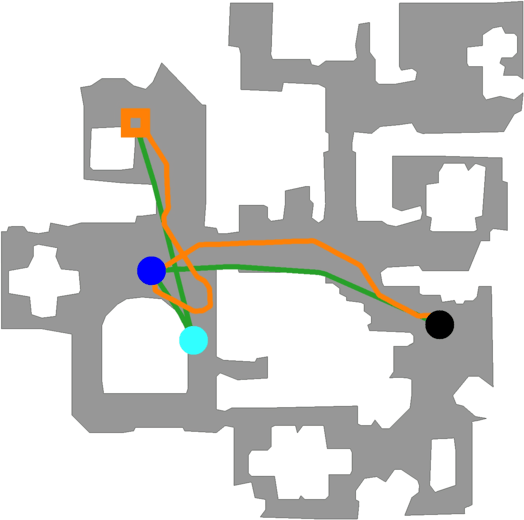}  \newline \vizc{1}{0.98}
    \\ \\
     \includegraphics[trim={9cm 3cm 1cm 2.2cm},clip,width=1\linewidth]{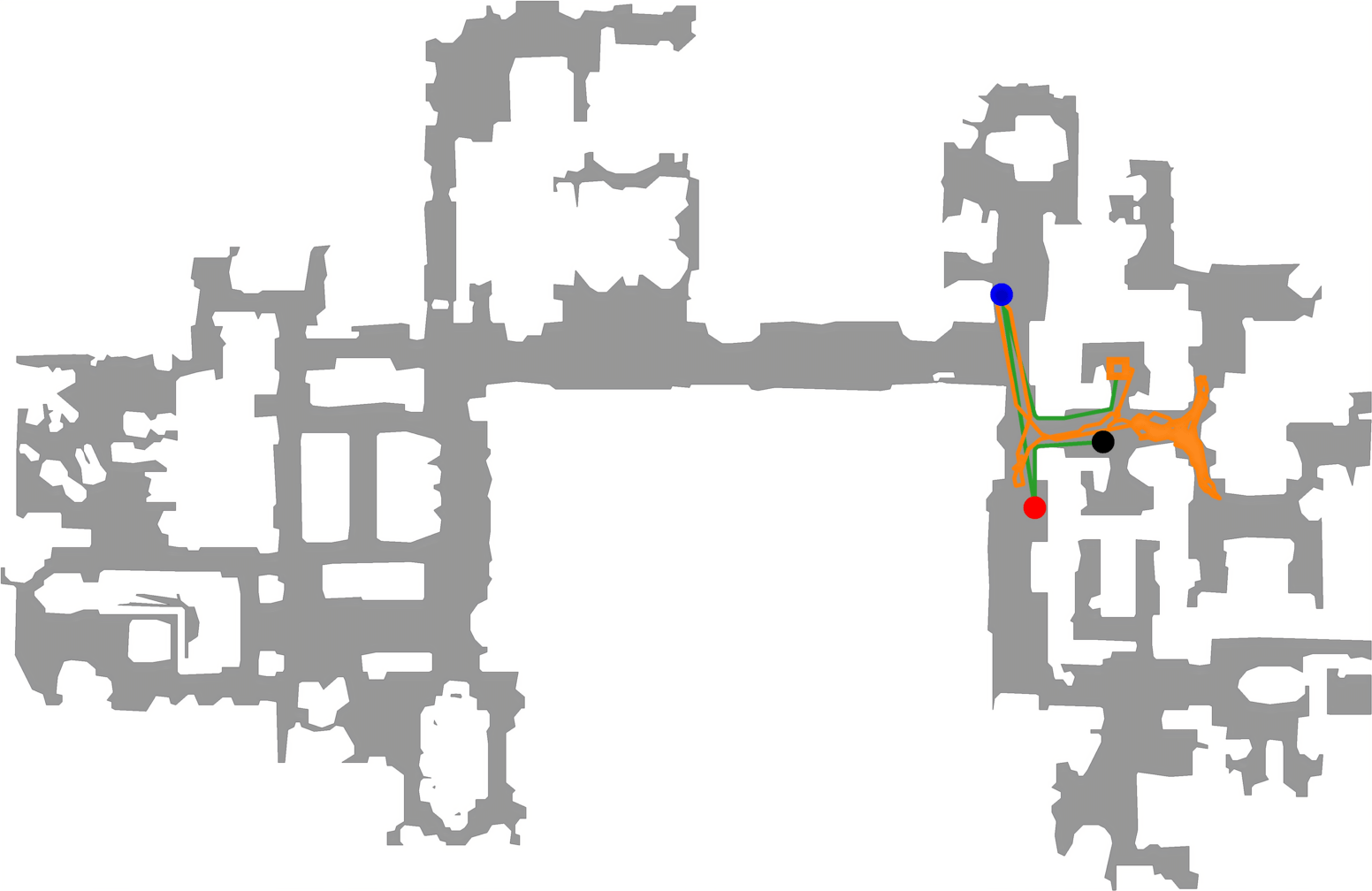} \newline
     \vizc{1}{0.23}   &    
     \includegraphics[trim={9cm 3cm 1cm 2.2cm},clip,width=1\linewidth]{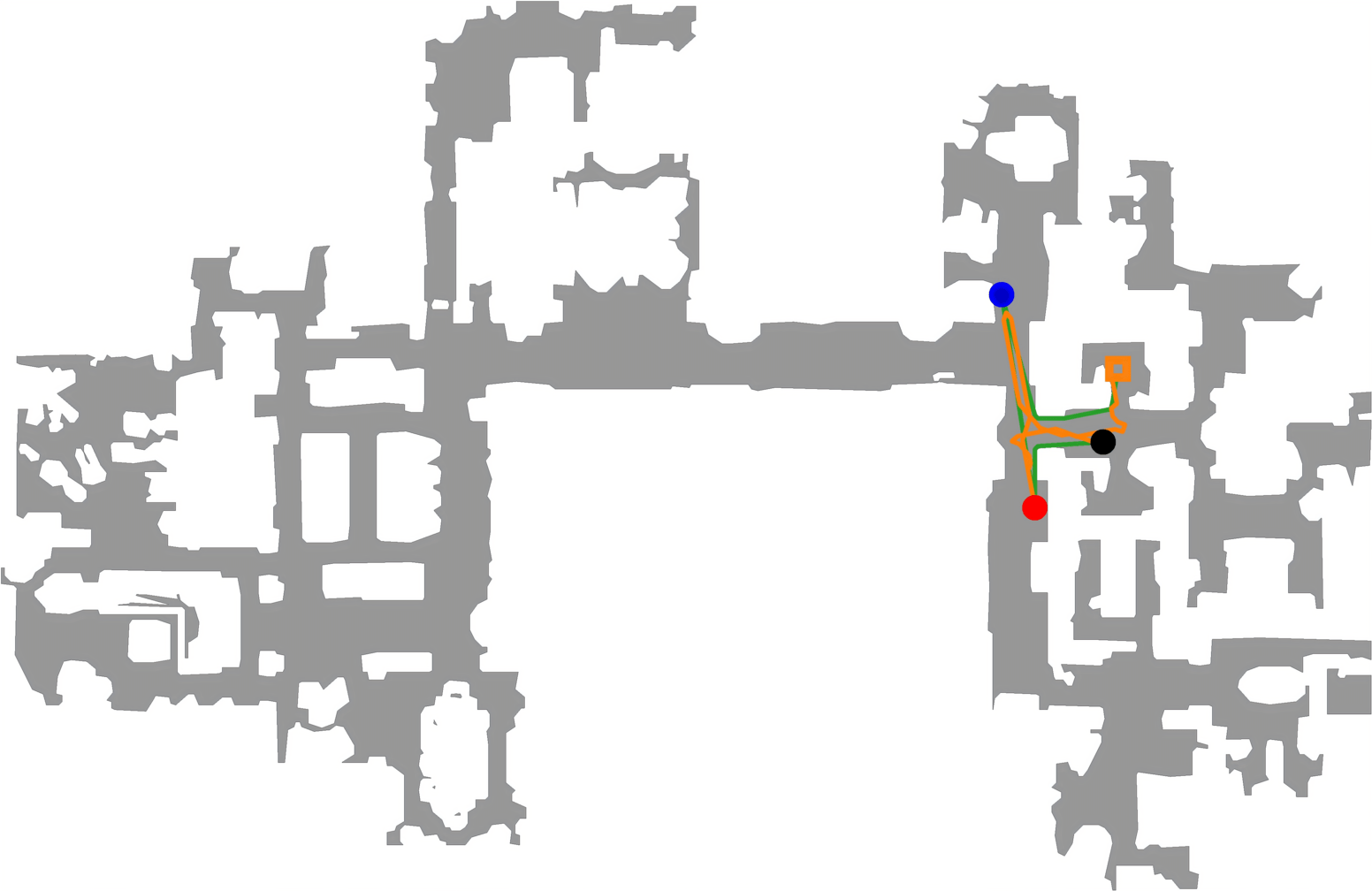} \newline \vizc{1}{0.92}  &  
     \includegraphics[trim={9cm 3cm 1cm 2.2cm},clip,width=1\linewidth]{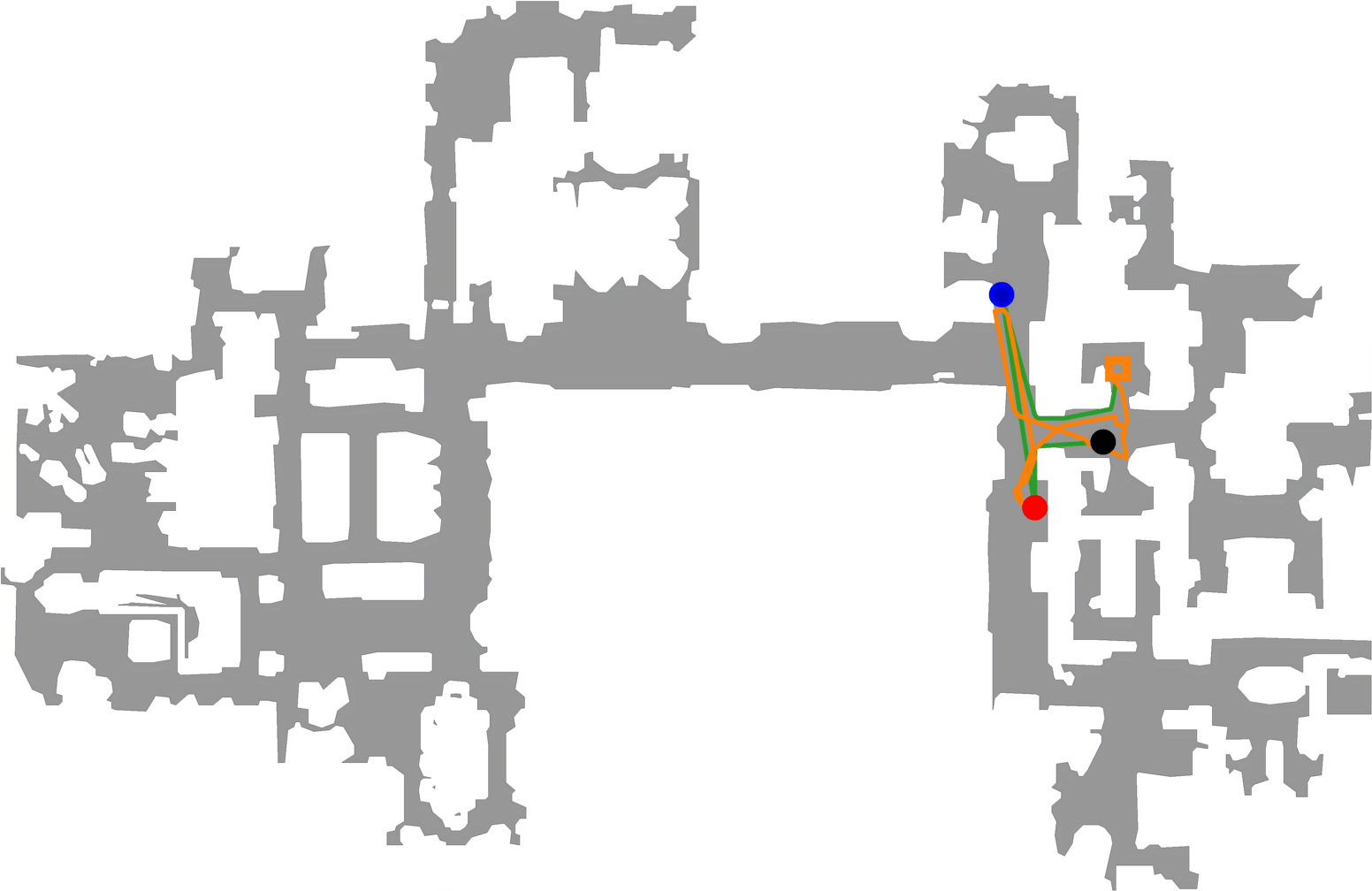} \newline \vizc{1}{0.79}        &   
     \includegraphics[trim={9cm 3cm 1cm 2.2cm},clip,width=1\linewidth]{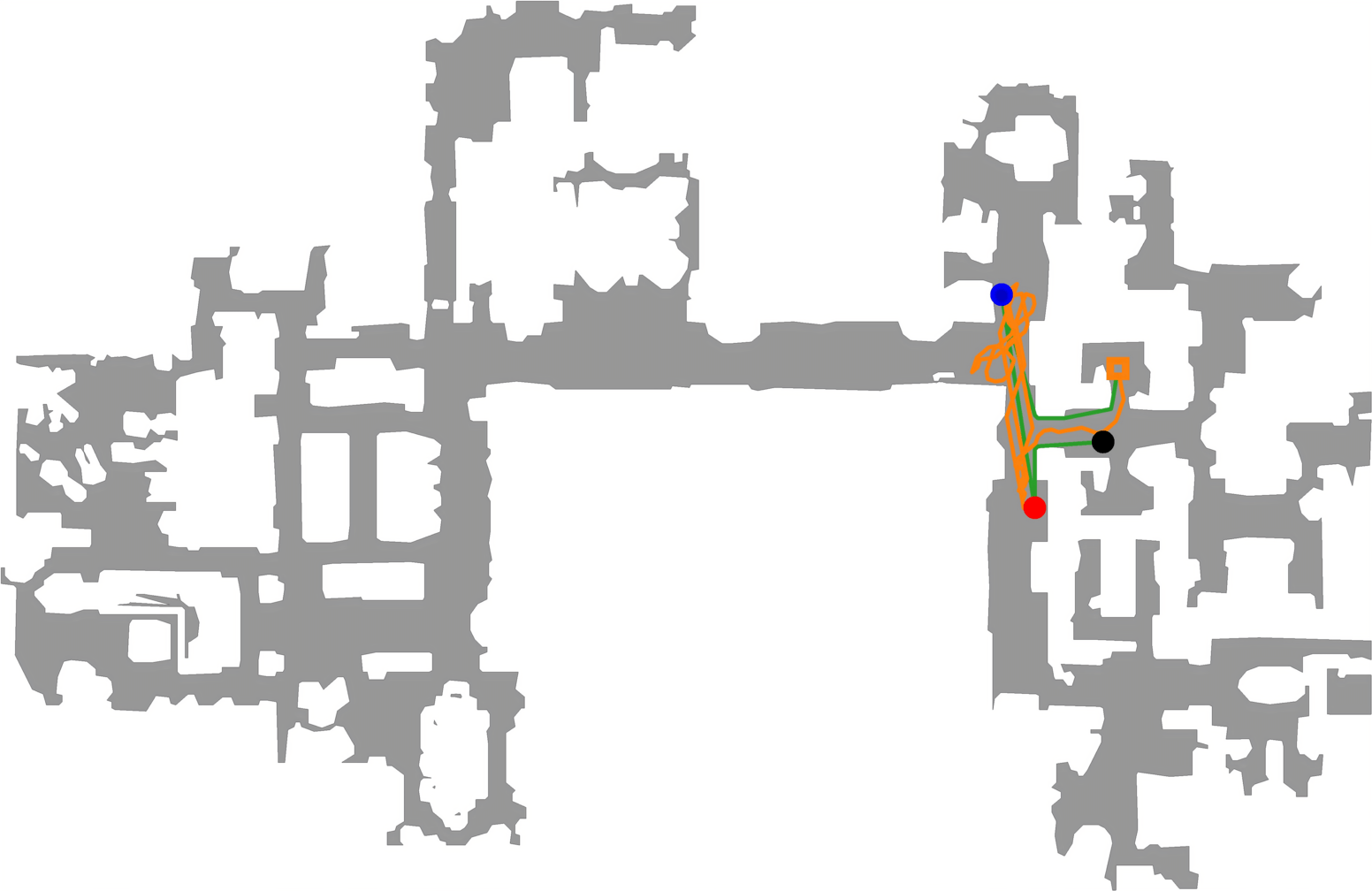}  \newline \vizc{0.66}{0.46}     &   
     \includegraphics[trim={9cm 3cm 1cm 2.2cm},clip,width=1\linewidth]{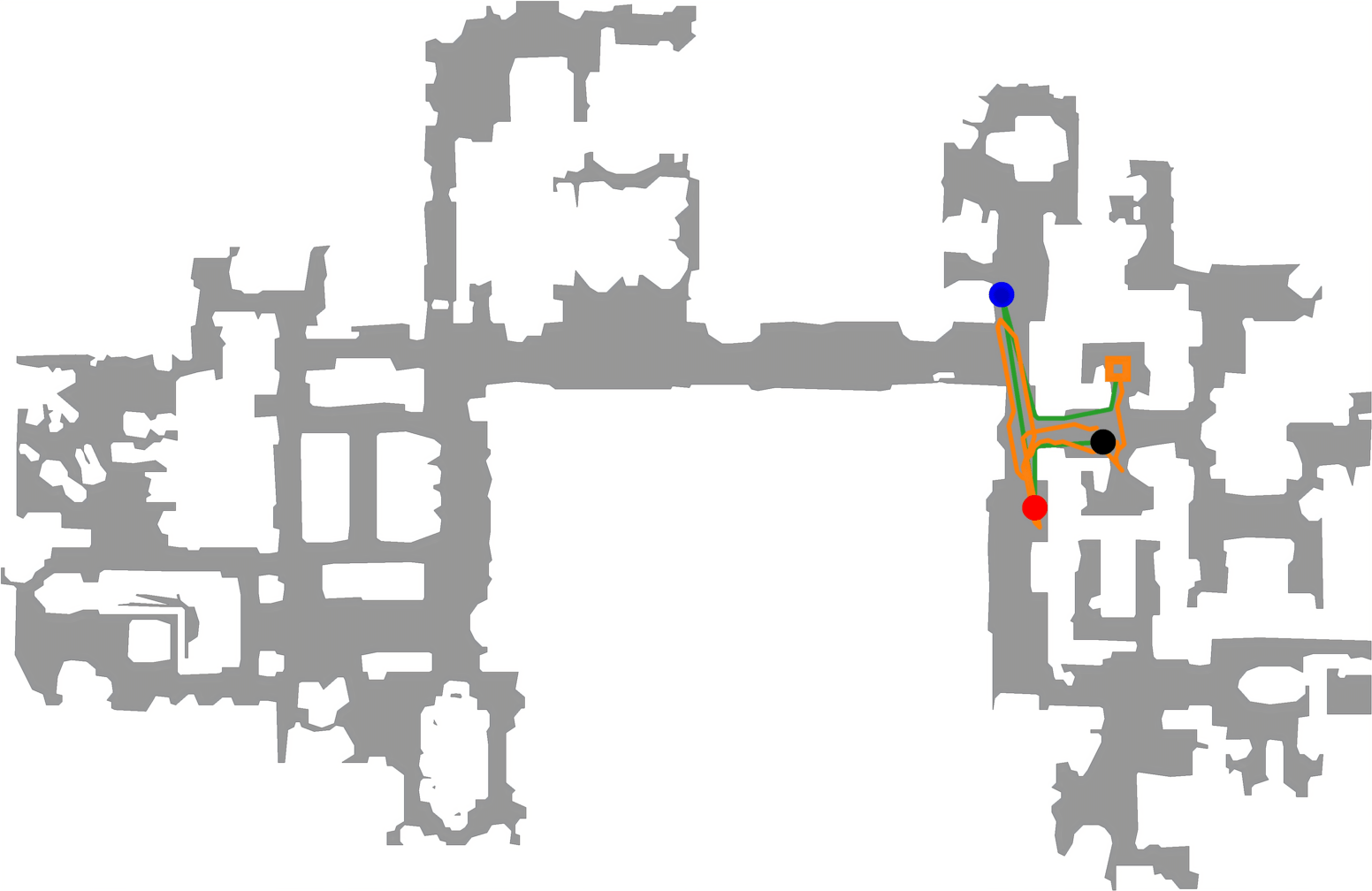} \newline \vizc{1}{0.71}
\end{tabular}
}
\caption{Example episodes for different agents.
{\color{orange}Agent path} and {\color{cadmiumgreen}shortest path} in orange and green, with the start shown by {\startsq} (orange square).
Goal order for top: 1\goal{cyan}, 2\goal{blue}, 3\goal{black}, and bottom: 1\goal{blue}, 2\goal{red}, 3\goal{black}.}
\label{fig:episode-maps}
\end{figure}

\begin{figure}
\hspace{2cm}\mon{1}\hspace{3.8cm}\mon{2}\hspace{3.8cm}\mon{3}\\
\includegraphics[width=0.32\linewidth]{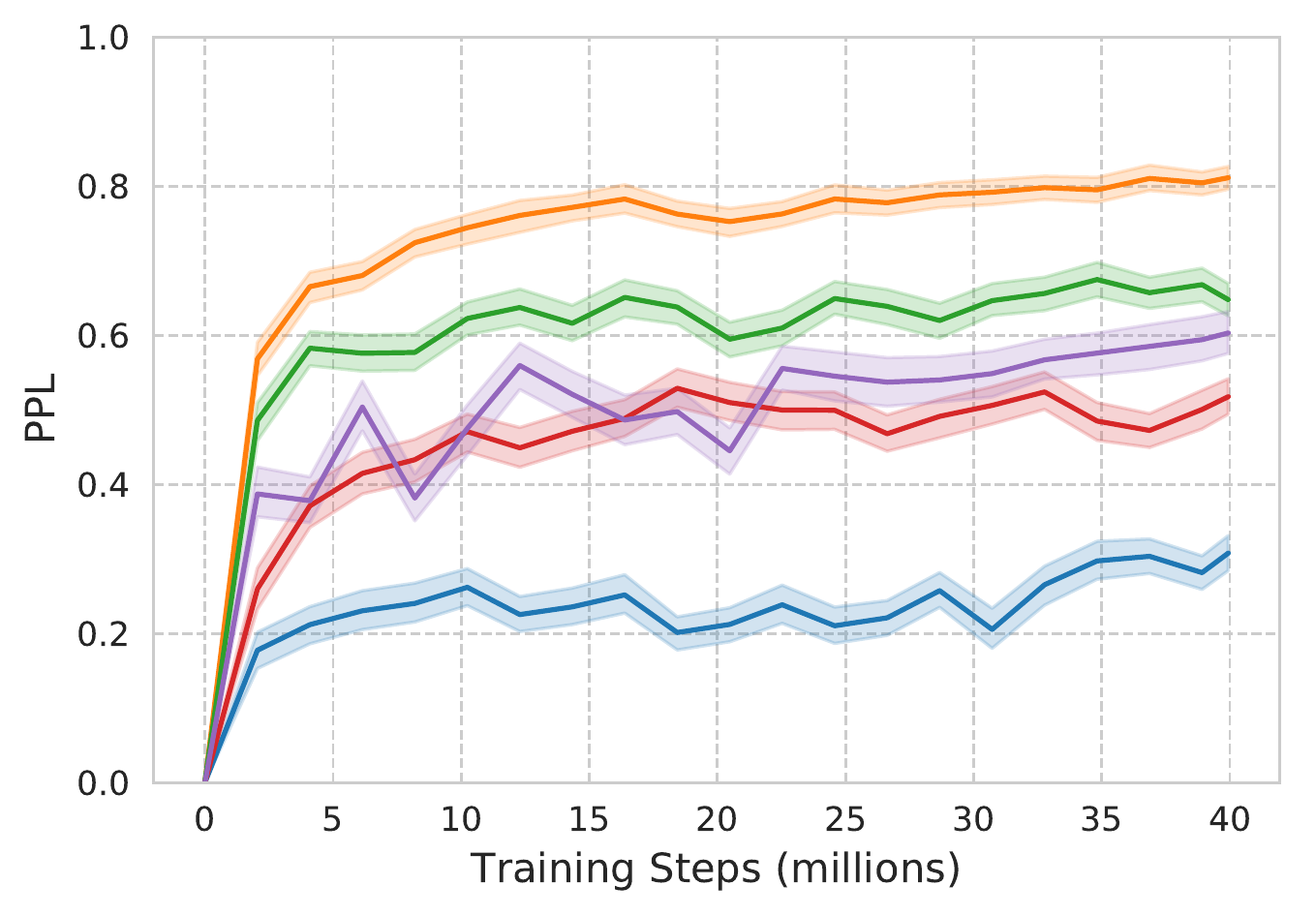}
\includegraphics[width=0.32\linewidth]{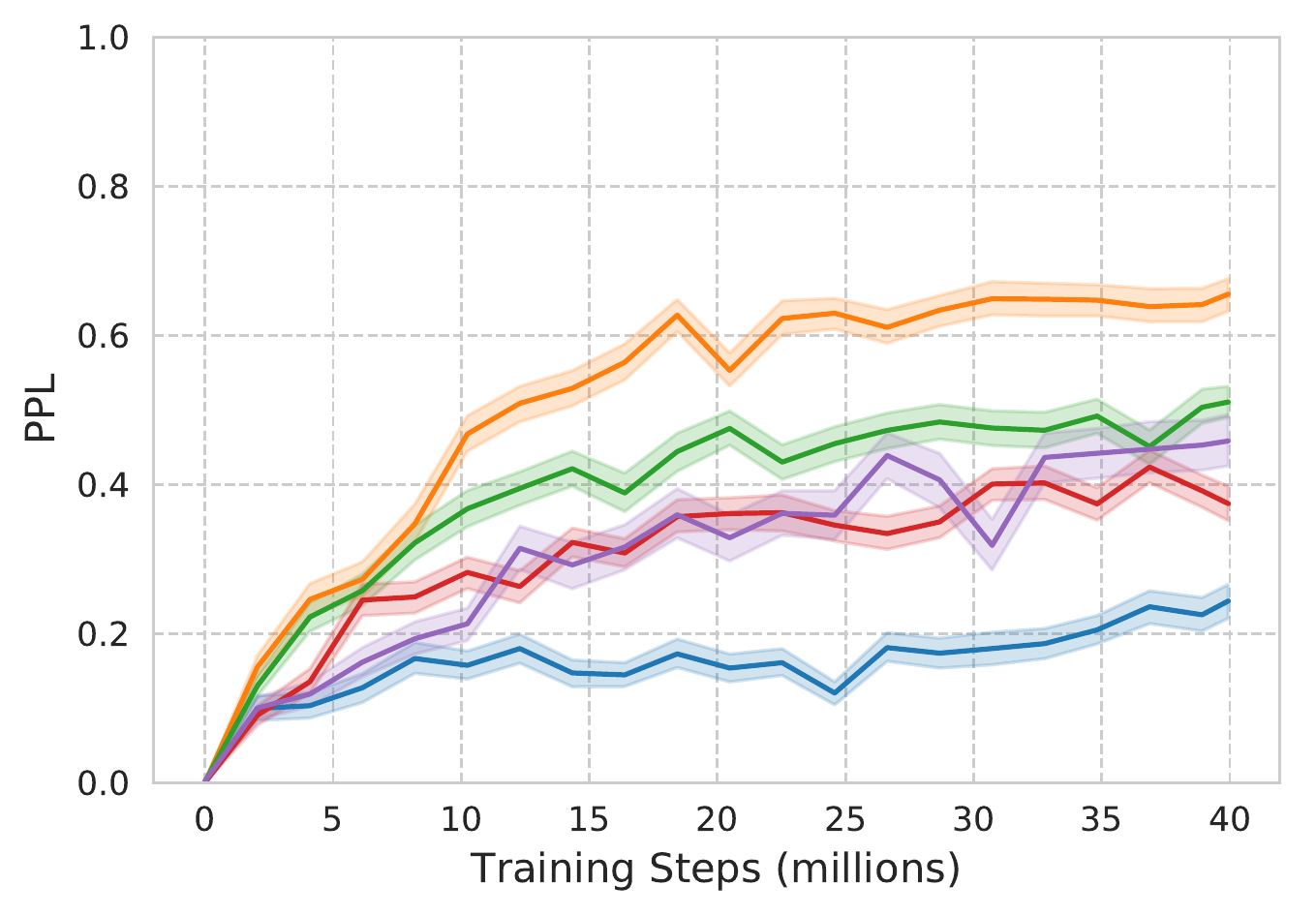}
\includegraphics[width=0.32\linewidth]{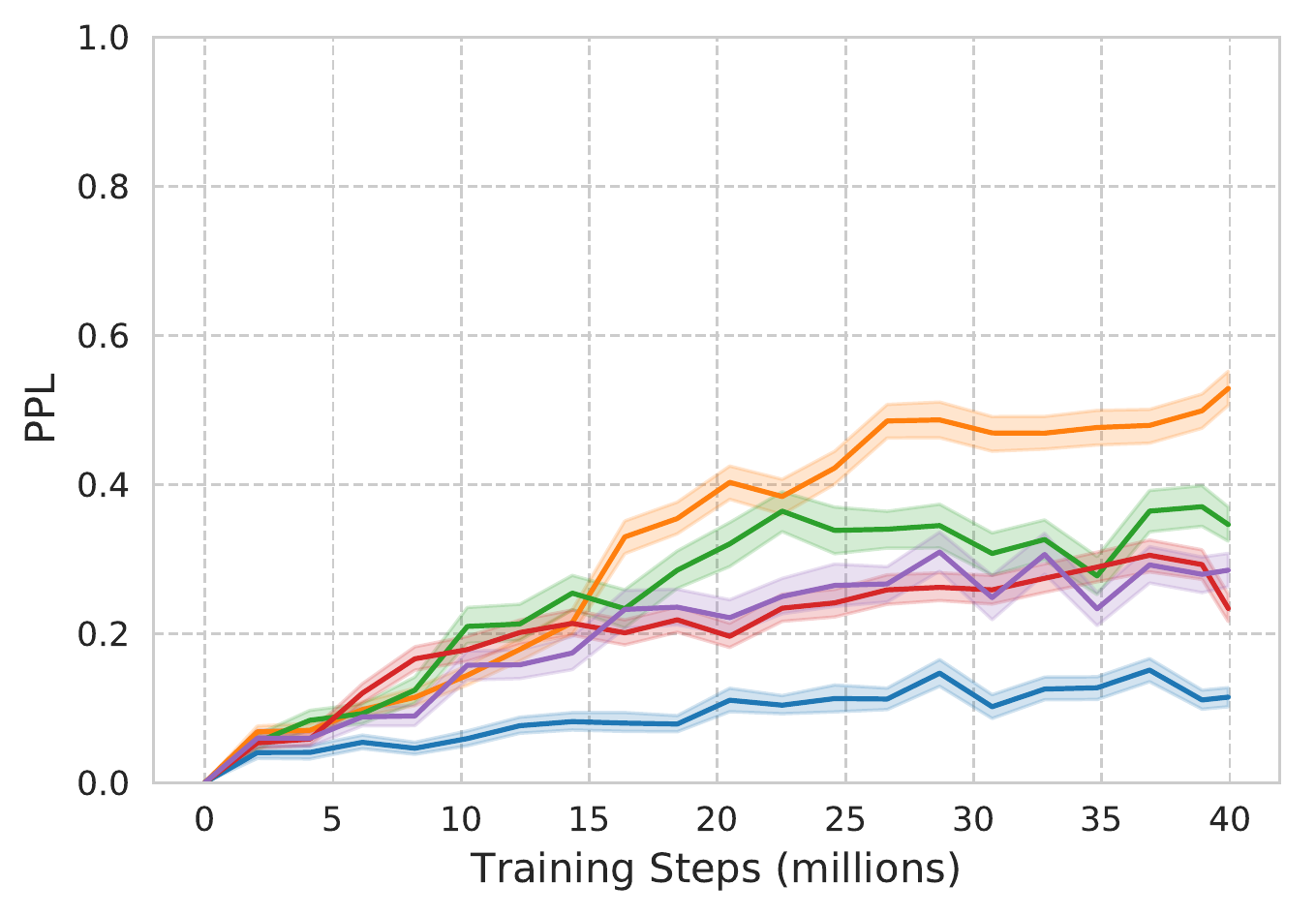}\\
\includegraphics[width=0.98\linewidth]{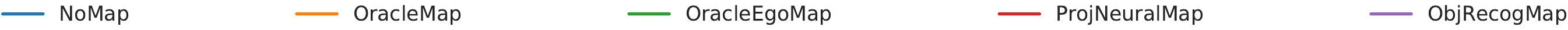}
\caption{\PPL of agents during training, evaluated on validation set with $95\%$ confidence intervals in shading. Overall performance decreases for all models as task complexity increases. \OracleMap trains the fastest and reaches the best overall performance. \OracleEgoMap and \ObjRecogMap follow closely, while the \ProjNeuralMap and \NoMap agents perform the worst.}
\label{fig:val-curves}
\end{figure}

\subsection{Example episodes}

\Cref{fig:episode-maps} shows example episodes visualizing the performance of different agent models on val set scenes.
\NoMap tends to wander and get low \PPL even when it is able to reach the goals.
When \found is called not in the vicinity of the current goal, the episode is terminated resulting in failure.
This happens in the top row for several agent models, as well as \ProjNeuralMap in the bottom row.

\subsection{Quantitative evaluation}

We train all agents for $40$ million steps on \mon{1}, \mon{2} and \mon{3} tasks.
During training we measure performance on $1@000$ validation scene episodes.
The maximum number of steps is $2@500$, which we found to be sufficiently large for all \mon{m} tasks in this paper.
\Cref{fig:val-curves} plots the \PPL metric for all agent models.
We pick the best performing checkpoint for each model and evaluate on $1@000$ episodes in $18$ unseen test scenes.
\Cref{tab:overall-metrics} summarizes all evaluation metrics averaged across test episodes (including for ablations of the \OracleMap agent using only occupancy or object category information).
As expected, the \OracleMap agents have the highest performance across the board, with significant gains over \NoMapAll indicating the value of explicit map information.
Of the \texttt{NoMap} variants, \FRMQN performed the best, surpassing the newer transformer-inspired model \SMT (potentially due to the need for more training).
Object category information in the map accounts for most of the gain.
The \OracleEgoMap agent follows, but \ObjRecogMap is surprisingly close, followed by \ProjNeuralMap.

\emph{How hard are the \task tasks?}
Even the oracle agents drop to $\leq50\%$ \PPL on \mon{3}.
Our \RandOracleFound agent is similar to \texttt{Random} in \citet{fang2019scene}.
At $500$ steps \citet{fang2019scene}'s \texttt{Random} finds $36.3\%$ of object classes, whereas \RandOracleFound finds $9\%$, $5\%$ and $3\%$ in \mon{1}, \mon{2}, \mon{3} respectively (note that \Cref{tab:overall-metrics} reports metrics at $2@500$ max steps).
This indicates the \mon{m} tasks are quite challenging.
In particular, objects need to be found in order, so we see a clear trend where \mon{m} is harder with more goals $m$.
The \task task is challenging and well-suited to investigating the usefulness of map information across a spectrum of difficulties.
The performance drop between \mon{1} and \mon{3} is dramatic.
We might have expected agents that remember past goal observations to have a higher rate of success when returning to a goal later.
In the supplement we carry out an analysis verifying that some agents can remember the location of objects seen during the course of the episode.
Despite this, we observe a worse than purely exponential drop in performance.
With exponential decay, we would expect \OracleMap \Success rates to be $0.94$, $0.94^2 = 0.88 $ and $0.94^3 = 0.83$ as $m$ is increased from $1$ to $3$.
The observed \Success rates are $0.94$, $0.79$ and $0.62$, indicating that there is room for improvement.

\emph{Are spatial maps useful?}
Yes. Having ground truth maps helps significantly (compare \OracleMap and \OracleEgoMap with \NoMap).
\OracleMap is almost perfect ($94\%$ \Progress/\Success) on \mon{1}, but does not always pick the shortest path (\SPL/\PPL of $77\%$).
However, performance drops for \mon{2} and \mon{3}.
Still, maps help more for \mon{2} and \mon{3} than for \mon{1} indicating the value of spatial memory for longer-horizon tasks.

\emph{What information in the spatial maps is useful?}
While it may be obvious that maps help, we see that object category information in the map indicating the goal location is much more useful than occupancy information (see drop from \OracleMap \texttt{(Obj)} to \OracleMap \texttt{(Occ)}).
Our hypothesis is that occupancy mainly helps to avoid collisions and this can be achieved using the depth sensor, since we have perfect depth and localization (i.e. no actuation or sensor noise).
Note that the \texttt{Obj} ablation is higher than \texttt{Occ+Obj}.
We suspect this is due to \texttt{Occ+Obj} being harder to train to leverage the added information channel, as \texttt{Occ+Obj} did not fully converge within $40M$ training steps.

\emph{What kind of learned map is useful?}
Surprisingly, \ObjRecogMap which simply learns to project predicted object categories into the map is quite competitive and outperforms the \ProjNeuralMap and \EgoMap agents for \mon{1} and \mon{2} based on the MapNet~\cite{henriques2018mapnet} projection architecture.
For \mon{3}, the \ProjNeuralMap and \EgoMap agents are able to outperform \ObjRecogMap.
Overall, the performance of \ProjNeuralMap and \EgoMap are very similar, suggesting limited contribution of using attention to read from the semantic map.
This indicates that simple recognition of the goal from visual features and direct integration into a map memory can be more effective than accumulating image features into a map.
Still, performance is not as good as \OracleEgoMap indicating that there is more work to be done on learned map representations to match oracle egocentric mapping.

\begin{table}
\ra{1.3}
\centering
\caption{Agent performance on \mon{1}, \mon{2} and \mon{3} test set (maximum $2@500$ steps). The \task task is challenging with \RandOracleFound achieving $26\%$ success (SPL $8\%$) for \mon{1}, and \Rand failing completely. Performance decreases for all agents as we add more objects. Overall, maps help considerably, with the ability to represent goal objects in the map being particularly valuable (compare \OracleMap \texttt{(Obj)} and \OracleMap \texttt{(Occ)} as well as \ObjRecogMap and \ProjNeuralMap).
}
\label{tab:overall-metrics}
\resizebox{\linewidth}{!}{
\begin{tabular}{@{}cl rrr rrr @{\hspace{7mm}} rrr rrr @{}}
\toprule
 & & \multicolumn{3}{c}{\Success (\%)} & \multicolumn{3}{c}{\Progress (\%)} & \multicolumn{3}{c}{\SPL (\%)} & \multicolumn{3}{c}{\PPL (\%)}\\ \cmidrule(lr){3-5} \cmidrule(lr{7mm}){6-8} \cmidrule(l{0mm}r){9-11} \cmidrule(lr){12-14}
 & & \mon{1} & \mon{2} & \mon{3} & \mon{1} & \mon{2} & \mon{3} & \mon{1} & \mon{2} & \mon{3} & \mon{1} & \mon{2} & \mon{3}\\
\midrule
&\Rand & $0$ & $0$ & $0$ & $0$ & $0$ & $0$ & $0$ & $0$ & $0$ & $0$ & $0$ & $0$\\
&\RandOracleFound & $26$ & $8$ & $2$ & $26$ & $16$ & $12$ & $8$ & $2$ & $1$ & $8$ & $5$ & $4$\\
\midrule
\multirow{3}{*}{\STAB{\rot{NoMap}}}
& \NoMap & $62$ & $24$ & $10$ & $62$ & $39$ & $24$ & $35$ & $13$ & $4$ & $35$ & $21$ & $14$\\
& \FRMQN \cite{oh2016control}  & $62$ & $29$ & $13$ & $62$ & $42$ & $29$ & $50$ & $24$ & $11$ & $50$ & $33$ & $24$\\
& \SMT \cite{fang2019scene} & $63$ & $28$ & $9$ & $63$ & $44$ & $22$ & $48$ & $26$ & $7$ & $48$ & $36$ & $18$\\
\midrule
\multirow{4}{*}{\STAB{\rot{Oracle}}}
& \OracleMap (Occ+Obj) & $94$ & $74$ & $48$ & $94$ & $79$ & $62$ & $77$ & $59$ & $38$ & $77$ & $63$ & $49$\\
& \OracleMap (Occ) & $66$ & $34$ & $16$ & $66$ & $47$ & $36$ & $48$ & $25$ & $12$ & $48$ & $35$ & $27$\\
& \OracleMap (Obj) & $94$ & $82$ & $59$ & $94$ & $86$ & $70$ & $79$ & $65$ & $42$ & $79$ & $67$ & $50$\\
& \OracleEgoMap (Occ+Obj) & $83$ & $64$ & $37$ & $83$ & $71$ & $54$ & $65$ & $49$ & $25$ & $65$ & $54$ & $36$\\
\midrule
\multirow{3}{*}{\STAB{\rot{Learned}}}
& \EgoMap~\cite{beeching2020egomap} & $69$ & $46$ & $26$ & $69$ & $59$ & $44$ & $49$ & $31$ & $\mathbf{18}$ & $49$ & $42$ & $30$\\
& \ProjNeuralMap & $70$ & $45$ & $\mathbf{27}$ & $70$ & $57$ & $\mathbf{46}$ & $51$ & $30$ & $\mathbf{18}$ & $51$ & $39$ & $\mathbf{31}$\\
& \ObjRecogMap & $\mathbf{79}$ & $\mathbf{51}$ & $22$ & $\mathbf{79}$ & $\mathbf{62}$ & $40$ & $\mathbf{56}$ & $\mathbf{38}$ & $17$ & $\mathbf{56}$ & $\mathbf{45}$ & $30$\\
\bottomrule
\end{tabular}
}
\end{table}

\emph{What's next?}
Even with \OracleMap success drops to only $59\%$ ($42\%$ \SPL) for \mon{3}, despite the apparent triviality of navigating to three distinct objects.
A key question for future work is how to design architectures that avoid a dramatic performance drop with increasing numbers of goal objects, which is a clear sign of non-robustness in embodied agents.
The drop between \OracleMap and \OracleEgoMap is likely attributed to partial observability of the environment under egocentric embodiment constraints and offers a more conservative `high bar' for future work.
The subsequent drop from \OracleEgoMap to both \ProjNeuralMap and \ObjRecogMap indicates that further study of approaches for aggregating egocentric information and integrating it into a map would be fruitful.

In this paper, we investigated a representative set of prior work on agent map representations and compared their performance on the MultiON task.
We have assumed perfect localization in our experiments, which is usually not a practical assumption.
An interesting direction would be to explore how neural inspired grid-cell architectures such as in \citet{banino2018vector} that predict the location and heading can be used to localize the agent.
We have also focused on the use of grid-based maps.
Other types of maps such as topological maps would be interesting to investigate and contrast.

\section{Conclusion}

We introduced \task, a task framework allowing for systematic analysis of embodied AI navigation agents utilizing semantic map representations.
Our experiments with several agent models show that semantic maps are indeed highly useful for navigation, with a relatively na\"ive integration of semantic information into map memory providing high gains against more complex learned map representations.
However, overall agent performance degrades dramatically with modest task difficulty increases even for the best performing agents.
Our experiments suggest several promising directions for future work, including improved learned modules for integrating egocentric information into map representations.
We hope that the \task framework provides a flexible benchmark for systematic study of spatial memory and mapping mechanisms in embodied navigation agents.
\clearpage
\section*{Acknowledgments}

We thank the anonymous reviewers for their helpful suggestions.
Unnat Jain thanks Alexander Schwing and Svetlana Lazebnik for their support.
Angel X. Chang is supported by the Canada CIFAR AI Chair program.
Manolis Savva is supported by an NSERC Discovery Grant.
This research was enabled in part by support provided by WestGrid (\url{https://www.westgrid.ca/}) and Compute Canada (\url{www.computecanada.ca}).
\section*{Broader Impact}

This work is a step toward enabling robots that can navigate and operate in real world environments.
We focus on the study of what kind of maps might be useful to such robotic agents, and how to benchmark their performance.
In the future, robust robotic agents may be able to assist the elderly (bringing them their medication), deliver things to hotel rooms, help in moving items in warehouses, and generally serve broader society in a variety of roles.
On the other hand, success along these directions may cause displacement of workers from related occupations and economic difficulties for large segments of the population currently employed in a variety of sectors that are amenable to automation.
Moreover, deployment of imperfect robotic agents, that are not guaranteed to be failure-free may cause injuries or damages.
We believe that developing and studying the behavior of such systems in simulation first, and then in controlled real environments is paramount for minimizing these risks, before they are deployed in the real world.

\clearpage
{\small
\bibliographystyle{abbrvnat}
\bibliography{unnat,semantic}
}

\clearpage

\appendix
\section{Supplementary material}

In this supplemental document, we first provide a summary of the notation used in the main paper and here (\Cref{sec:notation}).
Then, we describe the implementation details of agent models used for our experiments (\Cref{sec:details}).
In \Cref{sec:statistics} we provide statistics for the episodes used in our experiments.
Finally, \Cref{sec:additional} contains additional experiments and analysis.

\subsection{Notation}
\label{sec:notation}

\Cref{tab:notation} provides a summary of definitions for important symbols used in the main paper and in this supplemental document.
 
\begin{table}[h]
\ra{1.3}
\centering
\caption{Summary of notation used. Subscript $t$ denotes the corresponding notation at time step $t$}
\label{tab:notation}
\resizebox{\linewidth}{!}{
\begin{tabular}{H{2cm}H{6cm}H{2cm}H{6cm}}
\toprule
Notation & Description & Notation & Description\\
\midrule
\mon{m} & Episode with $m$ ordered object goals & $v_t$ & $\concat(v_i, v_m, v_g, v_a)$\\
$p_t$ & Agent's position and orientation & $s_t$ & Final state representation\\
$o_t$ & Egocentric RGBD sensor images & $\theta$ & Parameters of end-to-end trainable model\\
$c_t$ & Egocentric RGB sensor image & $\pi_\theta(\cdot|s_t)$ & Actor policy given state $s_t$ \\
$d_t$ & Egocentric Depth sensor image & $V(s_t)$ & Approximate value function\\
$g_t$ & Current goal object one-hot vector & $M_{GT}$ & Oracle Map\\
$a_t$ & Action taken by the agent & $r_t$ & Reward\\
$m_t$ & Egocentric Map & $r_\text{\tiny subgoal}$ & Subgoal discovery reward\\
$M_t$ & Global Map & $r_\text{\tiny closer}$ & Moving closer to subgoal reward\\
$i_t$ & Image Features & $r_\text{\tiny time-penalty}$ & Time penalty reward\\
$v_i$ & Transformed Image features after passing image through a CNN and a linear layer & $\alpha$ & Negative slack reward\\
$v_i$ & Embedding of $m_t$ after passing it through a CNN and a linear layer & $d_{i-1, i}$ & Geodesic distance of the shortest path between goal $i-1$ and $i$\\
$v_g$ & Embedding of one-hot goal vector $g_t$  & $s$ & Binary success indicator\\
$v_a$ & Embedding of previous action $a_{t-1}$& $\bar{s}$ & \Progress\\
\bottomrule
\end{tabular}
}
\end{table}

\subsection{Agent model details}
\label{sec:details}

We describe the details of the \OracleEgoMap, \ObjRecogMap, and \ProjNeuralMap agent models.
For each map-based model, the agent has access at time $t$ to a global map $M_t$ which is a $300 \times 300$ grid, with each cell corresponding to a  $0.8\text{m} \times 0.8\text{m}$ square in the environment.
Depending on the model, $M_t$ is revealed or built up over time.
The details of what goes into each map cell and how the information is built up over time are described below.

\xhdr{\OracleEgoMap}:
At each time step, the agent has access to a partially revealed map $M_t$ that is derived from the oracle map.
Each cell has two channels: i) an occupancy channel; and ii) an object category channel.
The occupancy channel stores a $16$-dimensional embedding learned from whether the cell is navigable, non-navigable or \textit{undiscovered}.
The object category channel stores a $16$-dimensional embedding of the the category of the object occupying that cell learned from a $k+1$ dimensional one-hot vector ($k$ dimensions for the goal categories and $(k+1)^{th}$ dimension for `no goal'.)
At the start of the episode, all the cells in the occupancy channel are `undiscovered'.
A cell is \emph{discovered} once the agent sees (the region of the environment corresponding to) it and it remains discovered through the rest of the episode.
At each time step, the agent sees only those locations in the environment that are: i) within a $79^{\circ}$ field-of-view in front of it; ii) less than $5\text{m}$ away from the agent; and iii) with no obstacle between the agent and that location.
All cells in the object category channel are initialized with the $(k+1)^{th}$ category embedding.
When an object is discovered, its category embedding is stored in the corresponding cell in the object category channel.

\xhdr{\ObjRecogMap}:
The map $M_t$ used in \ObjRecogMap is similar to that of \OracleEgoMap but instead of it being `revealed' from the oracle map, it is predicted based on egocentric views.
In addition, each cell has only the object category channel.
As in \OracleEgoMap, each cell stores one of the $k+1$ 16-dimensional embeddings ($k$ embeddings for the goal categories and $(k+1)^{th}$ for `no goal'.)

At each time step, the current RGB frame is passed through a $(k+1)$—classification network to predict the category of the goal in view (if no goal is in view, $(k+1)$ is used as the prediction target).
At training time, the network is supervised using the ground truth object category information to predict which goal object is in view.
An object is considered to be in the agent's view if it falls in a grid cell that satisfies the three conditions described in above for \OracleEgoMap.
We use categorical cross-entropy loss to supervise the network.
If more than one object is in the agent's current view, the agent is trained to predict the one that is closer to it.
Initially, all the cells in the map store the $(k+1)^{th}$ category embedding (corresponding to `no goal'.) If the agent predicts $(k+1)^{th}$ category at a time step, the map is not updated. If the agent predicts a category $l \in \{1, \dots, k\}$, it stores the encoding for the $l^{th}$ category at the cell containing its current position.

\xhdr{\ProjNeuralMap}:
As before, the agent has access to a partially built map $M_t$.
The agent builds up an egocentric map $m_t$ of size $7 \times 13$.
Note that this covers an area of $5.6m$ in front and to both sides of the agent.
Points that are farther than $5.6m$ from the agent are not projected.
Here also, the agent has a $79^{\circ}$ field-of-view.
We project the image feature $i_t(i, j, \cdot)$ on the ground using the depth frame $d_t$.
To do this, we downsample $d_t$ so it matches the image feature dimensions and use the downsampled depth for projection.
We observed that the alternative of upsampling and interpolating image features leads to reduced performance.
The agent is at the mid bottom edge of this egocentric map. $\mathsf{R}(m_t, M_t|p_t)$ registers the egocentric map $m_t$ into $M_t$.
The integration into $M_t$ uses element-wise max-pooling.%

\subsection{Episode statistics}
\label{sec:statistics}

\begin{figure}[ht]
\begin{tabular}{C{0.1cm}C{4cm}C{4cm}C{4cm}}
& \mon{1} & \mon{2} & \mon{3} \\
\begin{turn}{90}Train\end{turn} & \includegraphics[width=\linewidth]{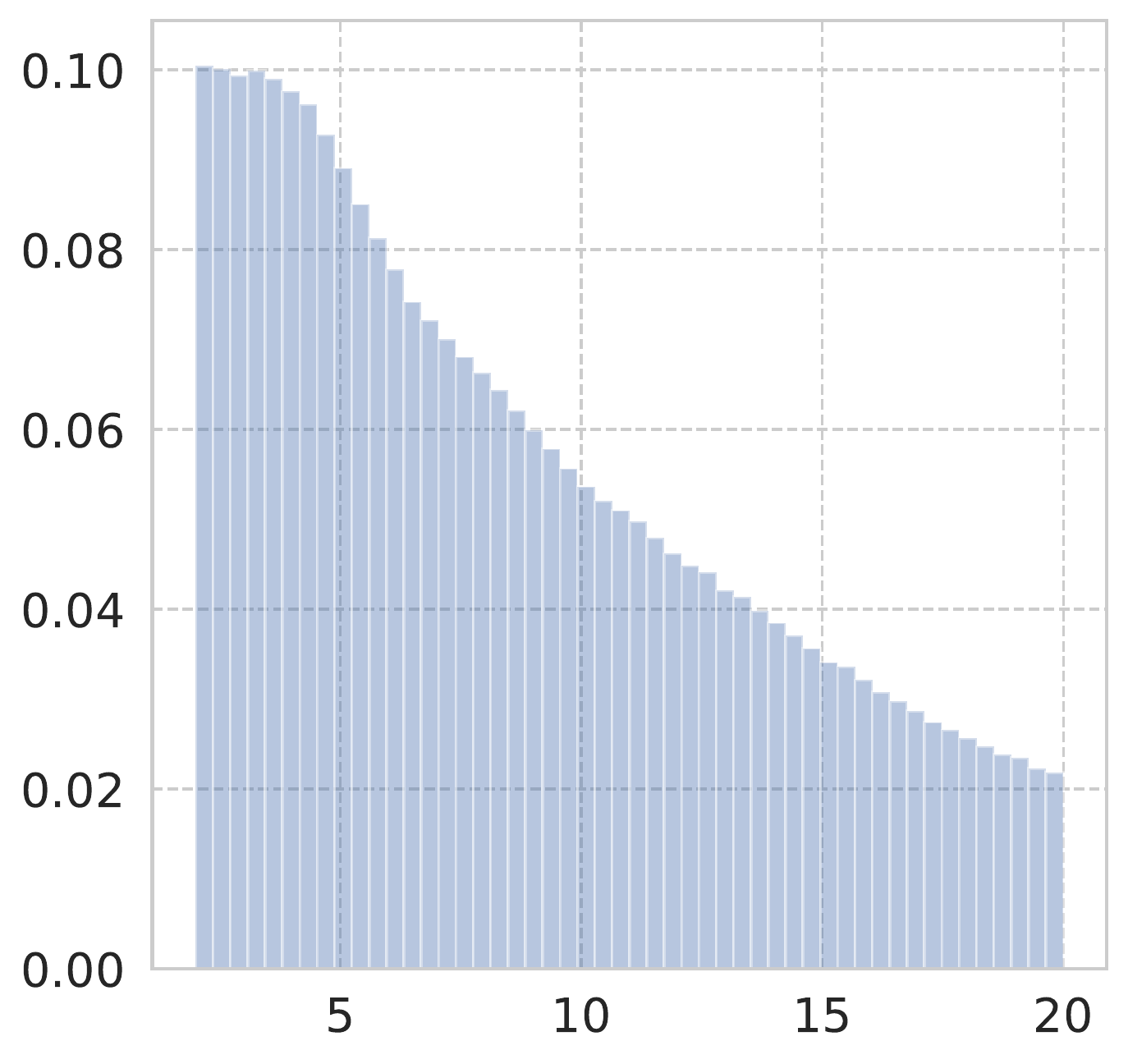} &
\includegraphics[width=\linewidth]{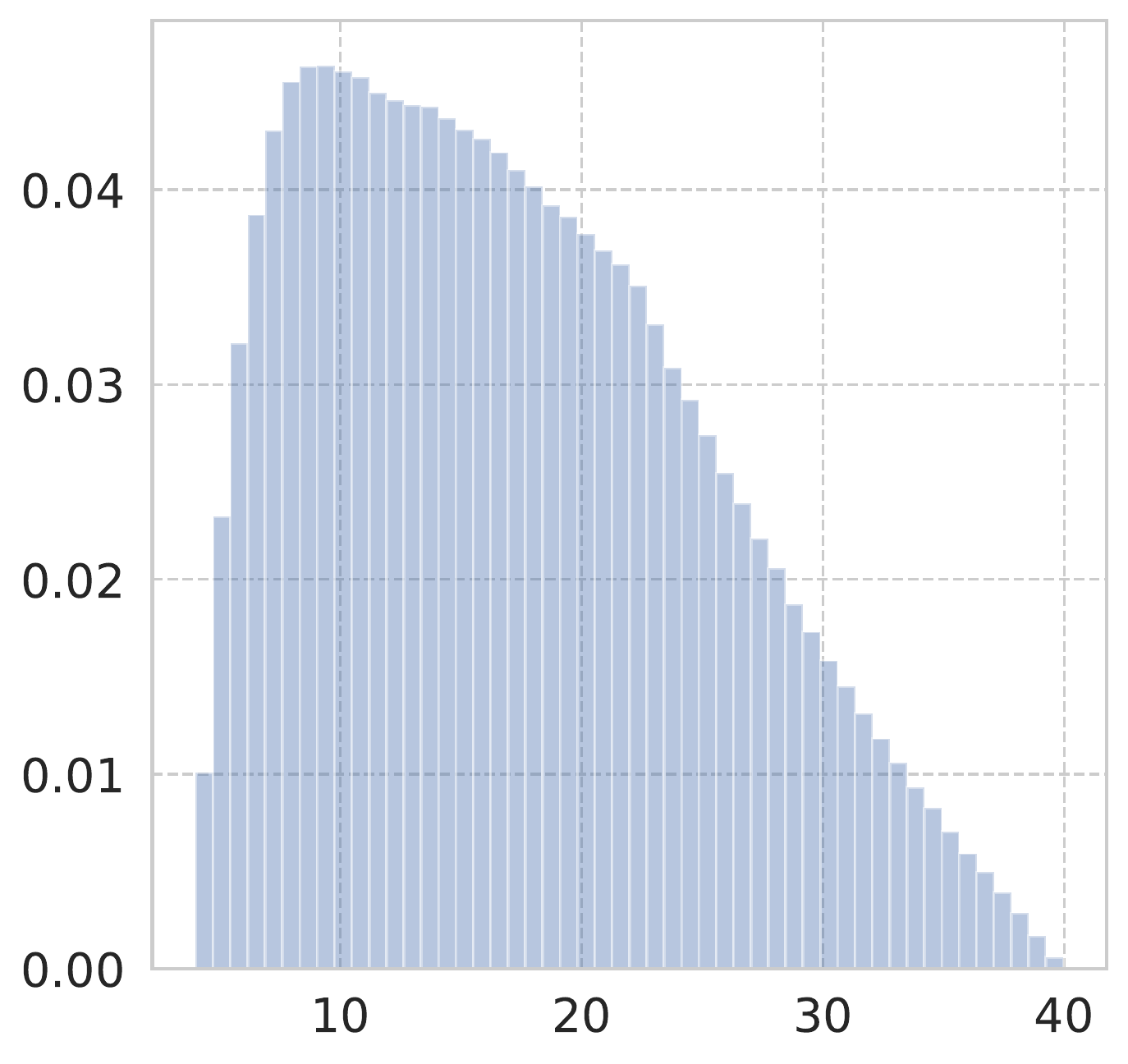} &
\includegraphics[width=\linewidth]{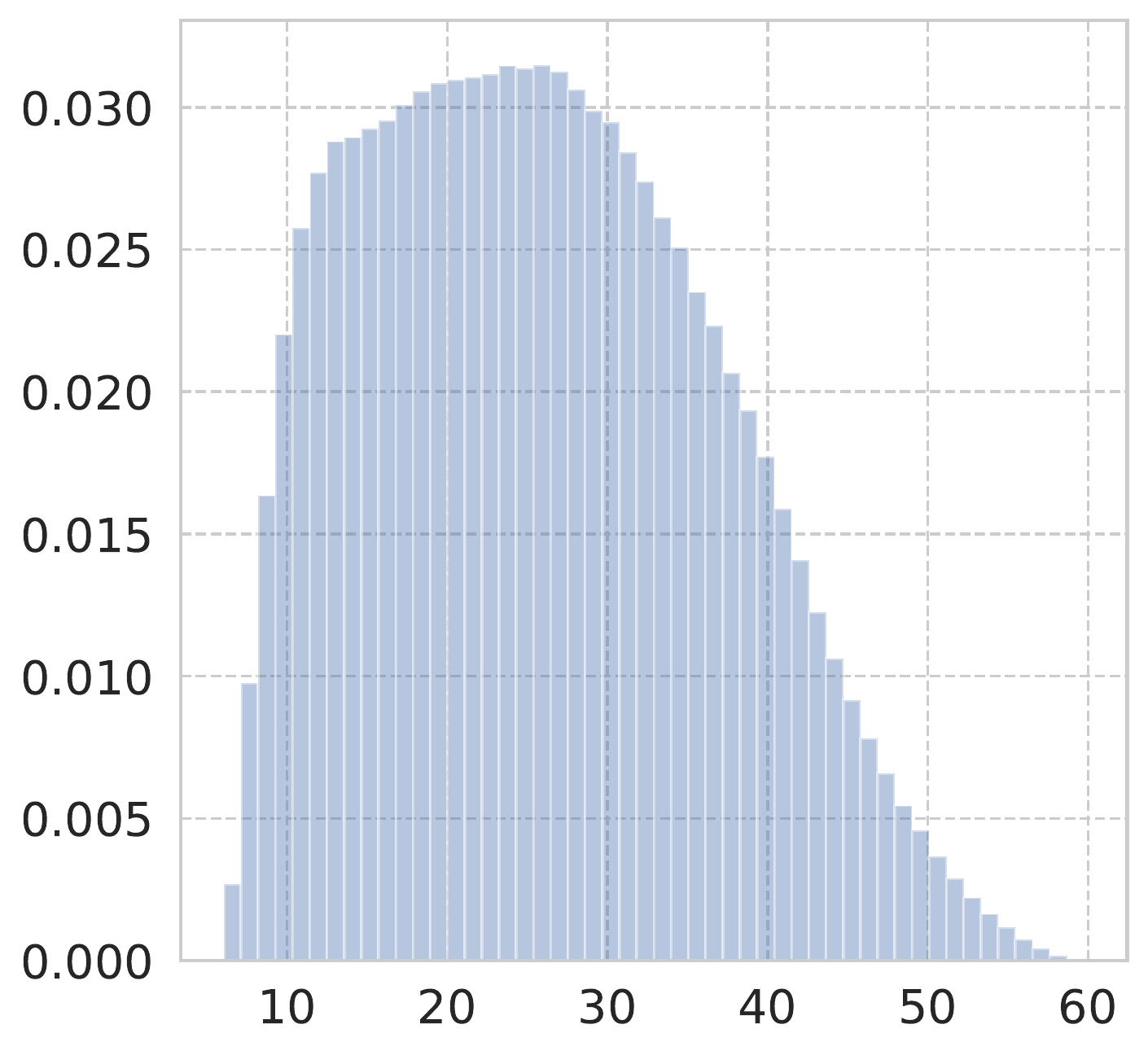}\\
\begin{turn}{90}Val\end{turn} & \includegraphics[width=\linewidth]{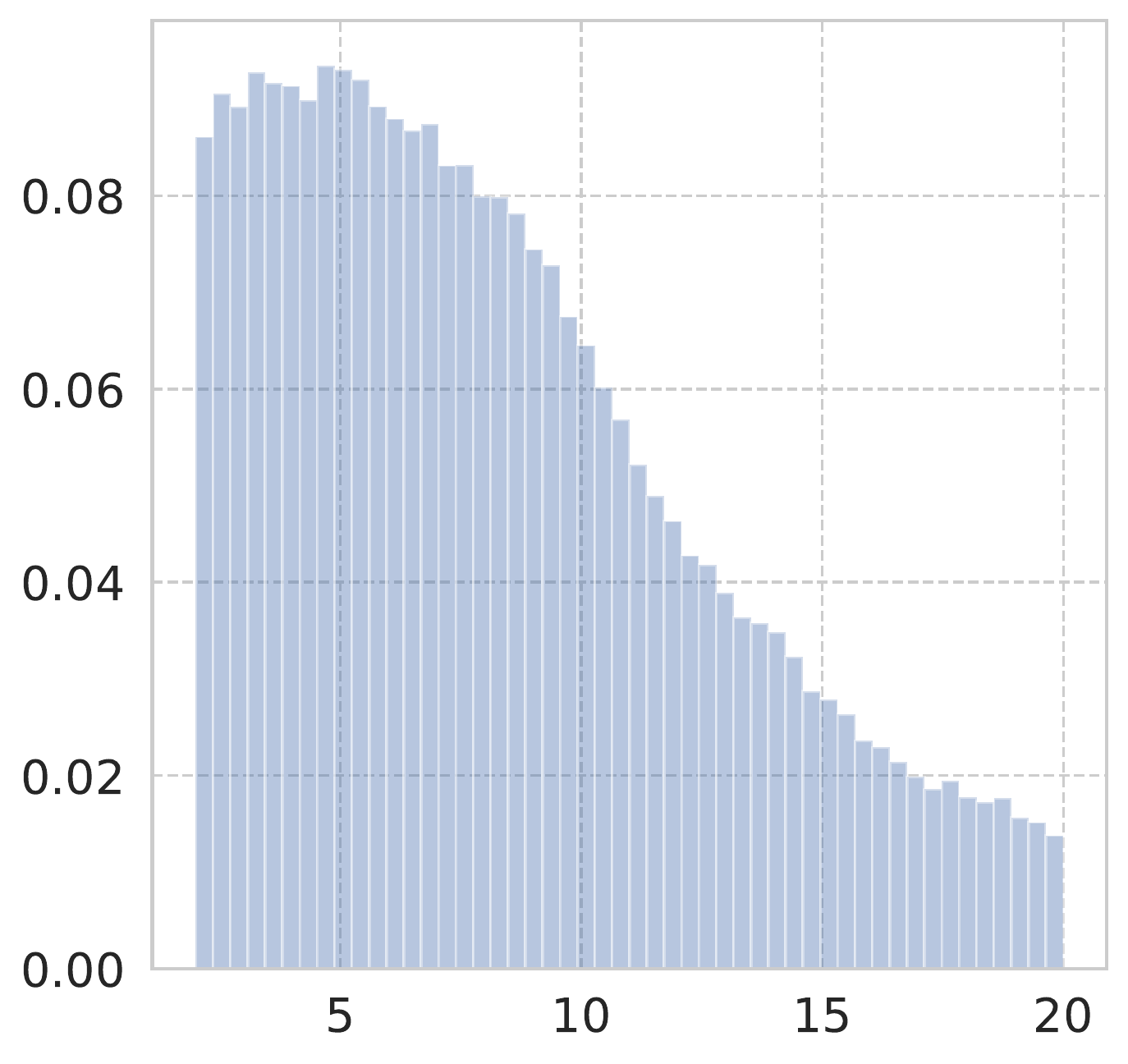} &
\includegraphics[width=\linewidth]{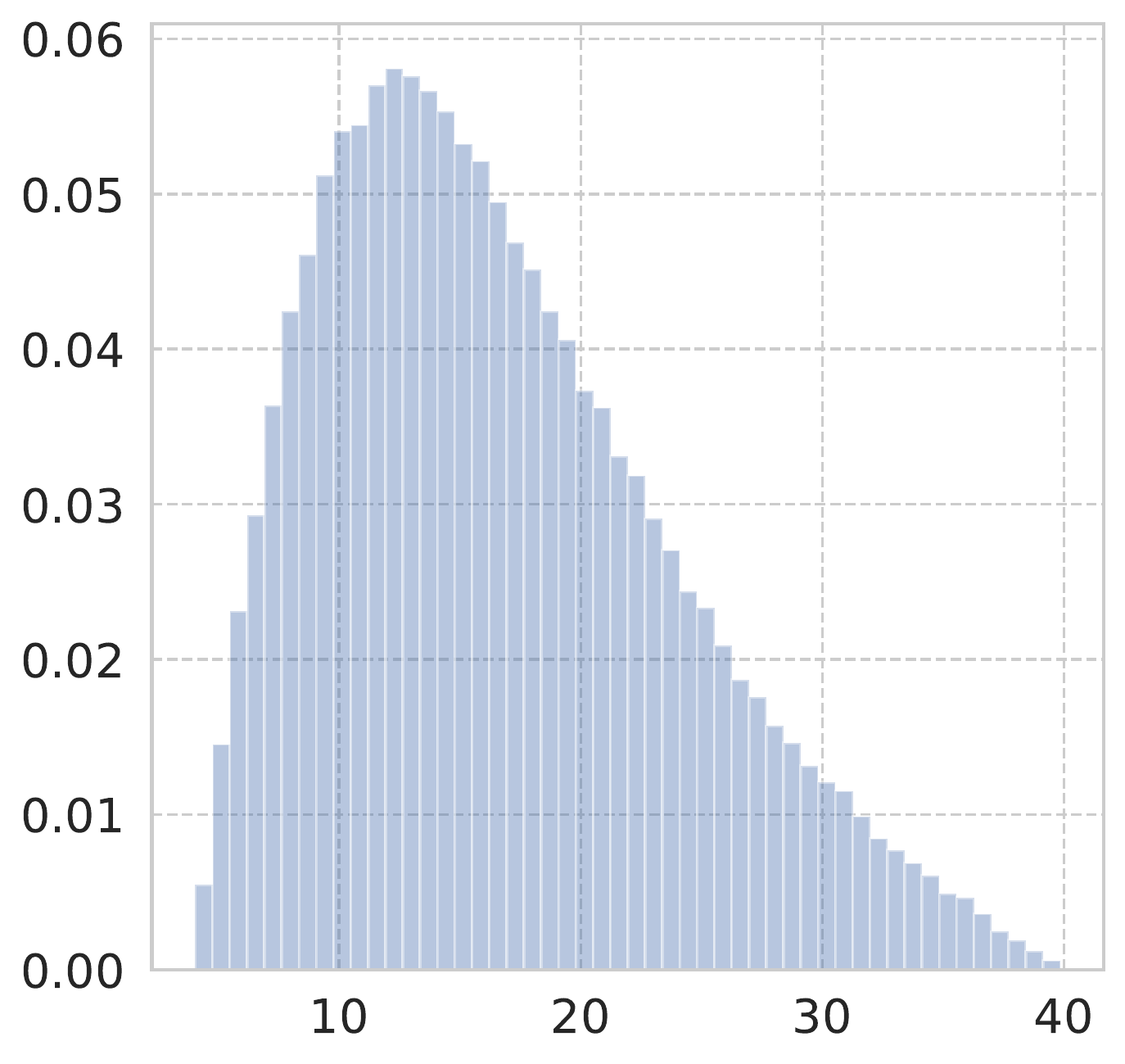} &
\includegraphics[width=\linewidth]{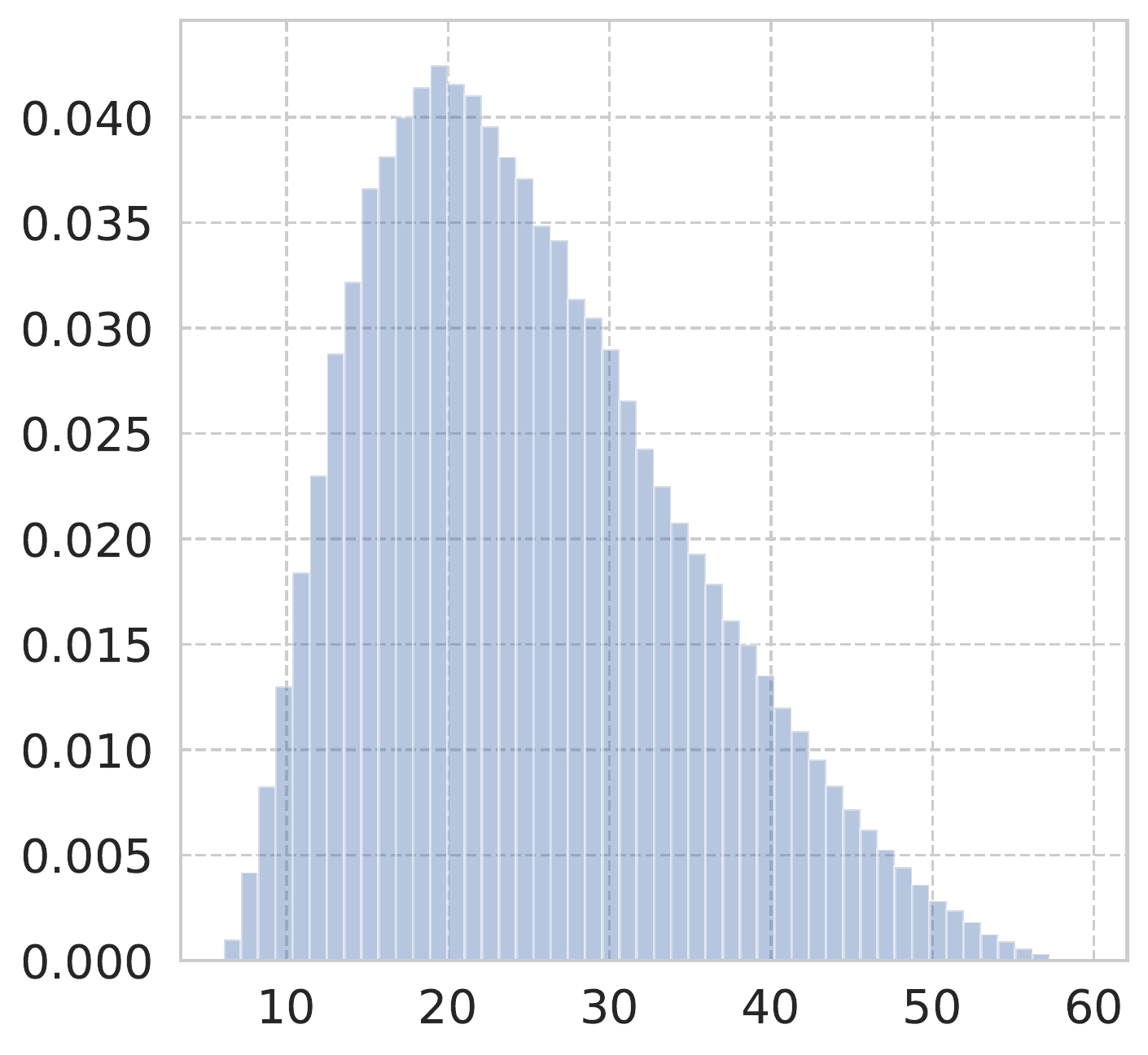}\\
\begin{turn}{90}Test\end{turn}  & \includegraphics[width=\linewidth]{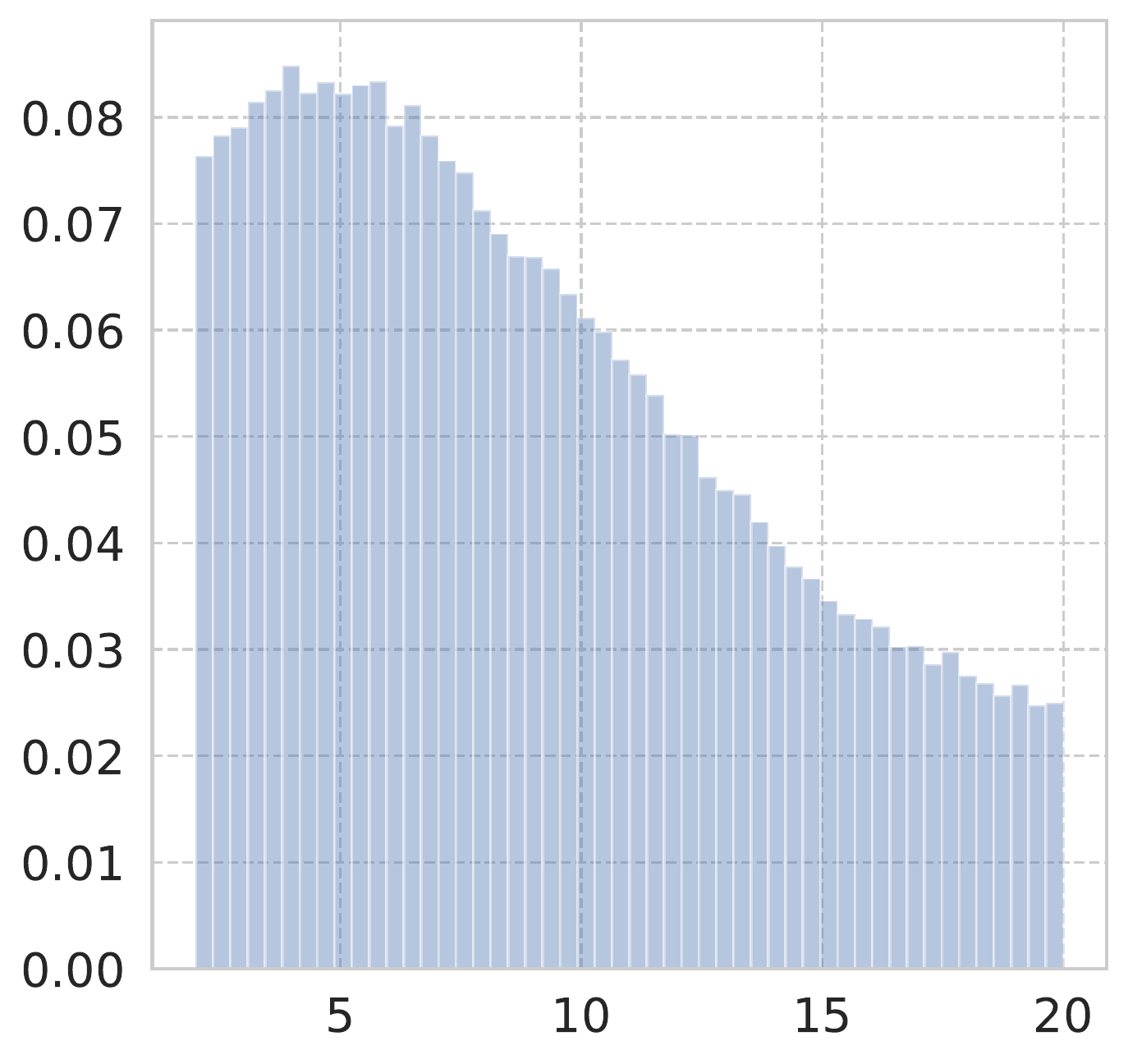} &
\includegraphics[width=\linewidth]{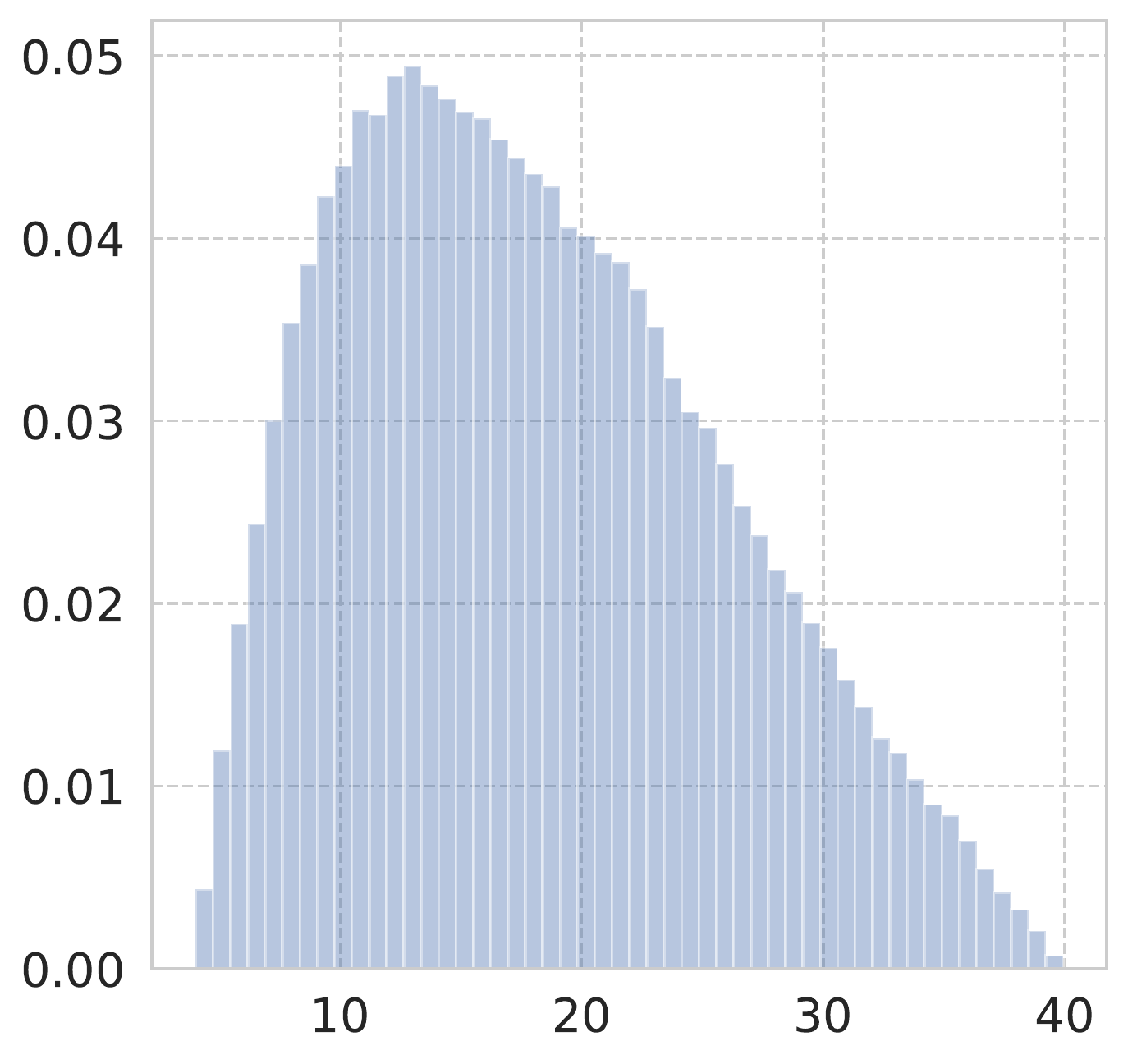} &
\includegraphics[width=\linewidth]{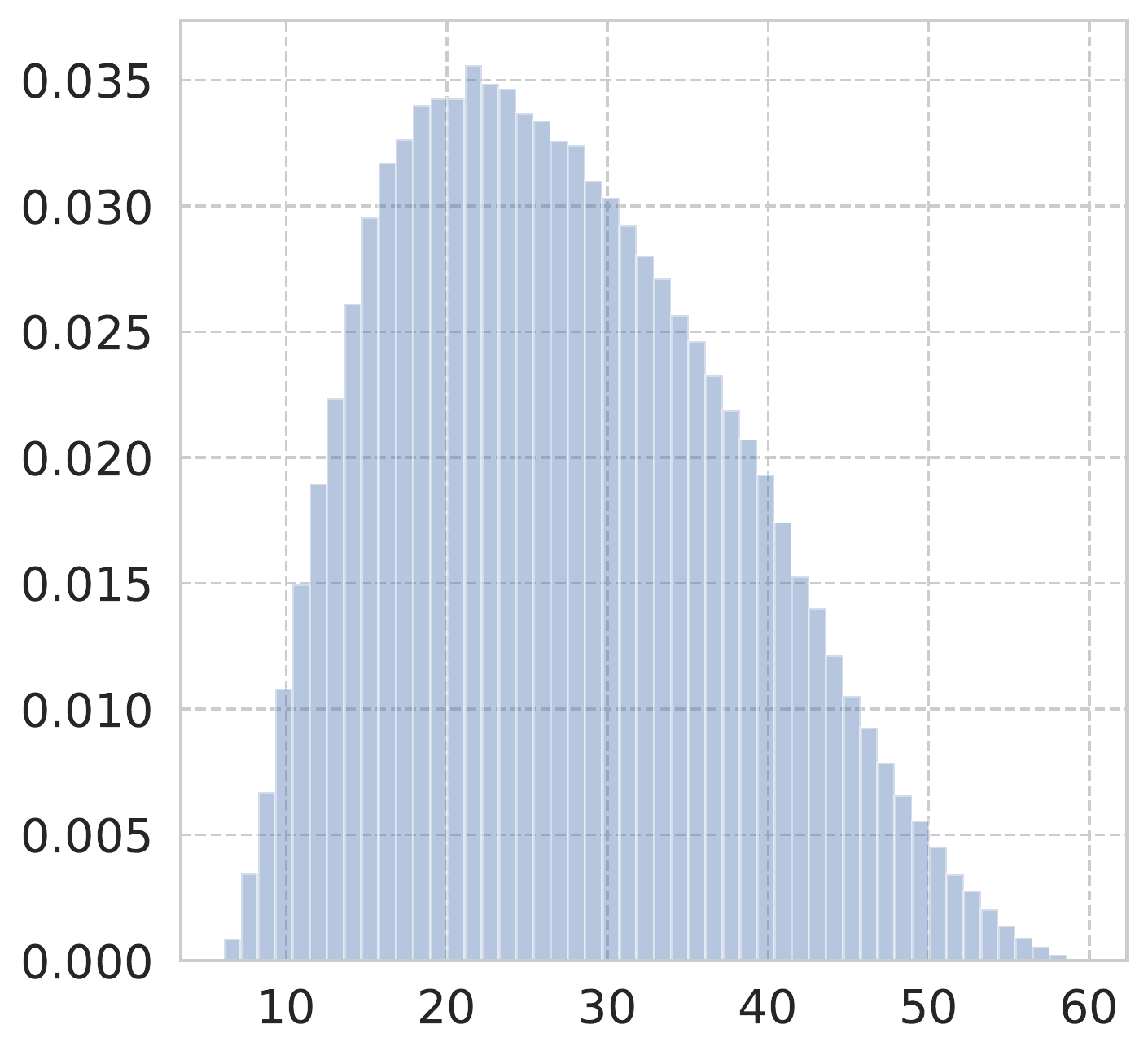}\\
\end{tabular}
\caption{Plots showing the distribution of train, val and test split episodes over the total geodesic distance for the oracle shortest path of the episode. The horizontal axis denotes the geodesic distance and the vertical axis denotes the fraction of total episodes in the corresponding histogram bin. The first row corresponds to the train split, the second row corresponds to the val split, and the third row corresponds to the test split. Each scene has $50@000$ episodes in the train split and $12@500$ episodes in the val and the test splits.}
\label{fig:dataset-stats}
\end{figure}

\Cref{fig:dataset-stats} plots the distribution of episodes from the train/val/test splits across total episode geodesic distance along the oracle shortest path.
Note that overall, the distribution tends to longer episodes as we go from \mon{1} to \mon{2} and \mon{3}.
There are small variations between the train/val/test splits for each \mon{m}, but the distributions are generally in correspondence.
\clearpage
\newpage
\subsection{Additional experiments and analysis}
\label{sec:additional}

\begin{table}[ht]
\ra{1.3}
\centering
\caption{Success rate ($\%$) for finding the $k^{th}$ subgoal in the \mon{3} task. The `seen' column has the accuracy with which the $k^{th}$ subgoal is found if it was \textit{seen} before $(k-1)^{th}$ subgoal was discovered. In contrast the `not seen' column represents episodes where the goal was not previously seen. The `improvement' column reports the difference. As expected the \NoMap and \OracleMap do not exhibit improvements. In contrast, storing previously observed goal information helps \OracleEgoMap, \ProjNeuralMap and \ObjRecogMap significantly, with \ObjRecogMap showing the largest gains.}
\label{tab:seen-vs-unseen}
\resizebox{0.7\linewidth}{!}{
\begin{tabular}{@{}l lll lll @{}}
\toprule
 & \multicolumn{3}{c}{Second goal ($k=2$)} & \multicolumn{3}{c}{Third goal ($k=3$)} 
\\ \cmidrule(lr){2-4} \cmidrule(lr){5-7} 
 & seen & not seen & improvement & seen & not seen & improvement \\
\midrule
\NoMap & $53$ & $51$ & $+2$ & $47$ & $46$ & $+1$\\
\OracleMap & $\mathbf{82}$ & $\mathbf{80}$ & $+2$ & $\mathbf{80}$ & $\mathbf{79}$ & $+1$\\
\OracleEgoMap & $80$ & $66$ & $+14$ & $76$ & $59$ & $+17$\\
\ProjNeuralMap & $74$ & $54$ & $\mathbf{+20}$ & $68$ & $44$ & $+24$\\
\ObjRecogMap & $72$ & $53$ & $+19$ & $69$ & $44$ & $\mathbf{+25}$\\
\bottomrule
\end{tabular}
}
\end{table}

\xhdr{Backtracking analysis.}
We compare the success rate for finding the $k^{th}$ ($k=2$, $k=3$) goal in two types of scenarios: i) the $k^{th}$ goal was \textit{seen} before the $(k-1)^{th}$ subgoal was \found; and ii) the $k^{th}$ subgoal was \textit{not seen} before the $(k-1)^{th}$ subgoal was \found.
The results for \mon{3} episodes are summarized in \Cref{tab:seen-vs-unseen}.
A goal object is \textit{seen} if it satisfies the three conditions mentioned in \Cref{sec:details}.
This analysis quantifies the ability of the agent to \textit{remember} the location of an object that was previously seen.
If an agent sees the second goal while it is looking for the first goal, it would benefit from this information at the time of searching for the second goal.
Understandably, this has little effect for the \NoMap agent since it has no way of storing the location of previously seen goals, beyond implicit encoding in the GRU module of the agent architecture.
\OracleMap too does not benefit much since it already has the full oracle map of the environment.
On the other hand, \OracleEgoMap and \ObjRecogMap exhibit a significant jump in success rate when the goal was previously observed. Notably, \ProjNeuralMap, which stores information about what was observed but does not explicitly convert it to goal category, also shows a small gain for previously observed goals but less than that of \OracleEgoMap and \ObjRecogMap. 
These results further demonstrate the value of map memory and the importance of the kind of information stored in the map for the \task task.

\begin{table}
\ra{1.3}
\centering
\caption{
Evaluation of agents trained on \mon{m} (in rows) test episodes of \mon{n} (in columns).
We report the \Progress metric averaged across all test set episodes.
Off the diagonal for each agent shows generalization capability to \task episodes with a different number of objects than in training.
The top row triplet reports absolute values, while the bottom one reports change in \Progress relative to the diagonal (evaluation on trained task).
As expected, performance drops for all agents in all generalization tests, with the drop being greater for testing on episodes with more objects.
Agents trained on \mon{1} have the lowest generalization capability, while \mon{2} and \mon{3} agents exhibit smaller but still significant drops in performance.}
\label{tab:generalization}
\resizebox{\linewidth}{!}{
\begin{tabular}{@{}l ccc ccc ccc ccc ccc @{}}
\toprule
 & \multicolumn{3}{c}{\NoMap} & \multicolumn{3}{c}{\OracleMap} & \multicolumn{3}{c}{\OracleEgoMap} & \multicolumn{3}{c}{\ProjNeuralMap} & \multicolumn{3}{c}{\ObjRecogMap}  
 \\ \cmidrule(lr){2-4} \cmidrule(lr){5-7} \cmidrule(lr){8-10} \cmidrule(lr){11-13} \cmidrule(lr){14-16} 
 & \mon{1} & \mon{2} & \mon{3}  & \mon{1} & \mon{2} & \mon{3}  & \mon{1} & \mon{2} & \mon{3}  & \mon{1} & \mon{2} & \mon{3}  & \mon{1} & \mon{2} & \mon{3} \\
\midrule
\mon{1}  & 62 & 23 & 11  & 
94 & 27 & 12  &
 83 & 21  & 9   &
 65 & 25 & 12  &
 79 & 21 & 12     \\
 \mon{2}  & 56 & 39 & 22  & 
 89 & 79 & 46  &
8 & 71 & 43  &
71 & 57 & 44  &
 77 & 62 & 38  \\
 \mon{3}  & 53 & 38 & 24  &
 84 & 76 & 62  &
77 & 7 & 54  &
 67 & 56 & 46  &
 64 & 55 & 40  \\
\midrule
\mon{1}  & 0 & -16 & -13 &  
0 & -52 & -5 &  
0 & -5 & -54 &  
0 & -32 & -32 & 
0 & -41 & -28   \\
\mon{2}  & -6 & 0 & -2 &
 -5 & 0 & -16 &
-3 & 0 & -11 &  
6 & 0 & -2 &  
-2 & 0 & -2   \\
\mon{3}  & -9 & -1 & 0 &  
 -1 & -3 & 0 &  
-6 & -1 & 0 &  
 2 & -1 & 0 &  
-15 & -7 & 0 \\
\bottomrule
\end{tabular}
}
\end{table}

\xhdr{Testing agent generalization.}
We report task generalization performance in \Cref{tab:generalization} where agent models trained on an \mon{k} task are evaluated on \mon{l} task where $k \neq l$.
Models trained on \mon{1} fail to generalize to more complex \task tasks for both agents with and without map memory.
This is likely caused by no prior training for `continuing' the task past the first found goal (i.e. the agents have not been trained to withhold calling \found until they navigate to another goal after the first one).
In contrast, the agents trained on \mon{2} and \mon{3} are able to generalize to some extent, especially for the case of going from a more complex task to a less complex one.
This is exhibited by smaller performance drops in the `lower triangle' than the `upper triangle' of the generalization table.

\begin{figure}
\begin{tabular}{C{0.1cm}C{4cm}C{4cm}C{4cm}}
& \mon{1} & \mon{2} & \mon{3} \\
\begin{turn}{90}\NoMap\end{turn} & \includegraphics[width=\linewidth]{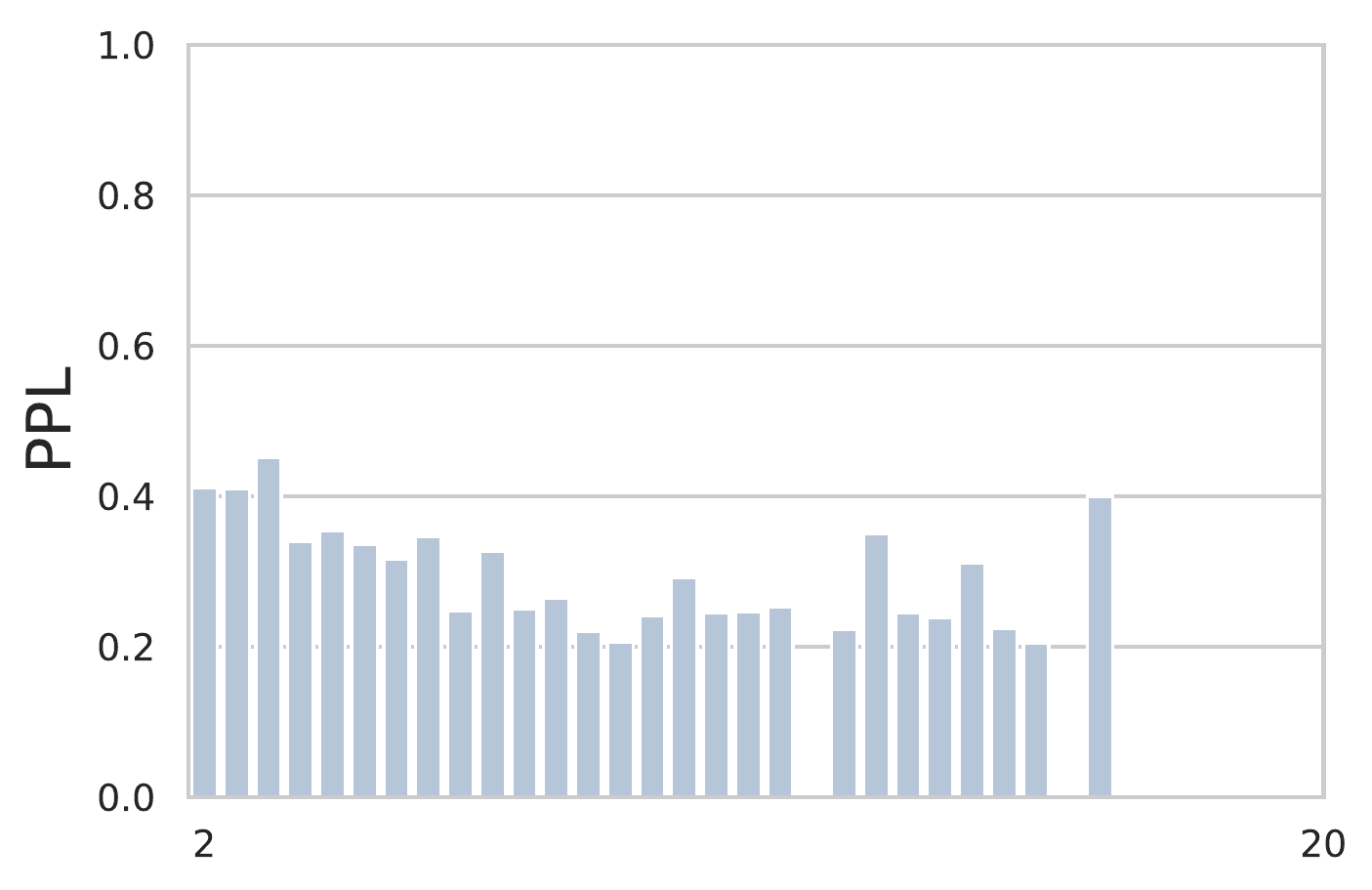} &
\includegraphics[width=\linewidth]{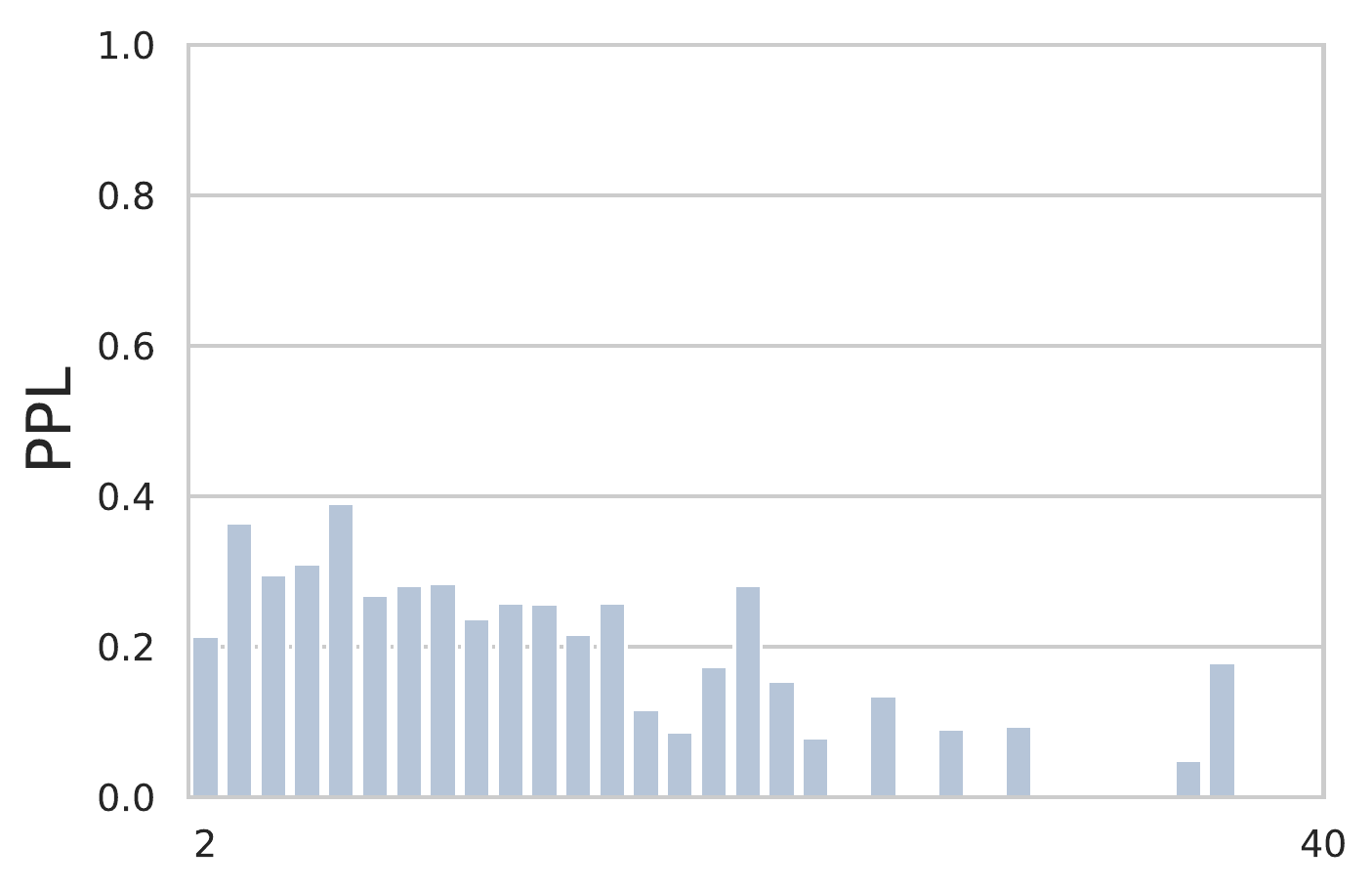} &
\includegraphics[width=\linewidth]{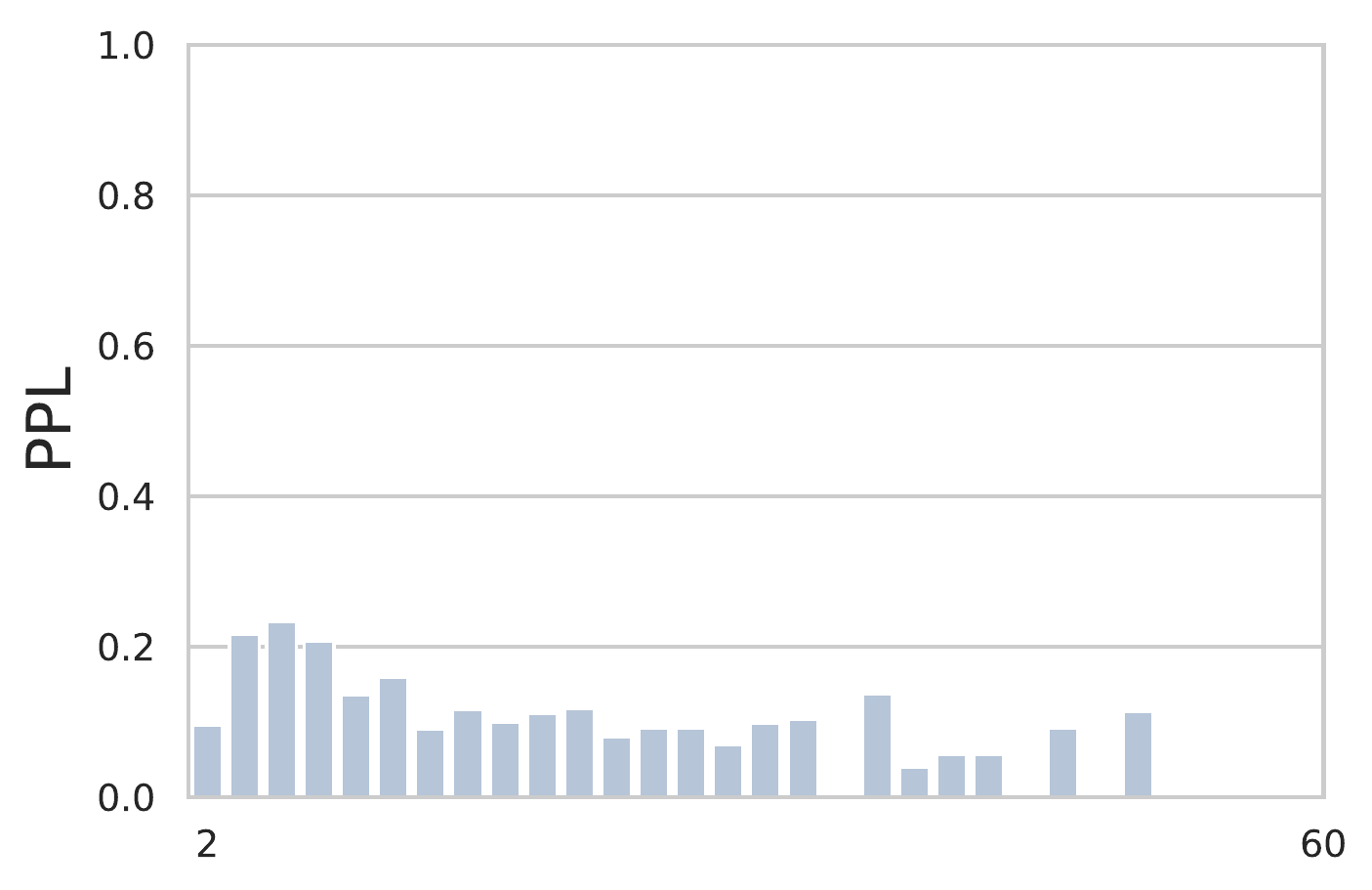}\\
\begin{turn}{90}\OracleMap\end{turn} & \includegraphics[width=\linewidth]{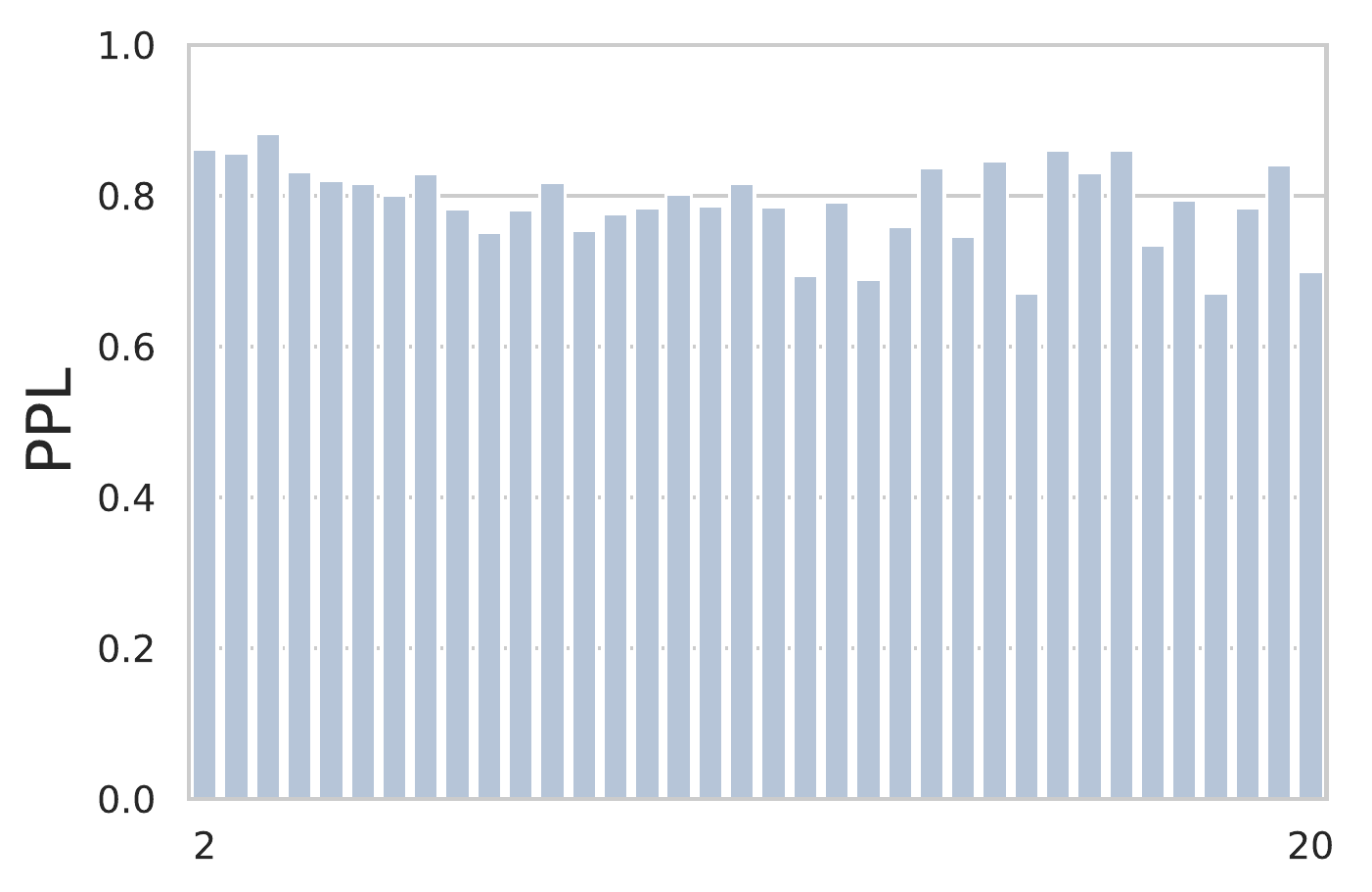} &
\includegraphics[width=\linewidth]{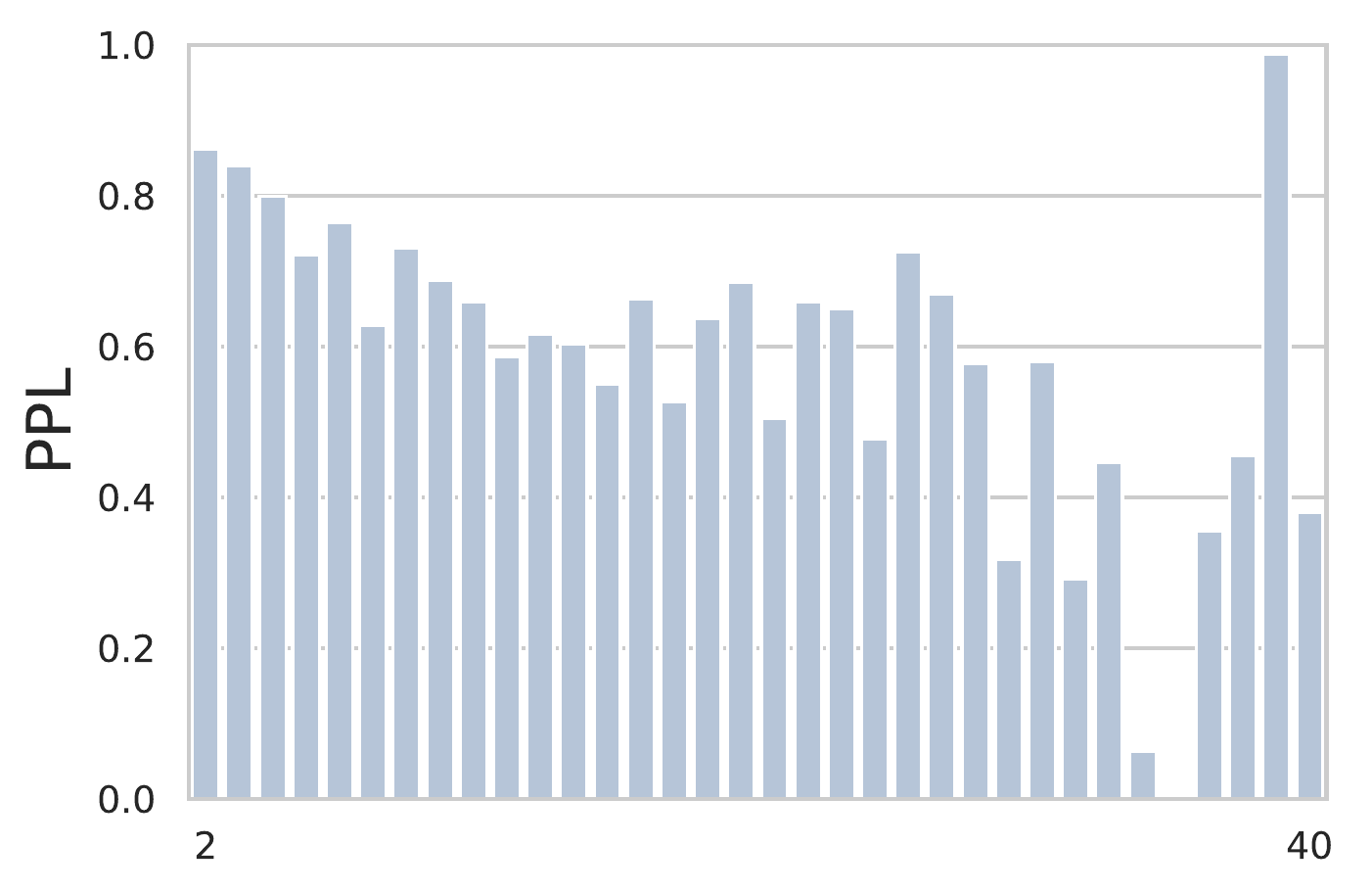} &
\includegraphics[width=\linewidth]{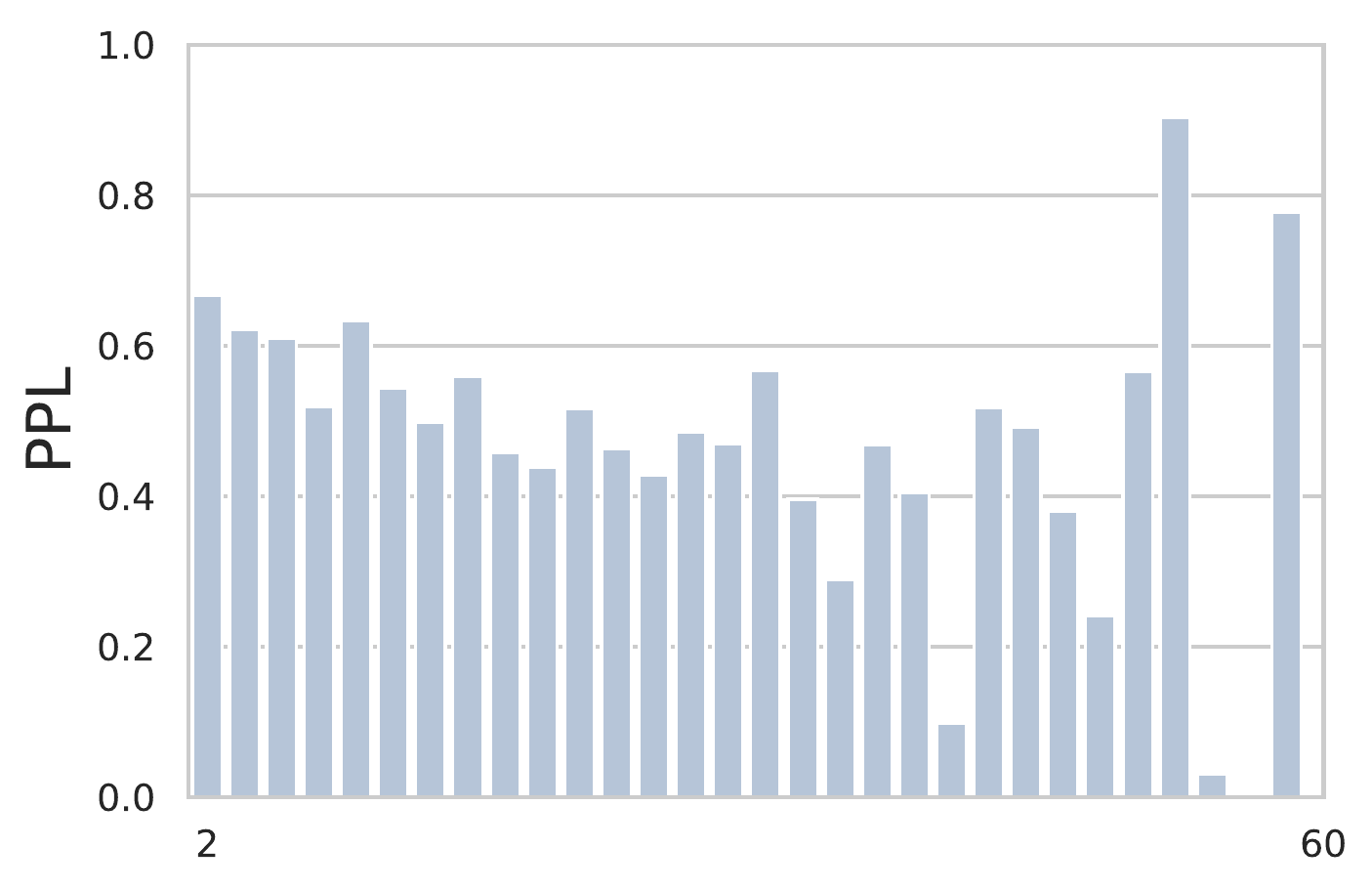}\\
\begin{turn}{90}\OracleEgoMap\end{turn}  & \includegraphics[width=\linewidth]{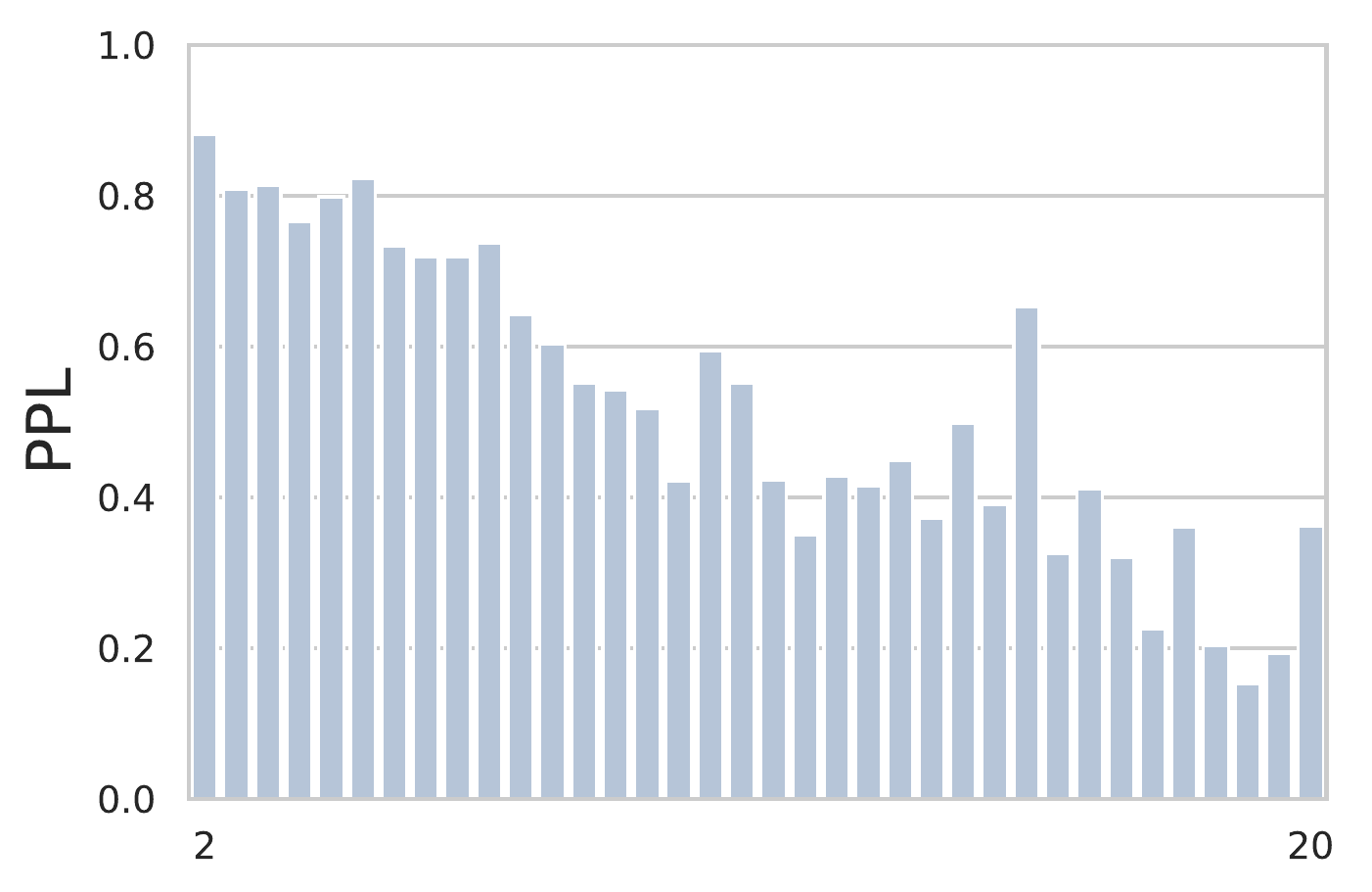} &
\includegraphics[width=\linewidth]{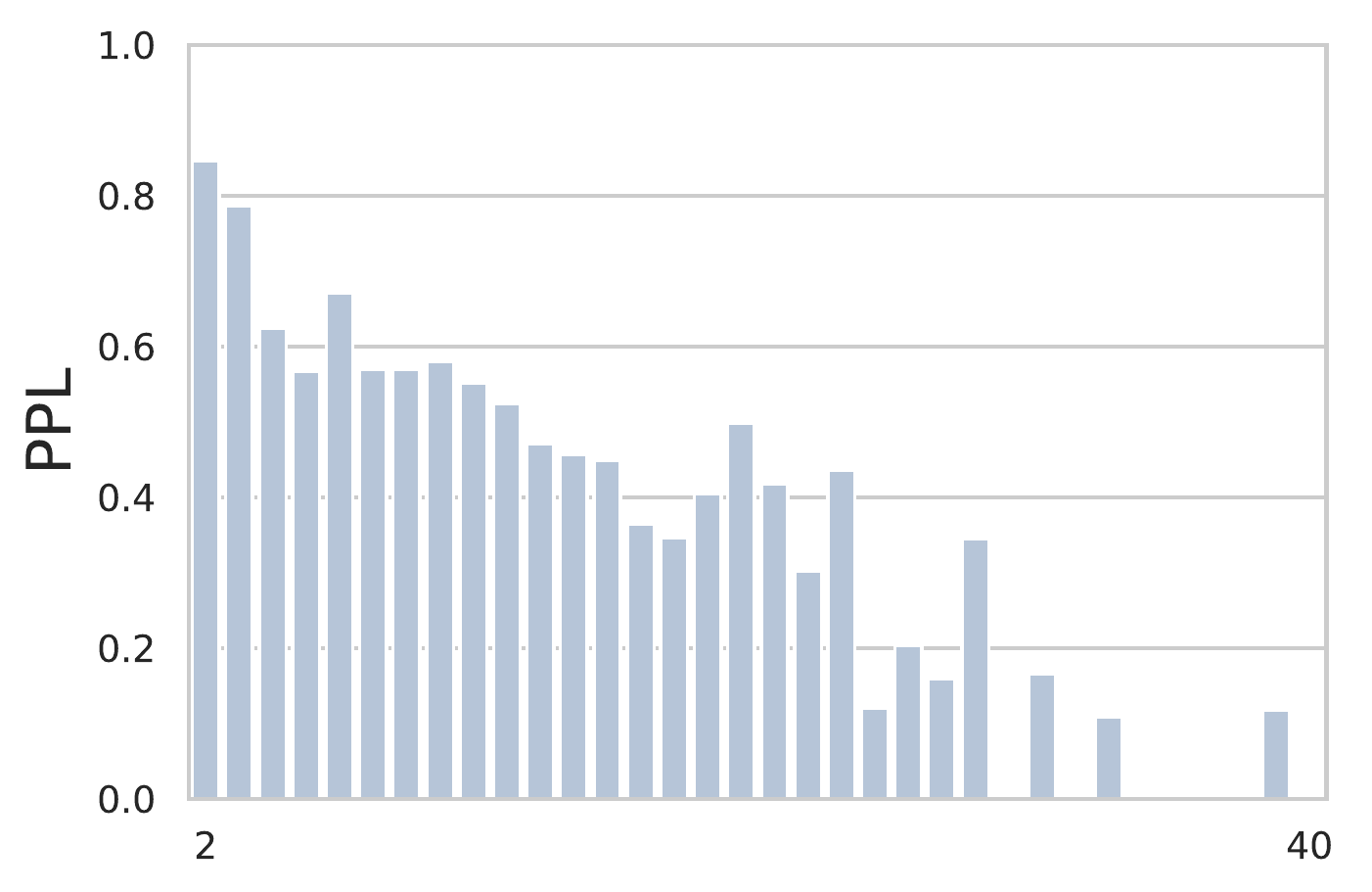} &
\includegraphics[width=\linewidth]{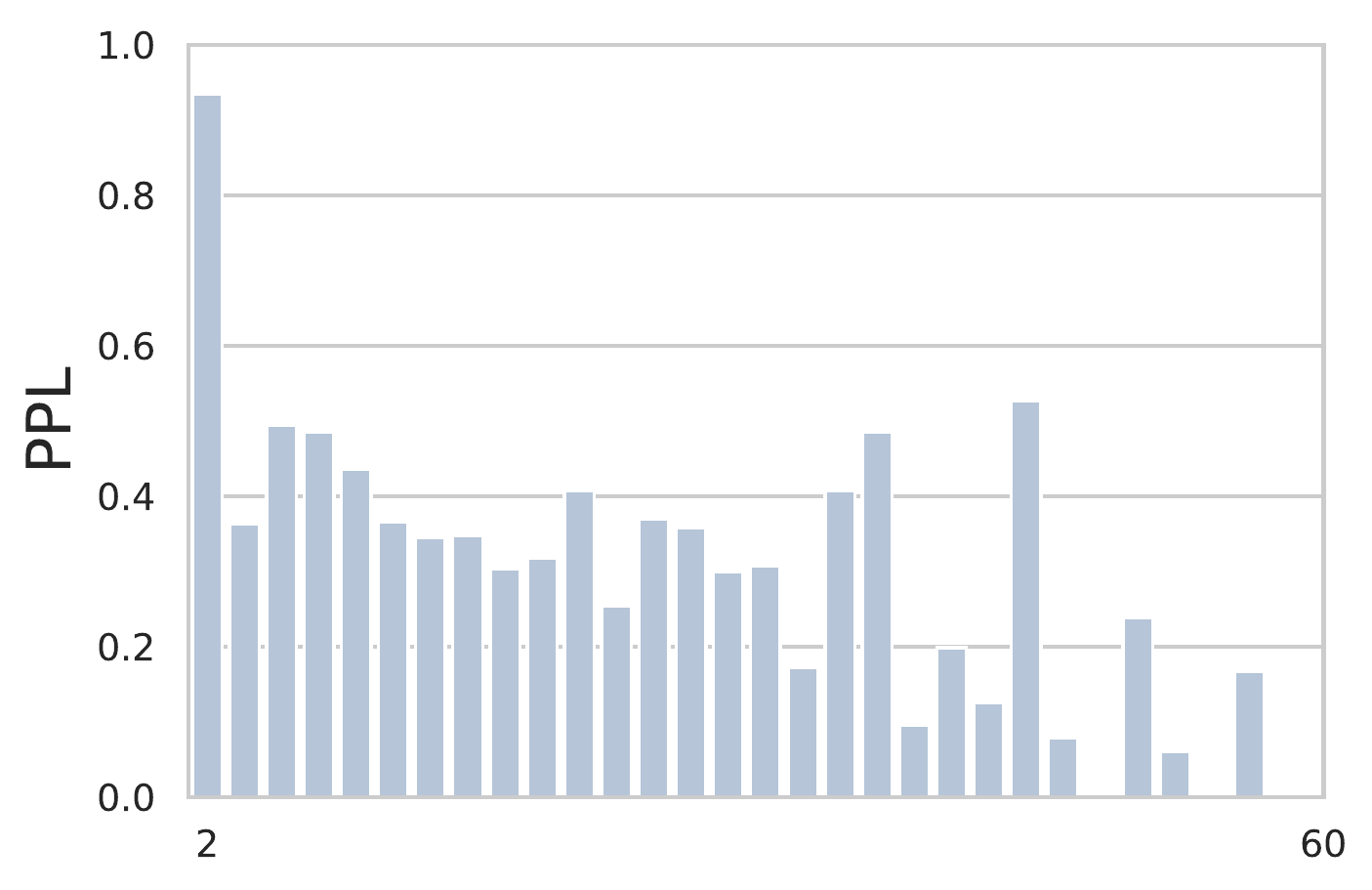}\\
\begin{turn}{90}\ProjNeuralMap\end{turn}  & 
\includegraphics[width=\linewidth]{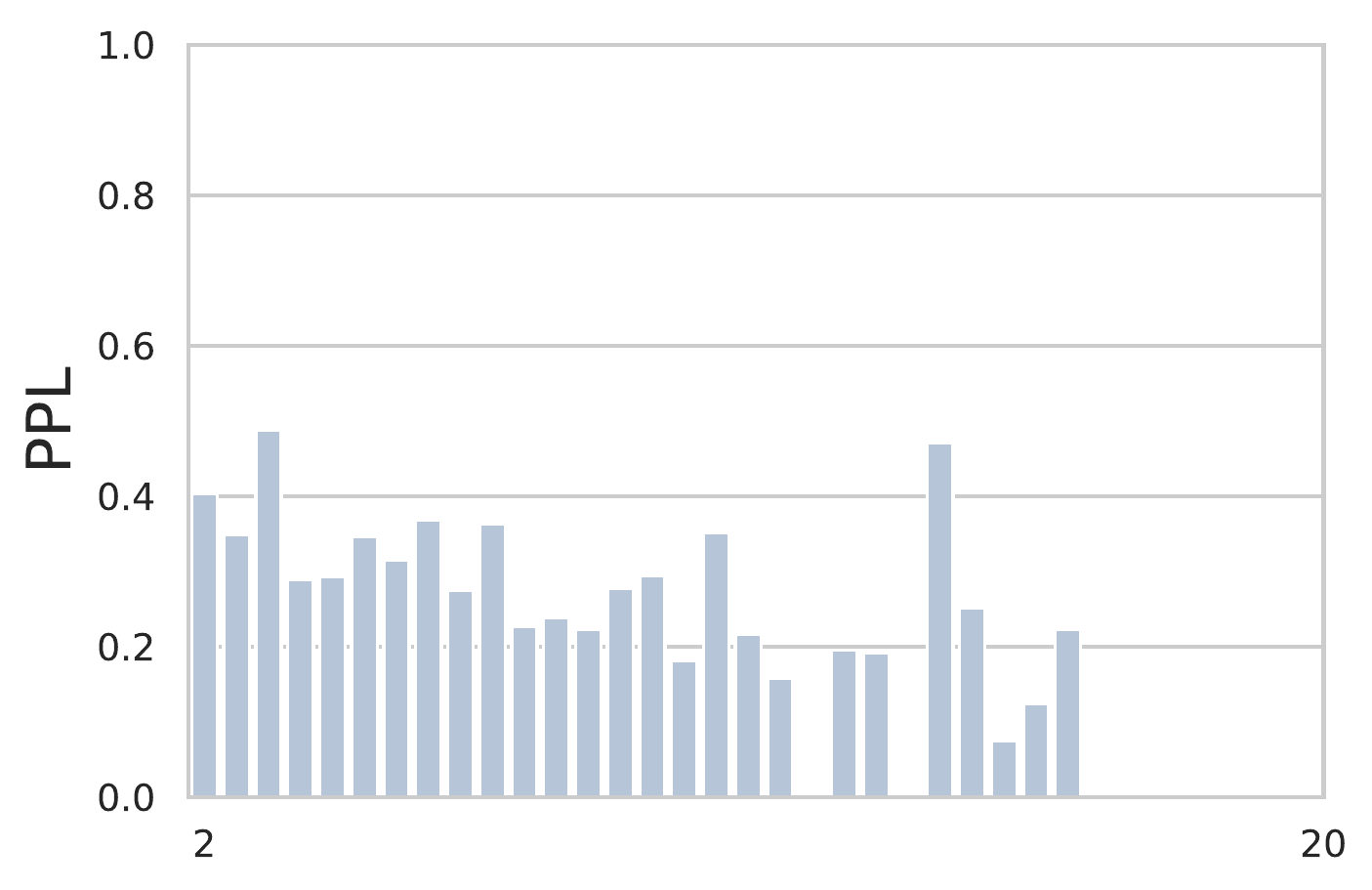} &
\includegraphics[width=\linewidth]{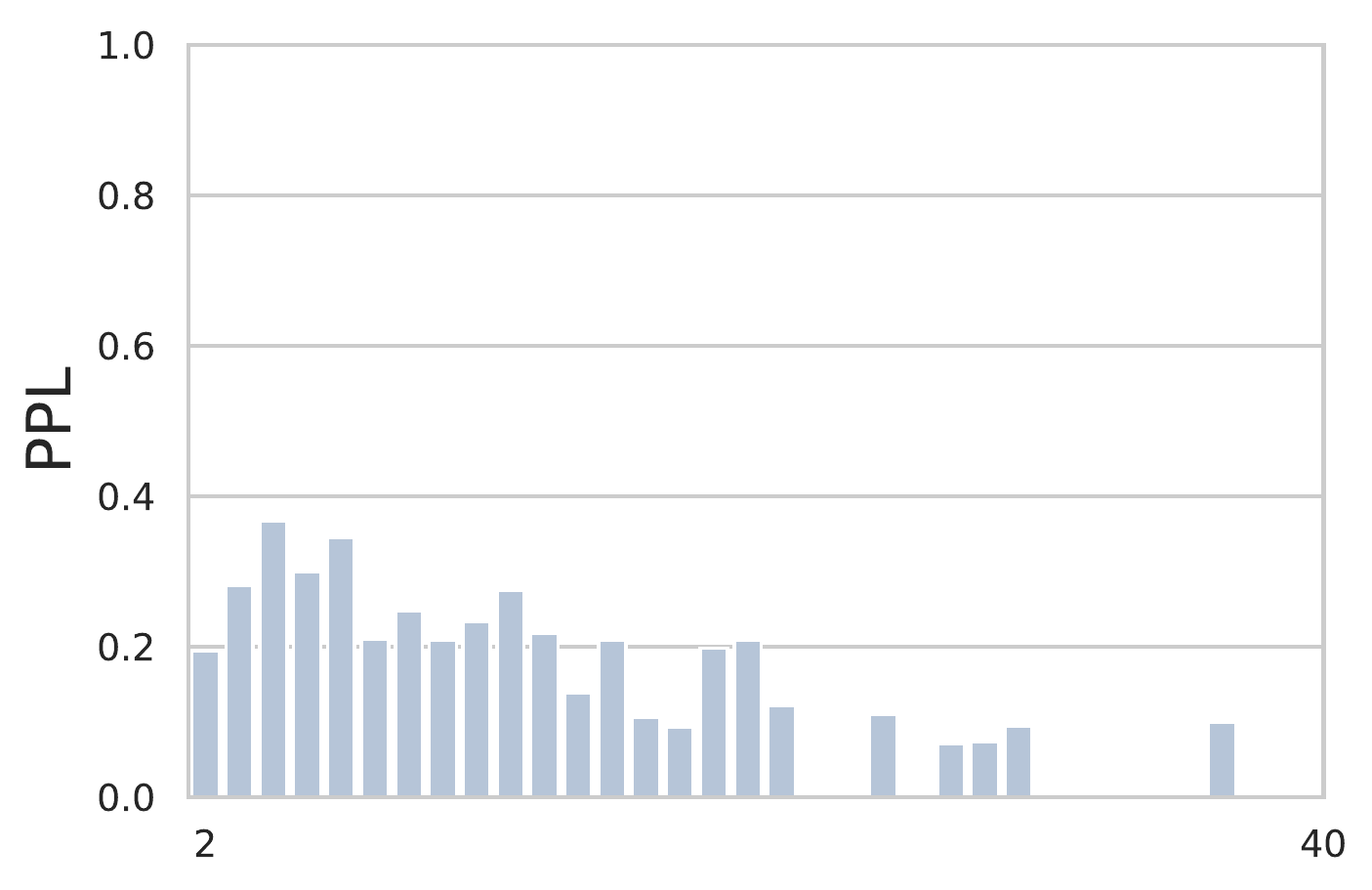} &
\includegraphics[width=\linewidth]{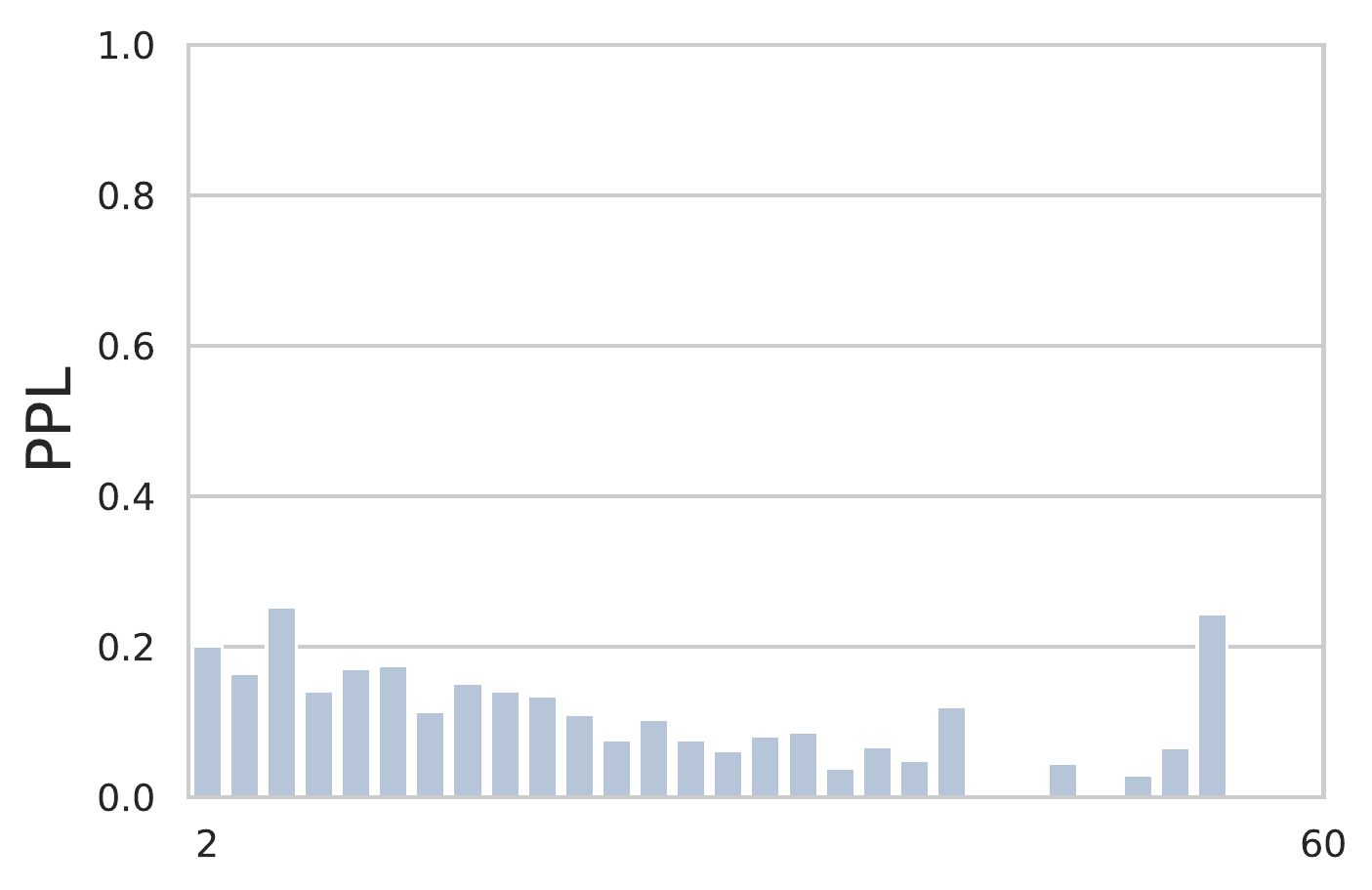}\\
\begin{turn}{90}\ObjRecogMap\end{turn}  & \includegraphics[width=\linewidth]{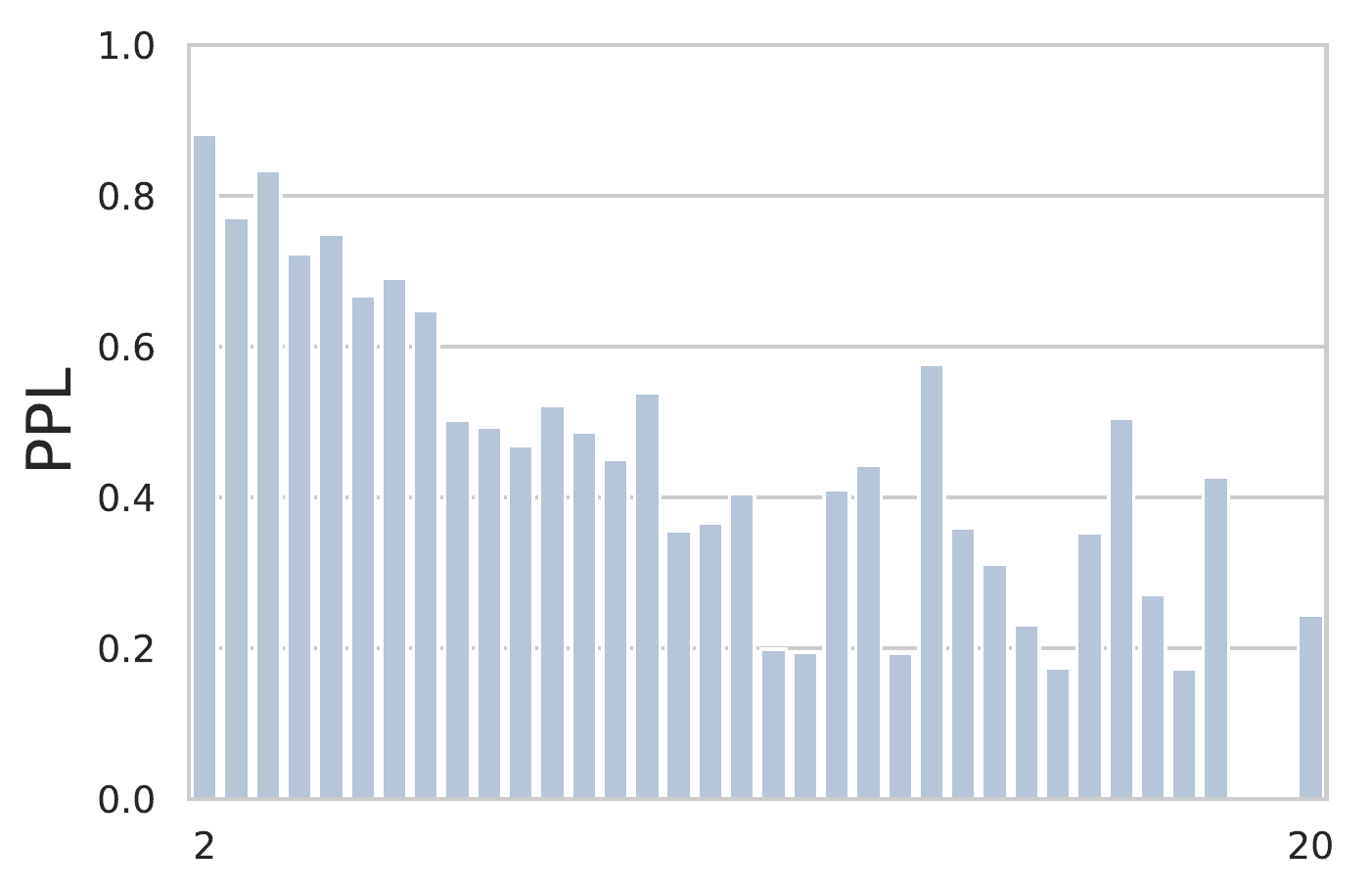} &
\includegraphics[width=\linewidth]{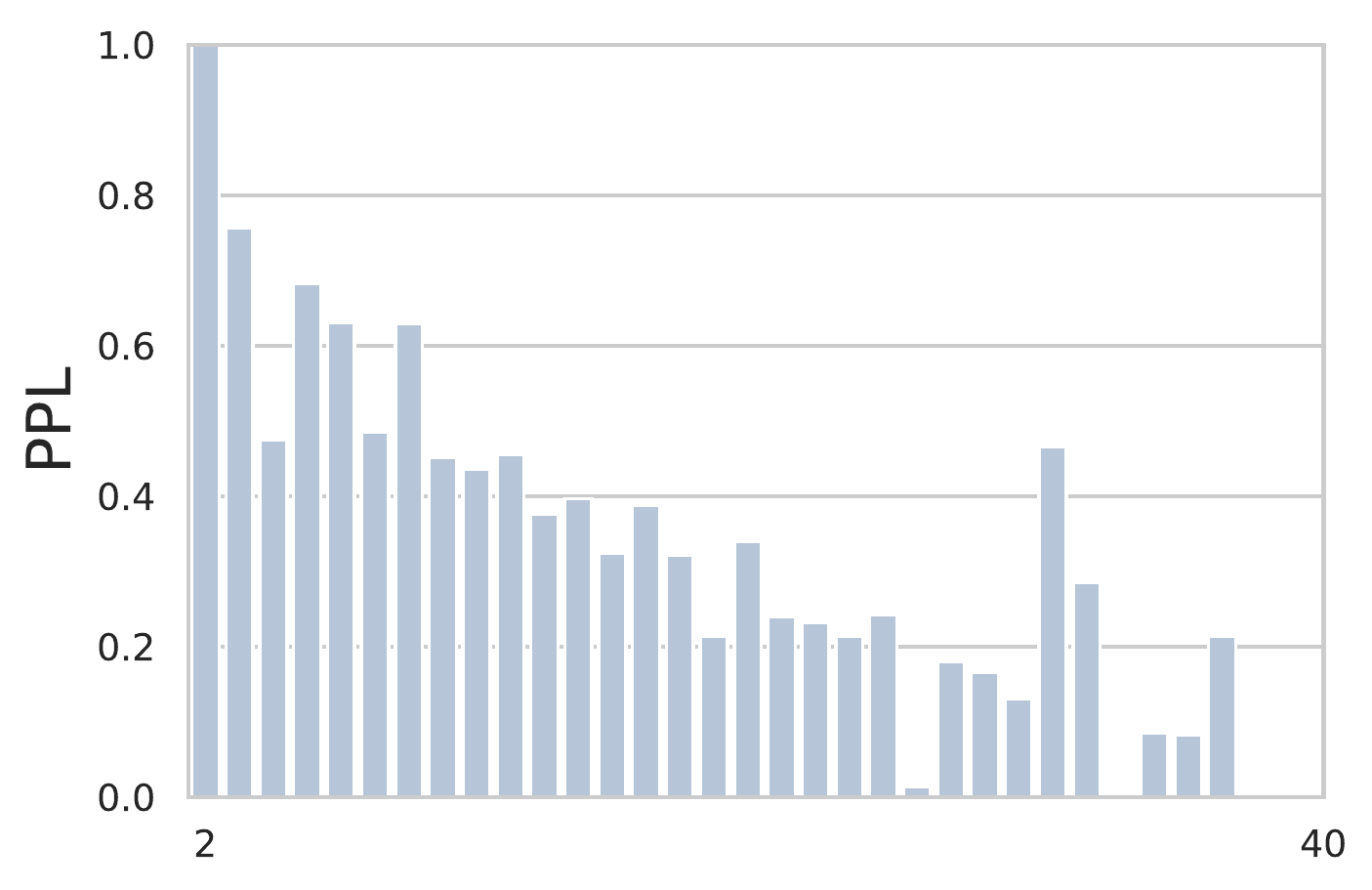} & 
\includegraphics[width=\linewidth]{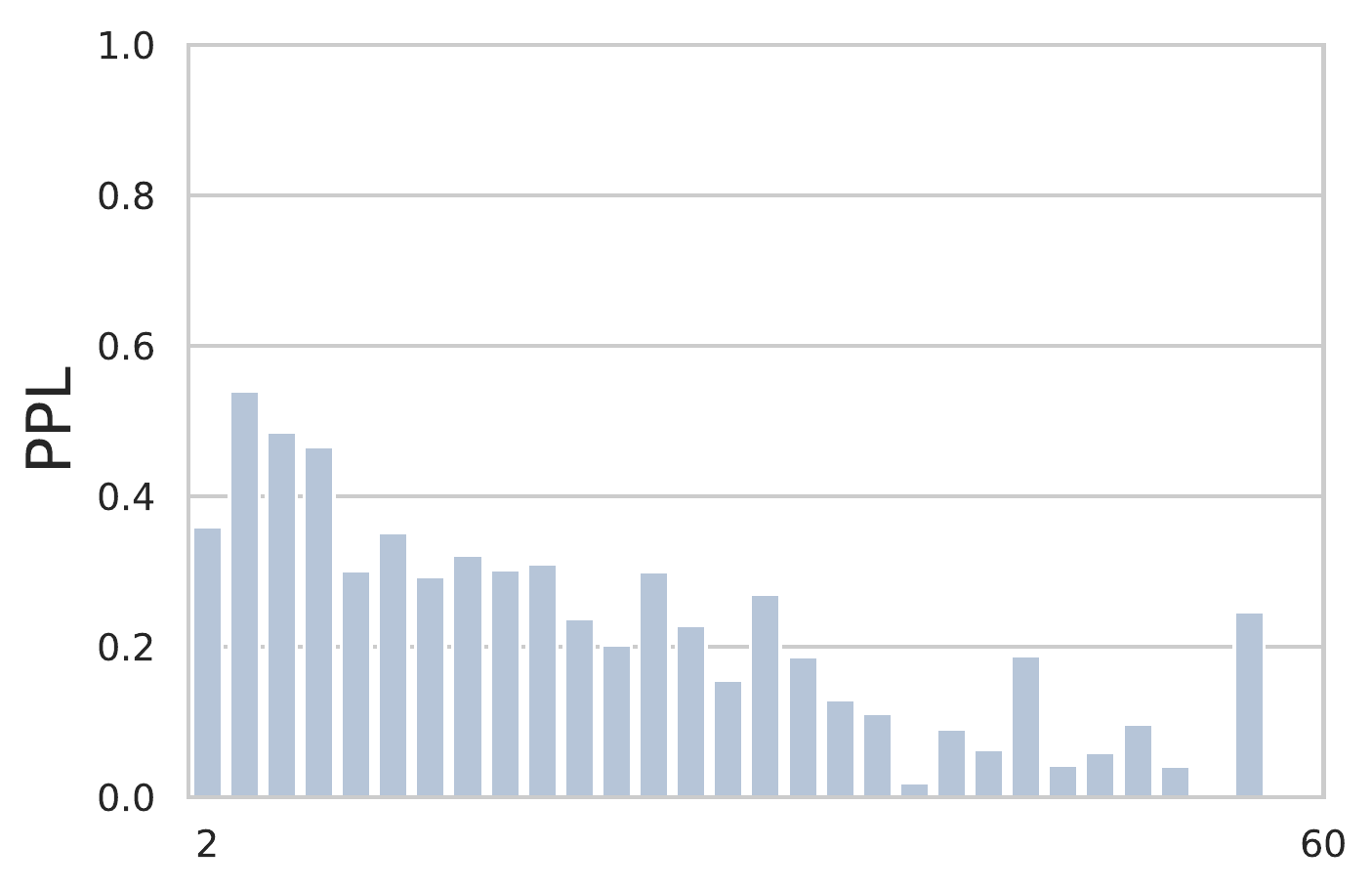}\\
\end{tabular}
\caption{
Plots of average \PPL against geodesic distance along oracle path (i.e. horizontal axis bins episodes into ranges of total episode geodesic distance along the oracle path, and the vertical axis shows average \PPL for all episodes in the corresponding bin).
There is a general trend of decreasing average \PPL with increasing episode distance for all the models (though \OracleMap shows a relatively small drop in performance).
This indicates the escalating difficulty of the \task task as the length of the episodes is increased.
}
\label{fig:difficulty-geodesic-axis-new}
\end{figure}

\xhdr{Analysis of agent performance vs episode length.}
We analyze agent performance based on the episode complexity, as measured by the geodesic distance of the oracle shortest path between the start position and all the goals in the episode.
\Cref{fig:difficulty-geodesic-axis-new} shows the average \PPL for each agent across a range of total episode geodesic distance along the oracle path.
Note that the \OracleMap agent is able to have relatively high \PPL regardless of the episode geodesic distance (with the overall \PPL dropping somewhat from \mon{1} to \mon{3}).
\NoMap has relatively low \PPL and often fails to reach any goals for harder episodes (i.e. episodes with higher geodesic distance).
For both \OracleEgoMap and \ObjRecogMap, we see that the \PPL decreases as we go from easier episodes to harder episodes, indicating the challenge in achieving agent robustness.

\begin{figure}
\begin{tabular}{C{0.1cm}C{4.2cm}C{4.2cm}C{4.2cm}}
& \mon{1} & \mon{2} & \mon{3} \\
\begin{turn}{90}\Progress\end{turn} & \includegraphics[width=\linewidth]{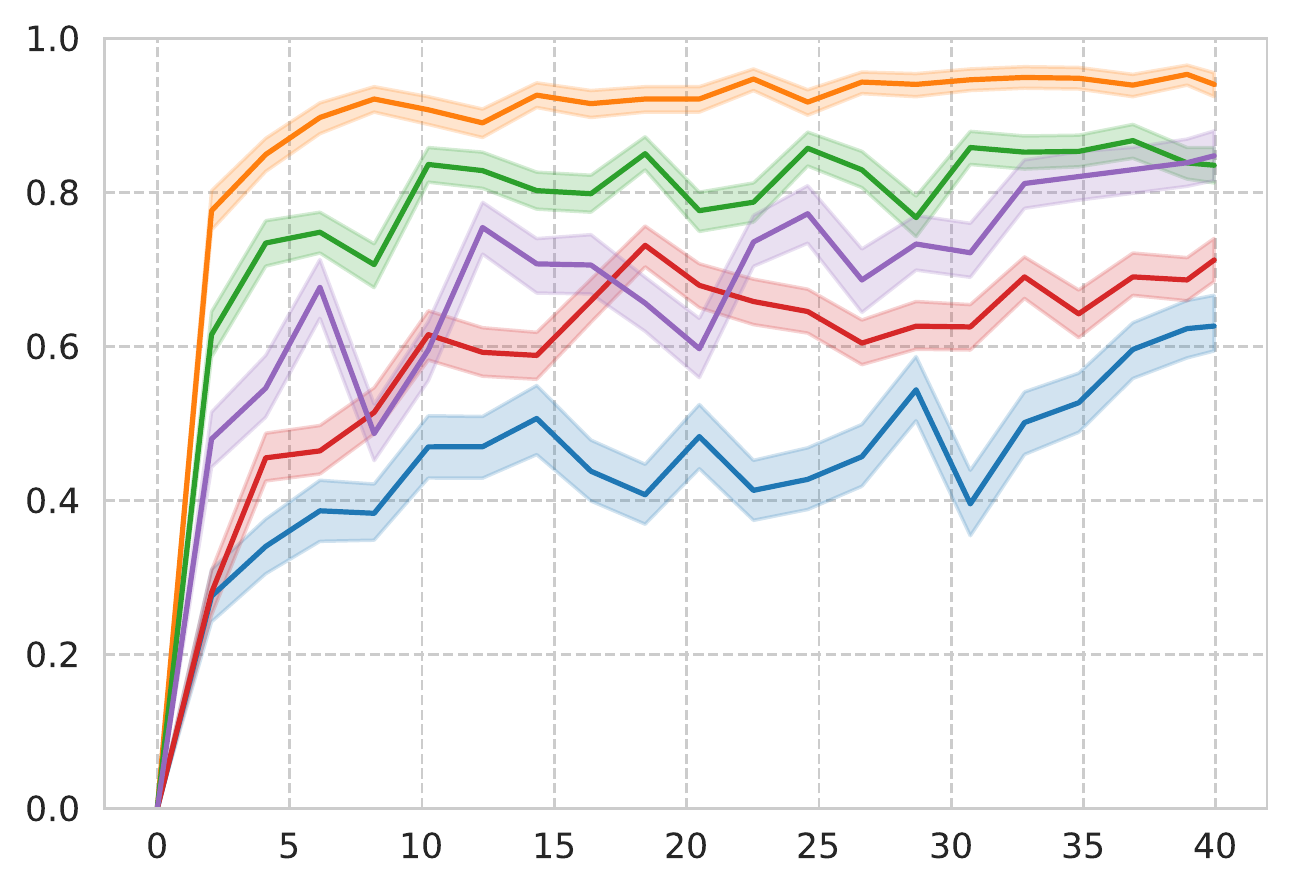} &
\includegraphics[width=\linewidth]{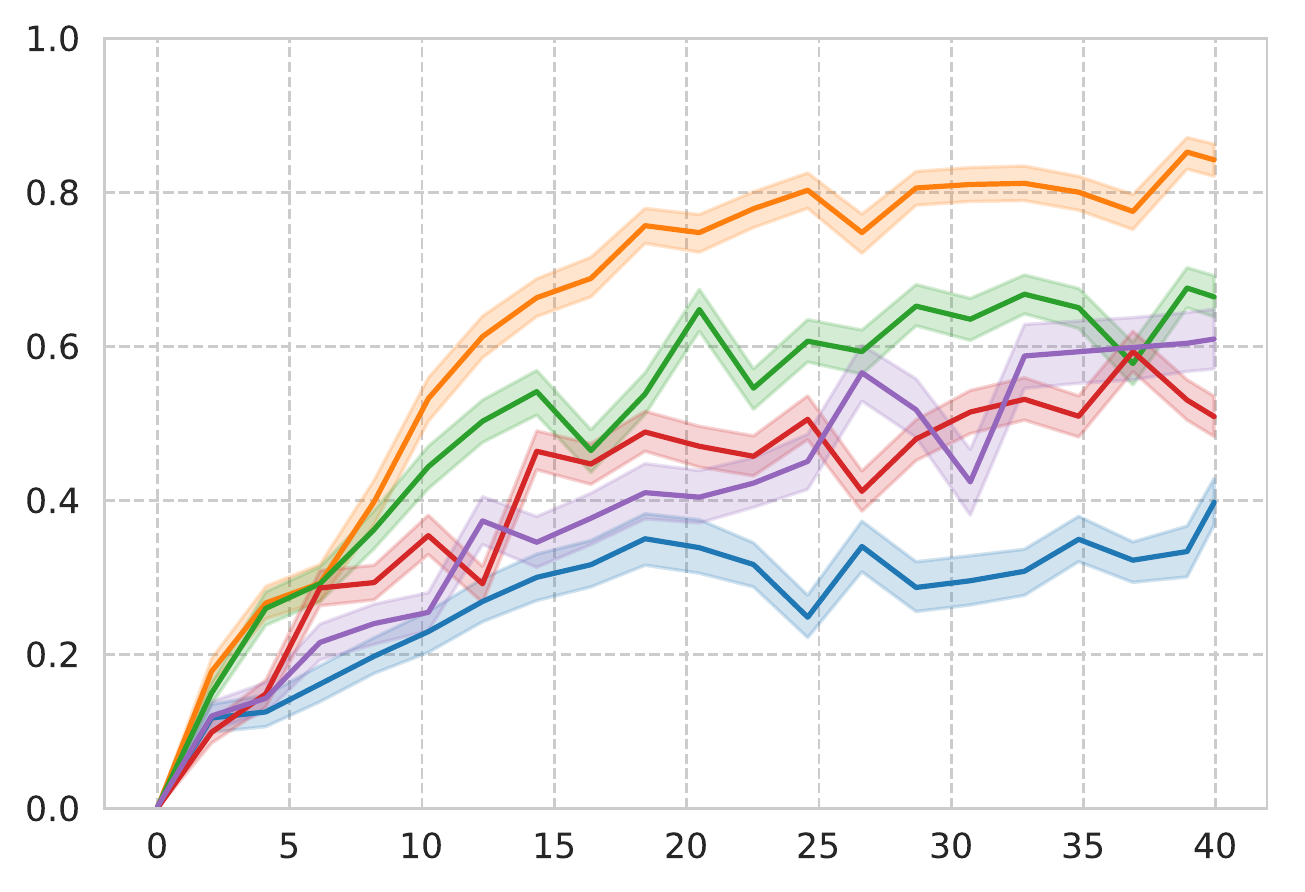} &
\includegraphics[width=\linewidth]{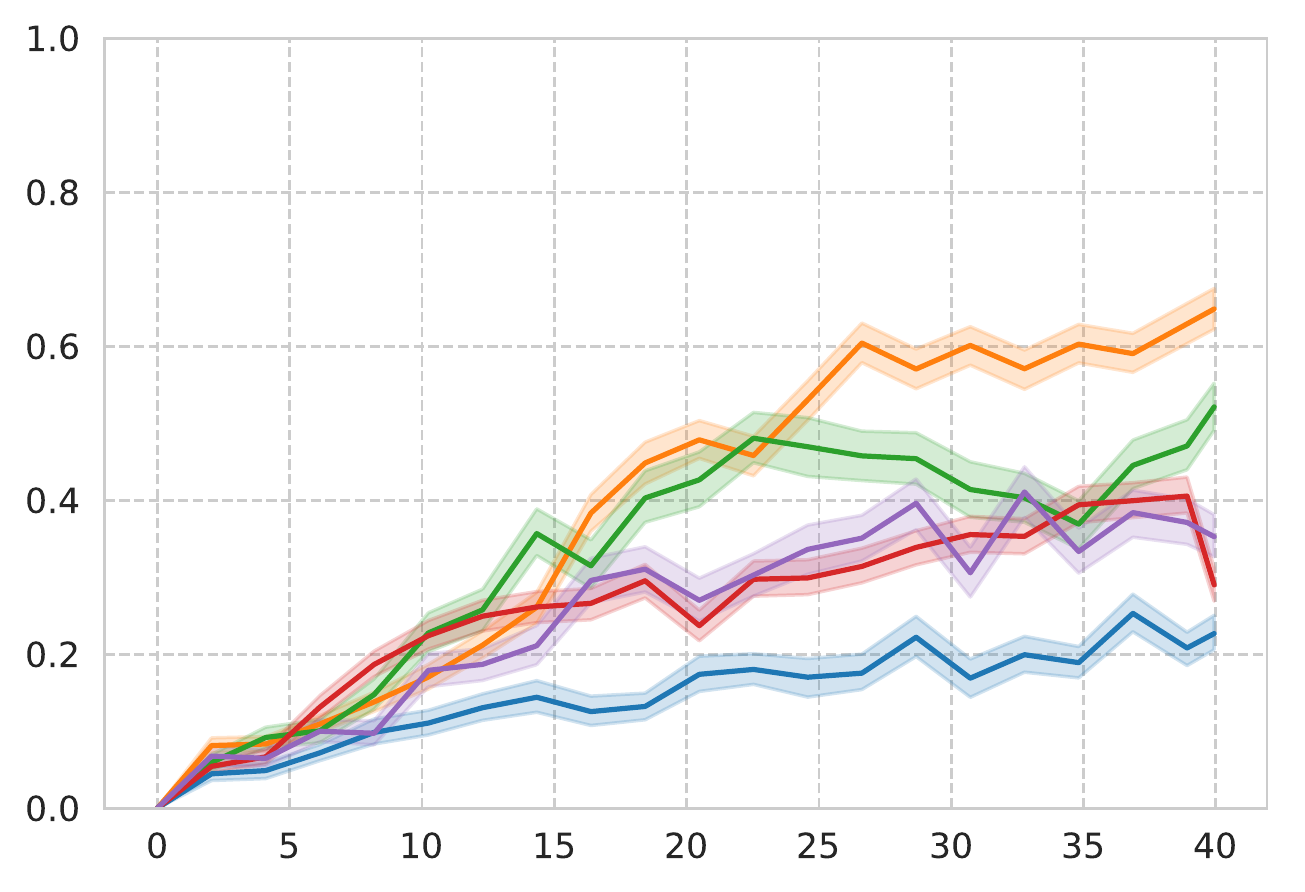}\\
\begin{turn}{90}\Success\end{turn} & \includegraphics[width=\linewidth]{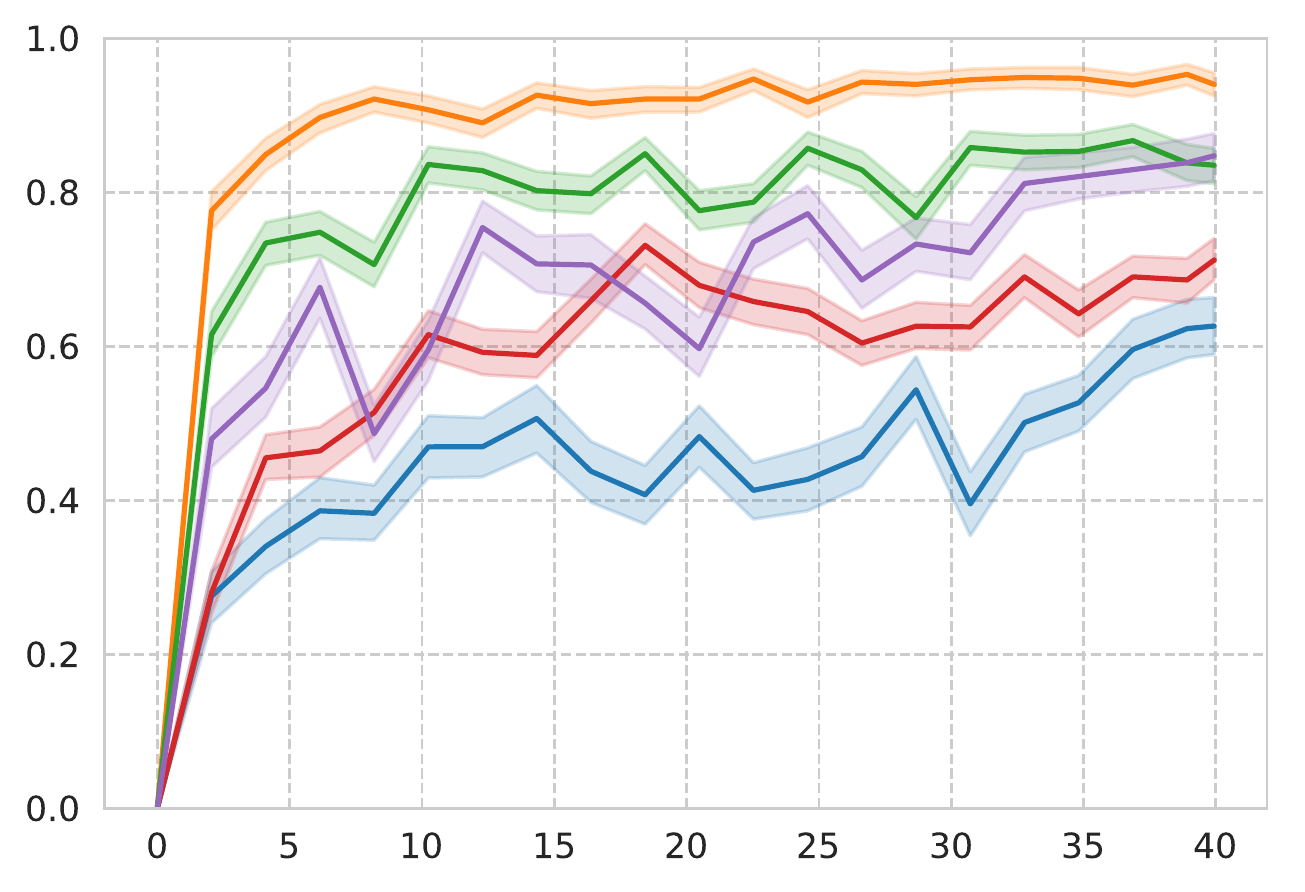} &
\includegraphics[width=\linewidth]{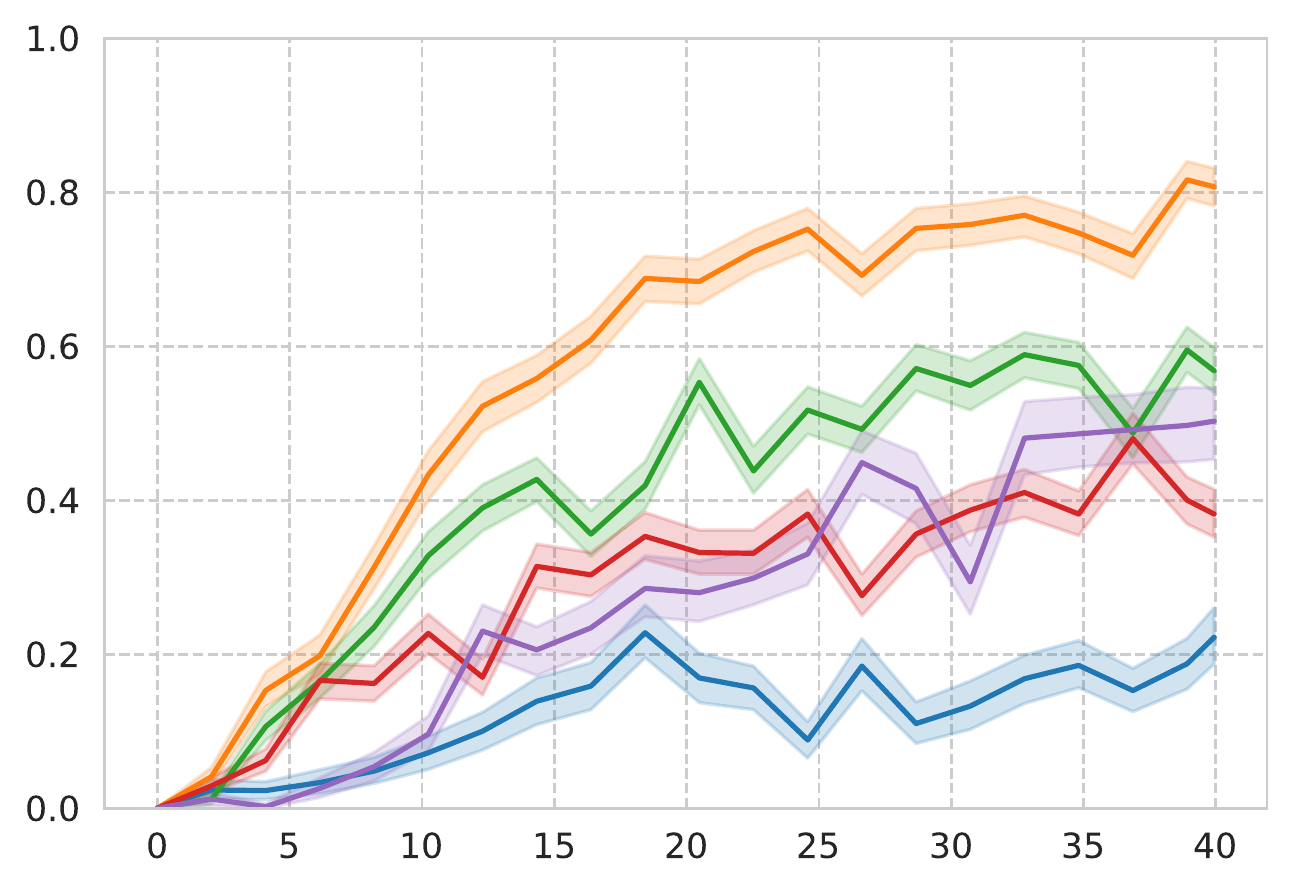} &
\includegraphics[width=\linewidth]{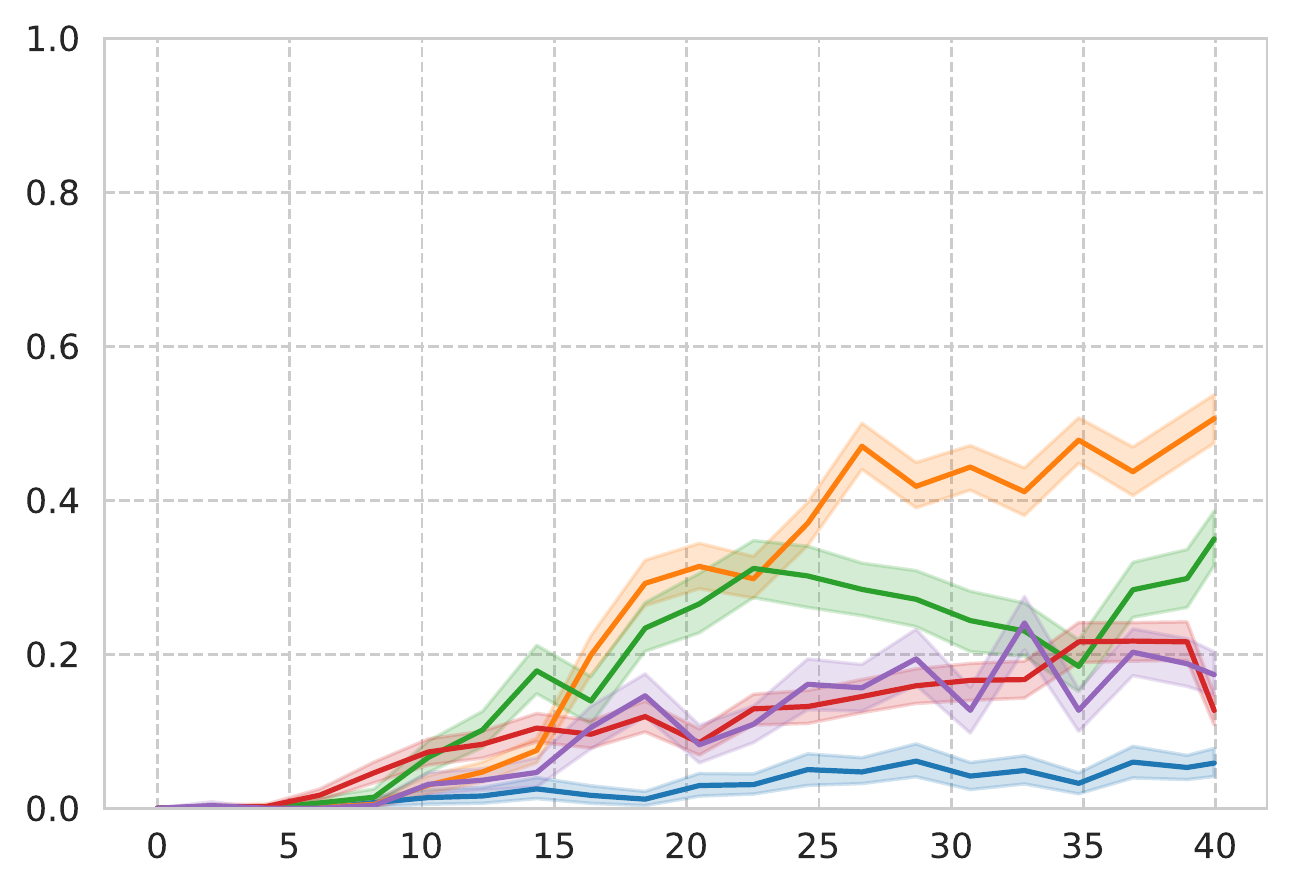}\\
\begin{turn}{90}\SPL\end{turn}  & \includegraphics[width=\linewidth]{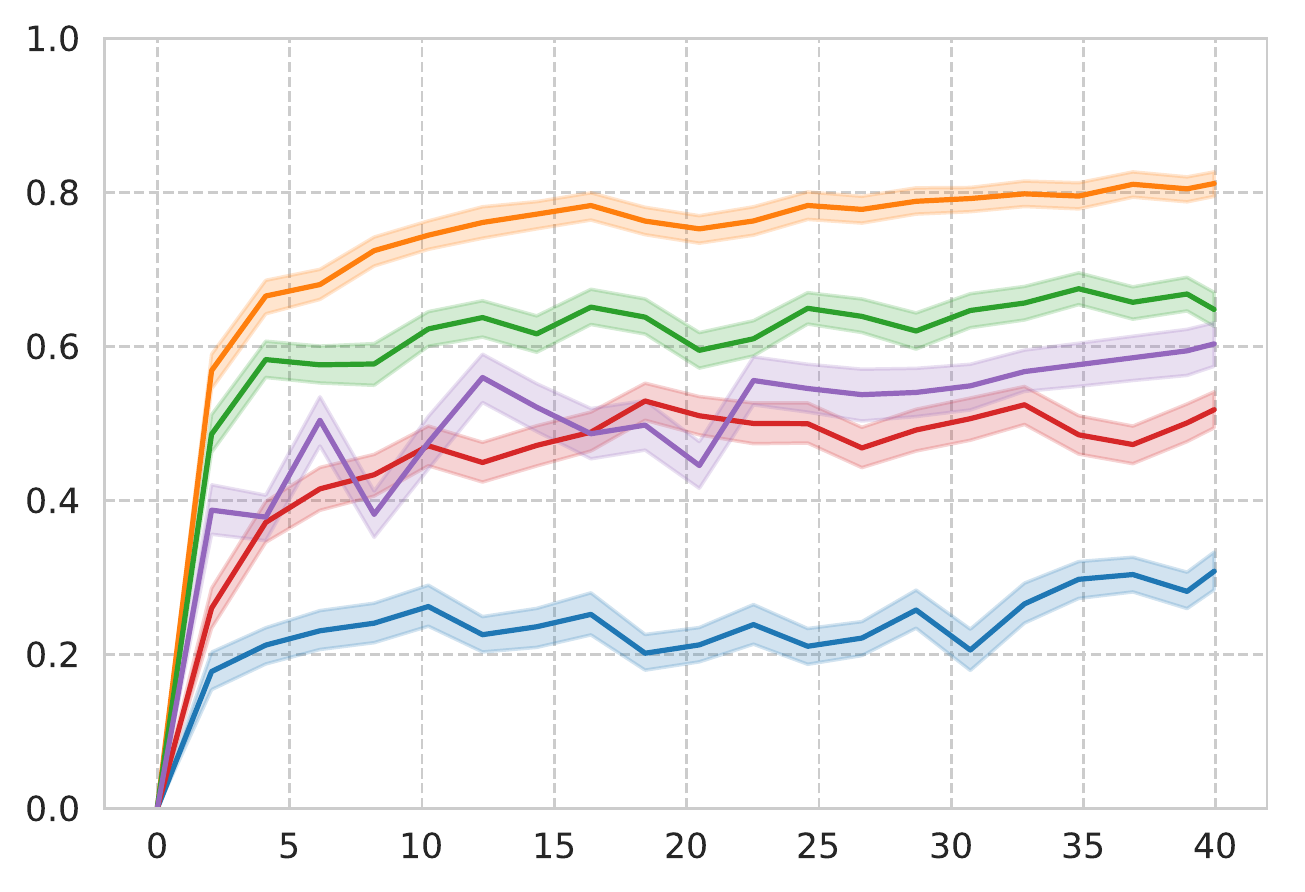} &
\includegraphics[width=\linewidth]{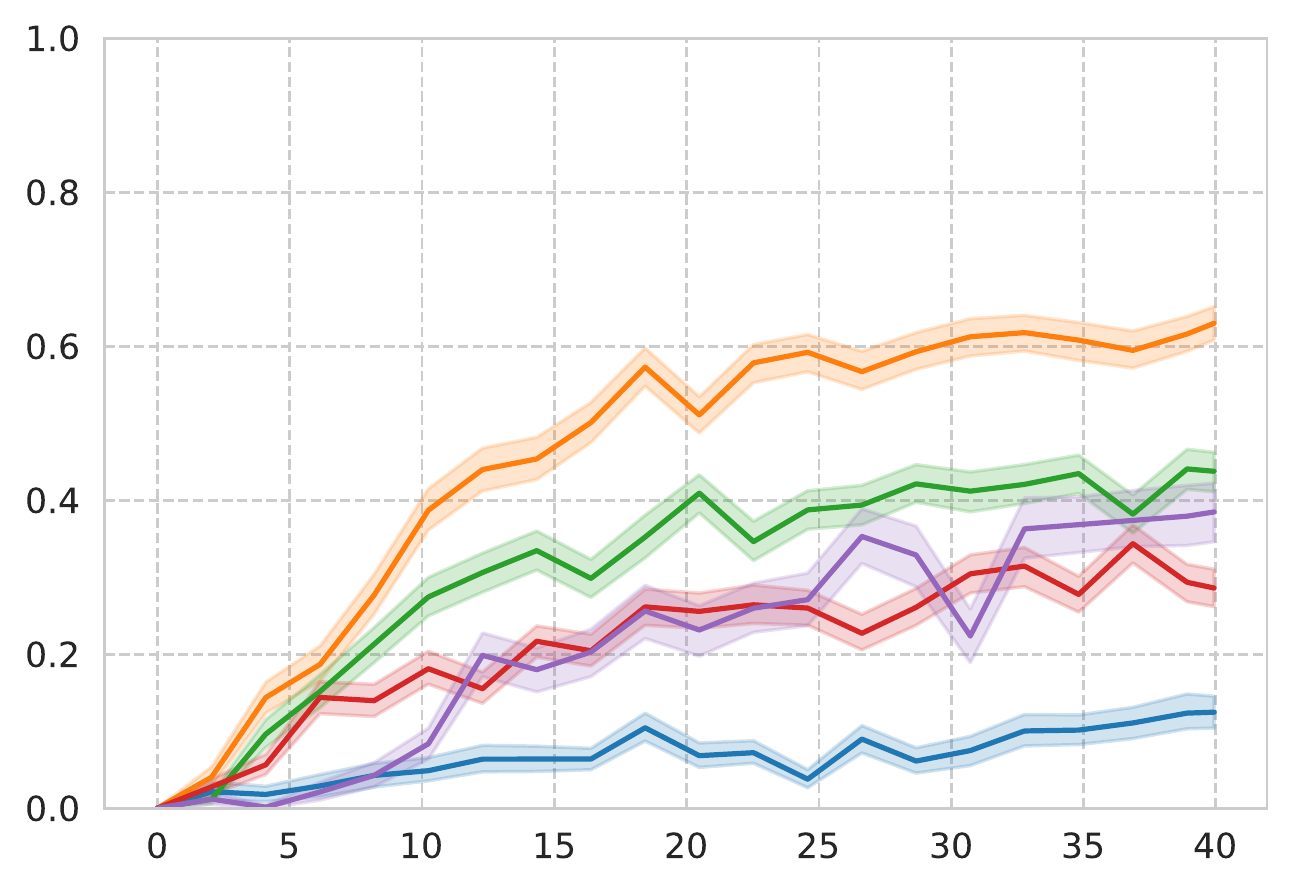} &
\includegraphics[width=\linewidth]{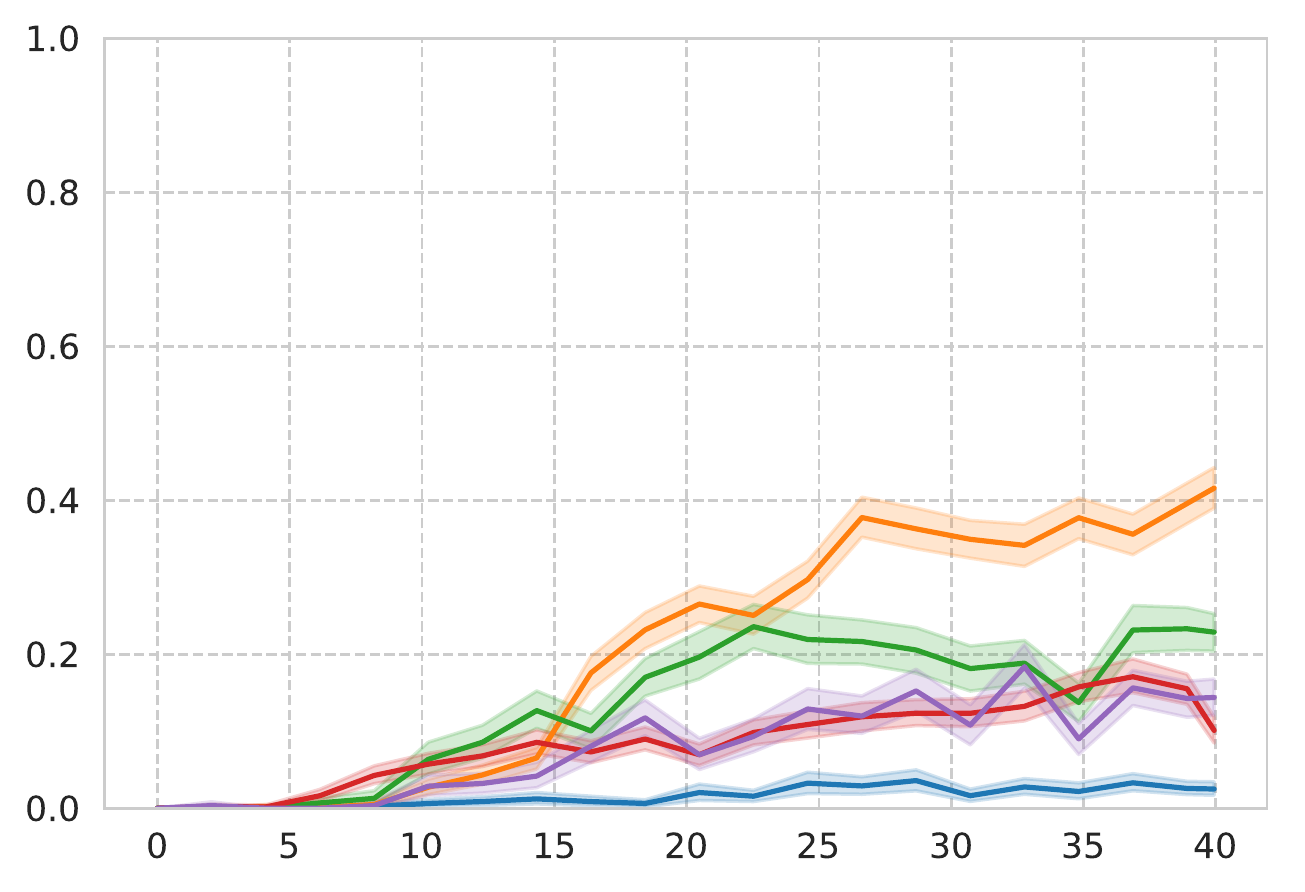}\\
\end{tabular}
\includegraphics[width=0.98\linewidth]{plots/legend.pdf}
\caption{Evaluation metrics of agents during training, evaluated on validation set with $95\%$ CI indicated by shading.
Overall performance decreases across all models as task complexity increases. The {\color{orange}{\OracleMap}} agent
trains the fastest and reaches the best overall performance. {\color{cadmiumgreen}{\OracleEgoMap}} and {\color{amethyst}{\ObjRecogMap}} follow
closely, while the {\color{red}{\ProjNeuralMap}} and {\color{blue}{\NoMap}} agents perform the worst.}

\label{fig:more-valplots}
\end{figure}

\xhdr{Agent performance during training.}
In \Cref{fig:more-valplots} we plot the \SPL, \Success, and \Progress metrics for the different agents on the validation set (see main paper Figure 4 for the \PPL).
All the metrics follow the same general trends as observed in the main paper.

\begin{figure}[ht]
\resizebox{\linewidth}{!}{
\newcolumntype{C}{>{\centering\arraybackslash} m{2.7cm} }
\begin{tabular}{@{}CCCCC@{}}
\toprule
\NoMap & \OracleMap & \OracleEgoMap & \ProjNeuralMap & \ObjRecogMap \\ \midrule

     \multicolumn{5}{c}{Goal order: 1\goal{yellow}, 2\goal{red}, 3\goal{cyangoal}} \\
        \includegraphics[trim={10cm 55cm 72cm 0cm},clip,width=1\linewidth]{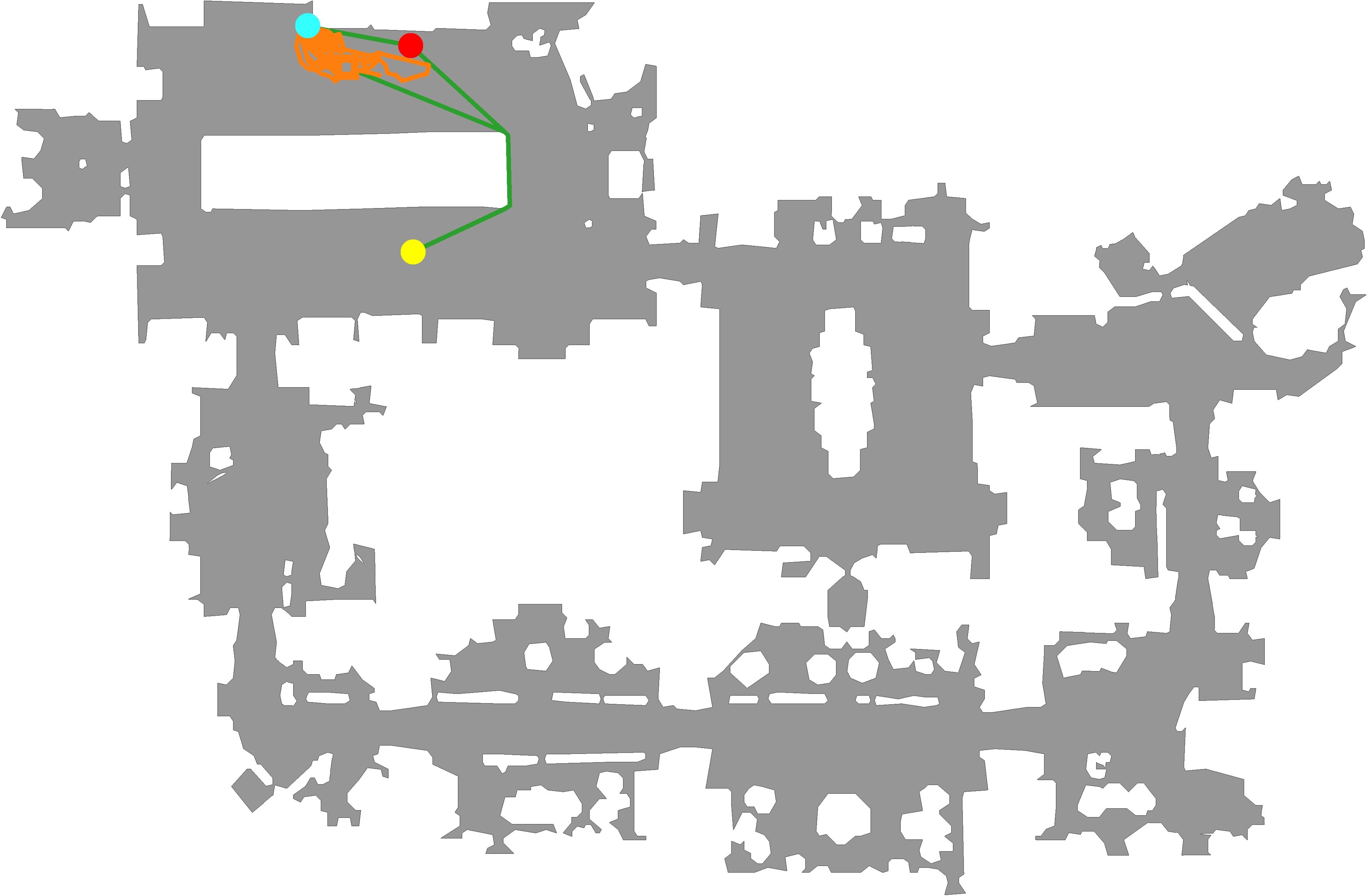} \newline \vizc{0}{0}    &    \includegraphics[trim={10cm 55cm 72cm 0cm},clip,width=1\linewidth]{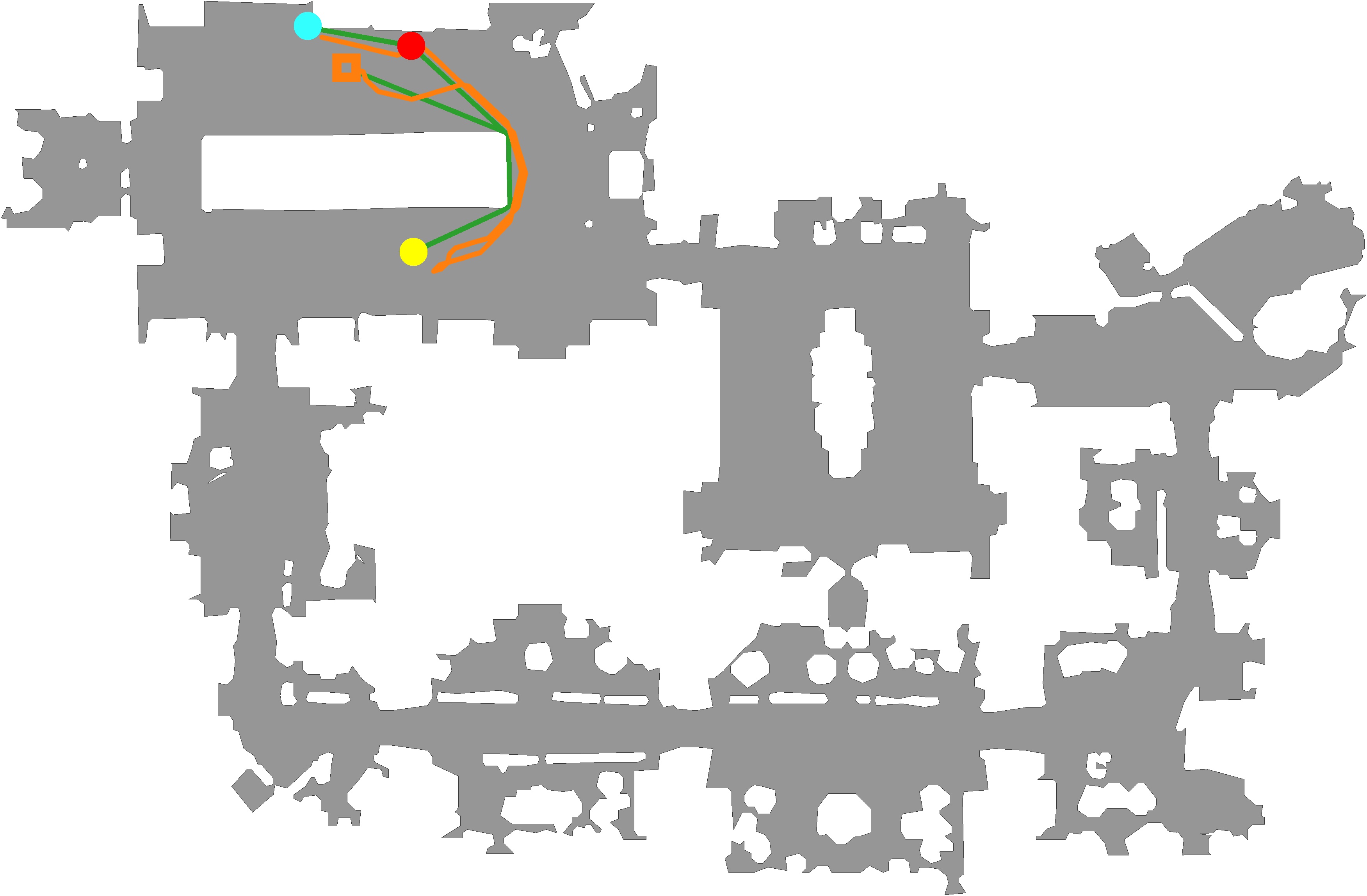} \newline \vizc{1}{0.95}           & \includegraphics[trim={10cm 55cm 72cm 0cm},clip,width=1\linewidth]{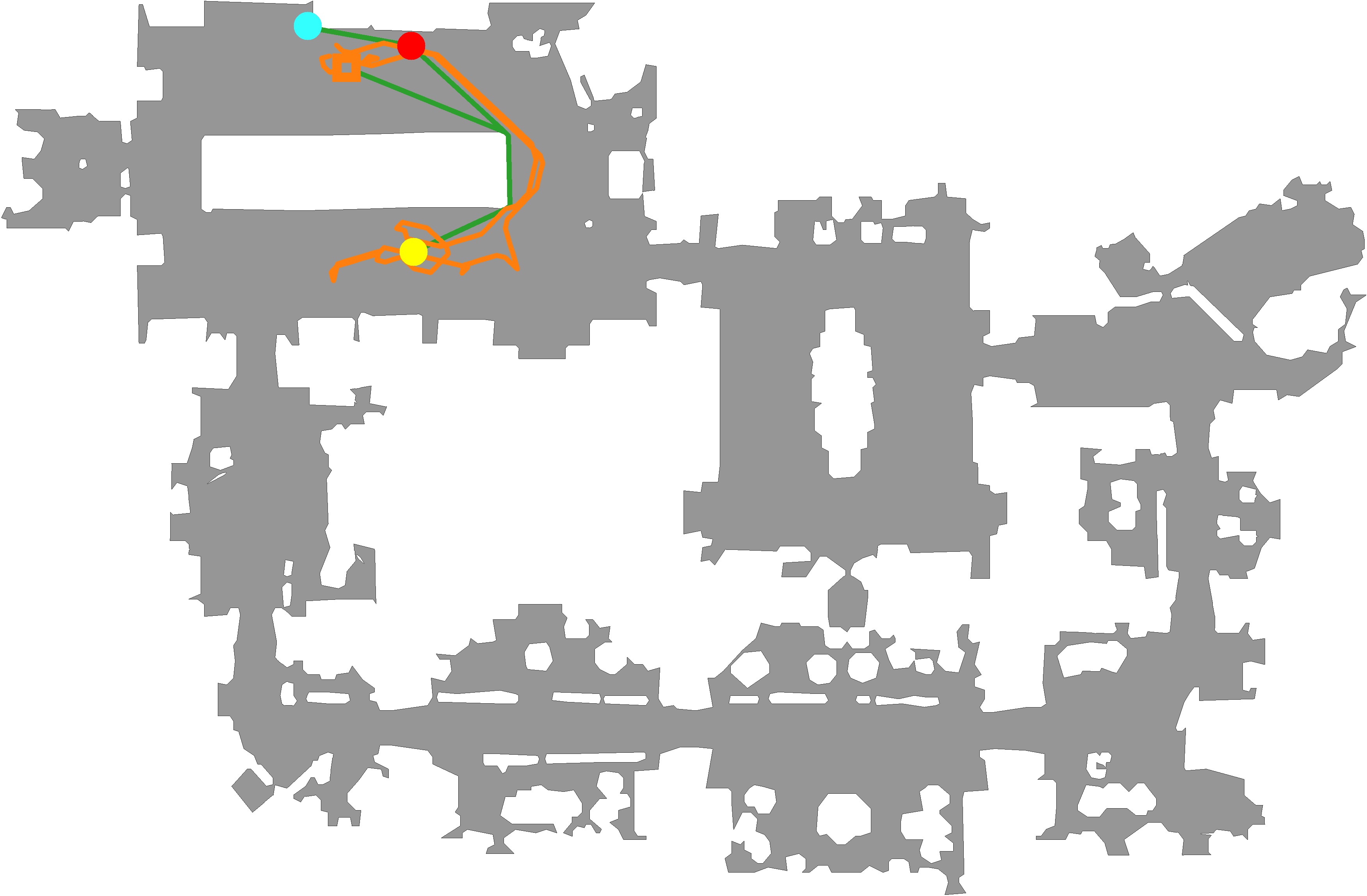} \newline \vizc{1}{0.54}         &   \includegraphics[trim={10cm 55cm 72cm 0cm},clip,width=1\linewidth]{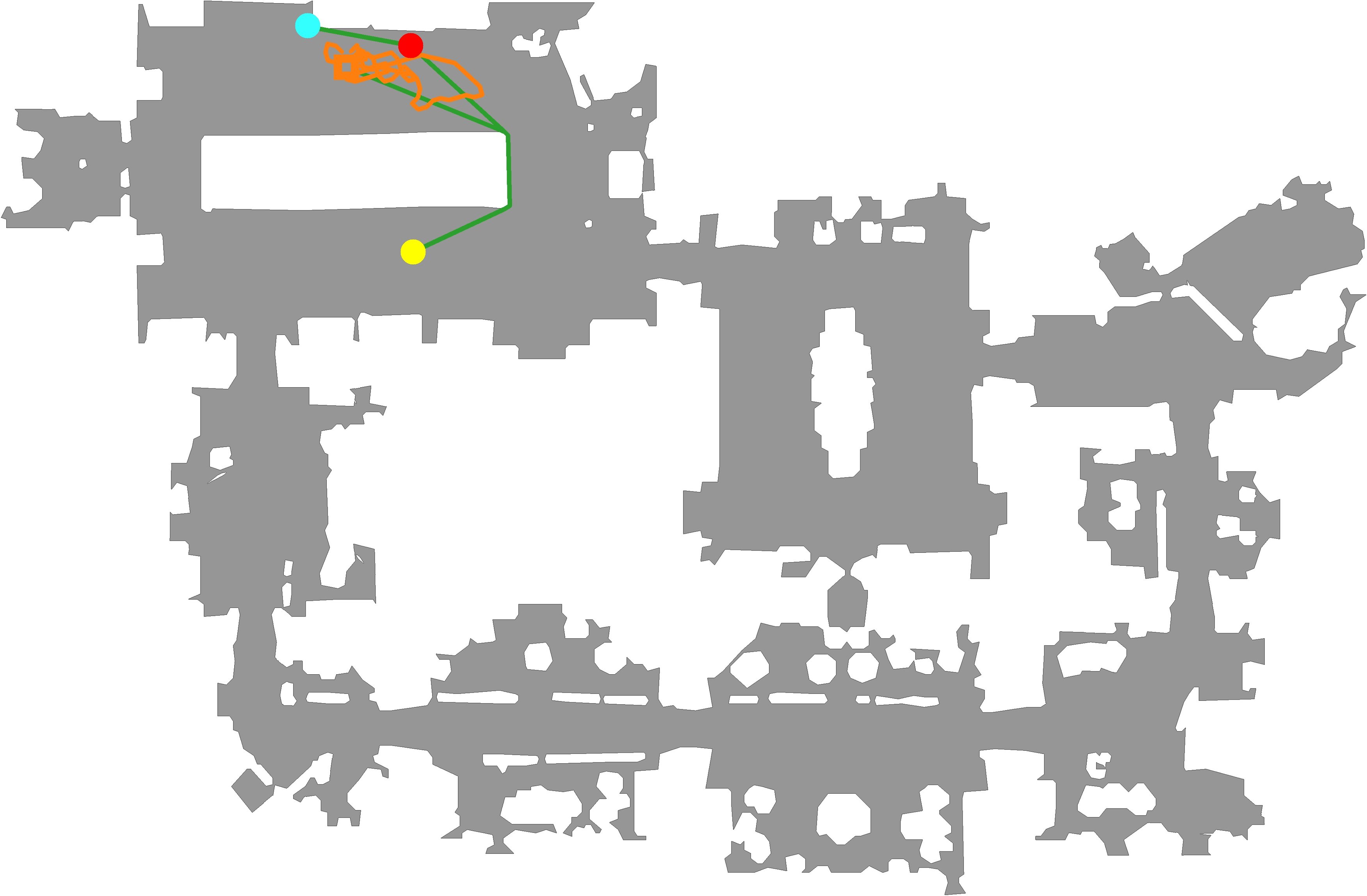} \newline \vizc{0}{0}    &   \includegraphics[trim={10cm 55cm 72cm 0cm},clip,width=1\linewidth]{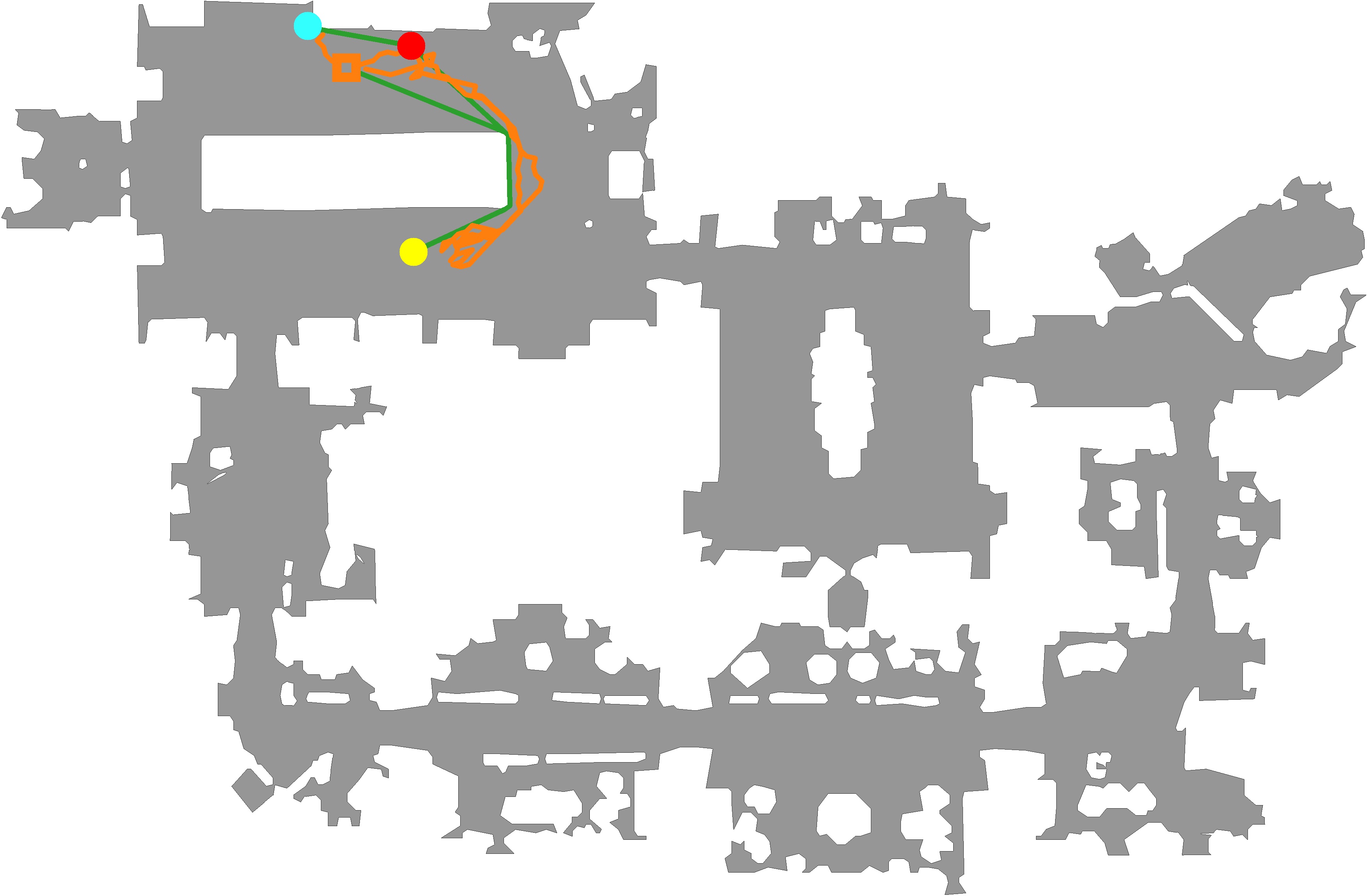}  \newline \vizc{1}{0.66}    \\ \\
        
     \multicolumn{5}{c}{Goal order: 1\goal{blue}, 2\goal{black}, 3\goal{magentagoal}} \\
         \includegraphics[trim={35cm 45cm 32cm 10cm},clip,width=1\linewidth]{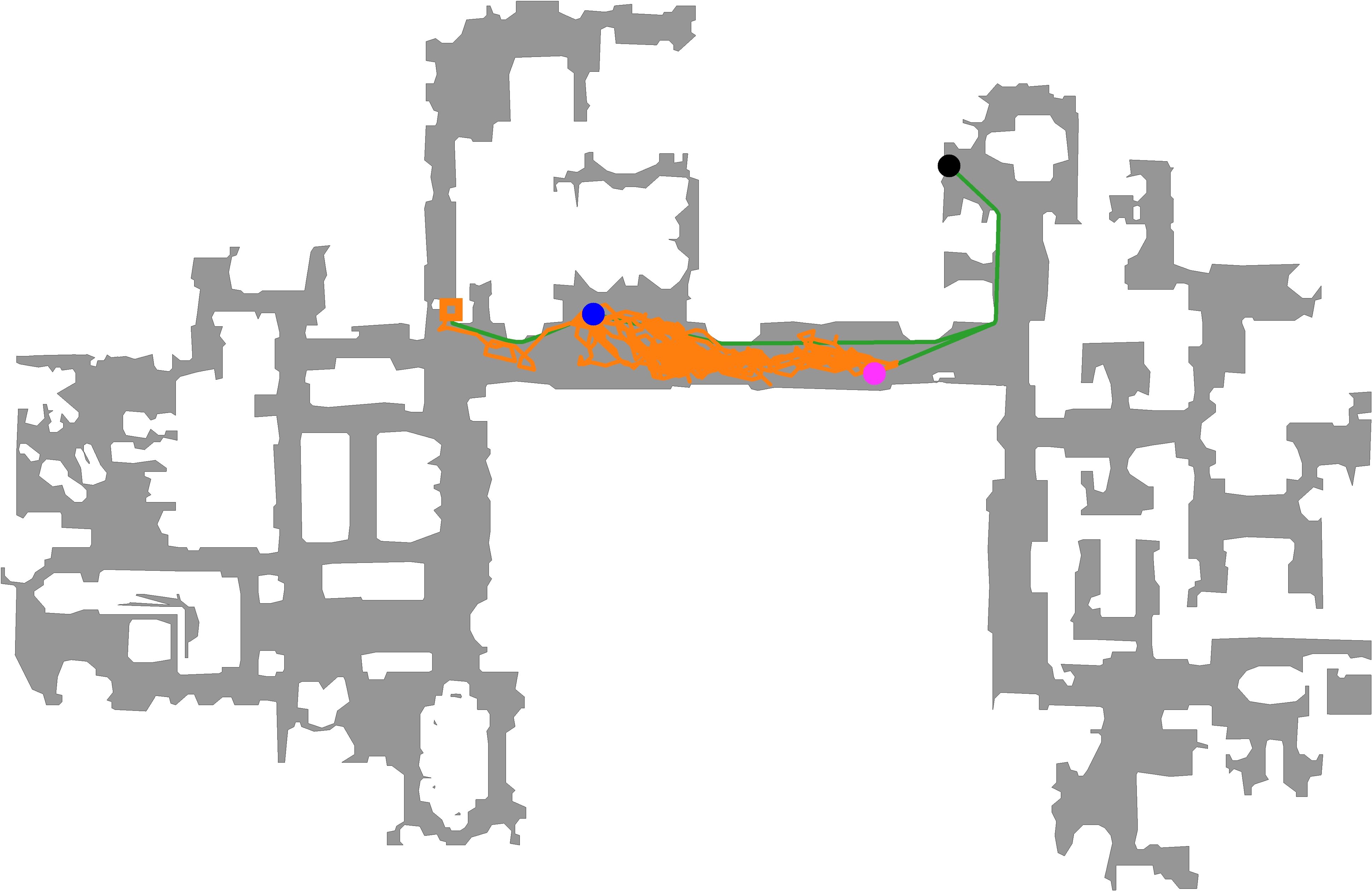}  \newline \vizc{0.33}{0.06}     &    \includegraphics[trim={35cm 45cm 32cm 10cm},clip,width=1\linewidth]{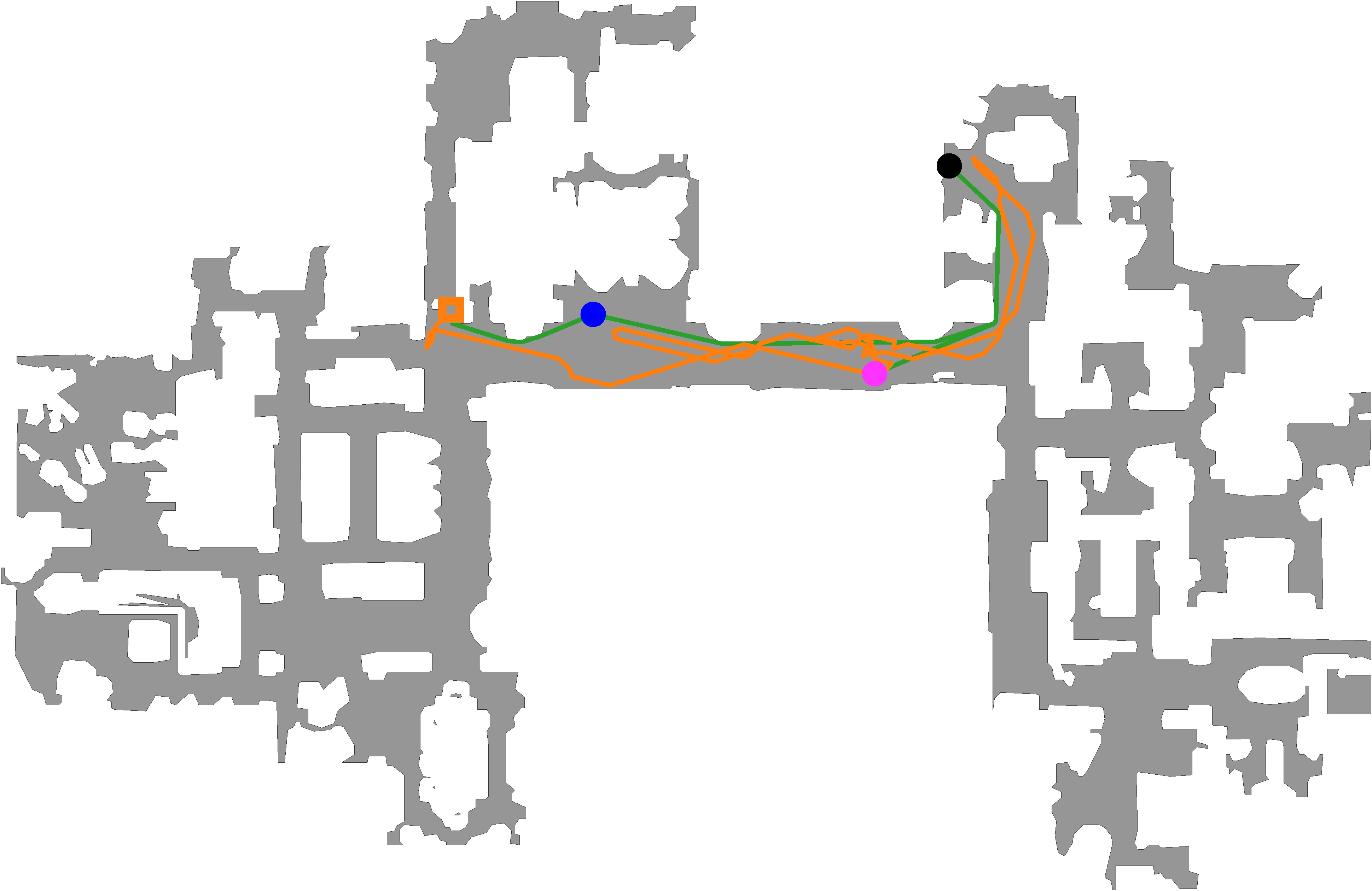}     \newline \vizc{1}{0.53}     & \includegraphics[trim={35cm 45cm 32cm 10cm},clip,width=1\linewidth]{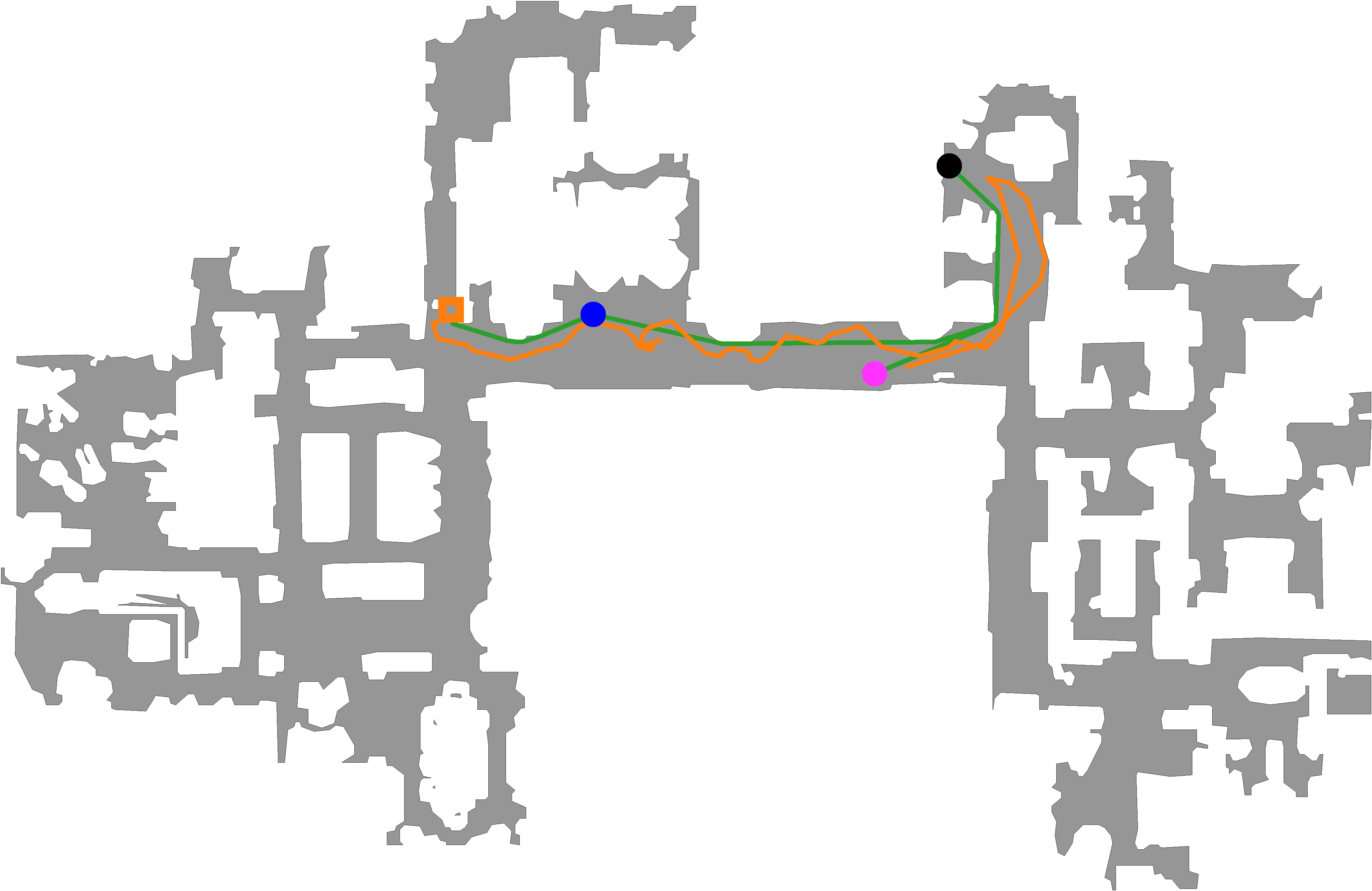} \newline \vizc{1}{0.87}        &   \includegraphics[trim={35cm 45cm 32cm 10cm},clip,width=1\linewidth]{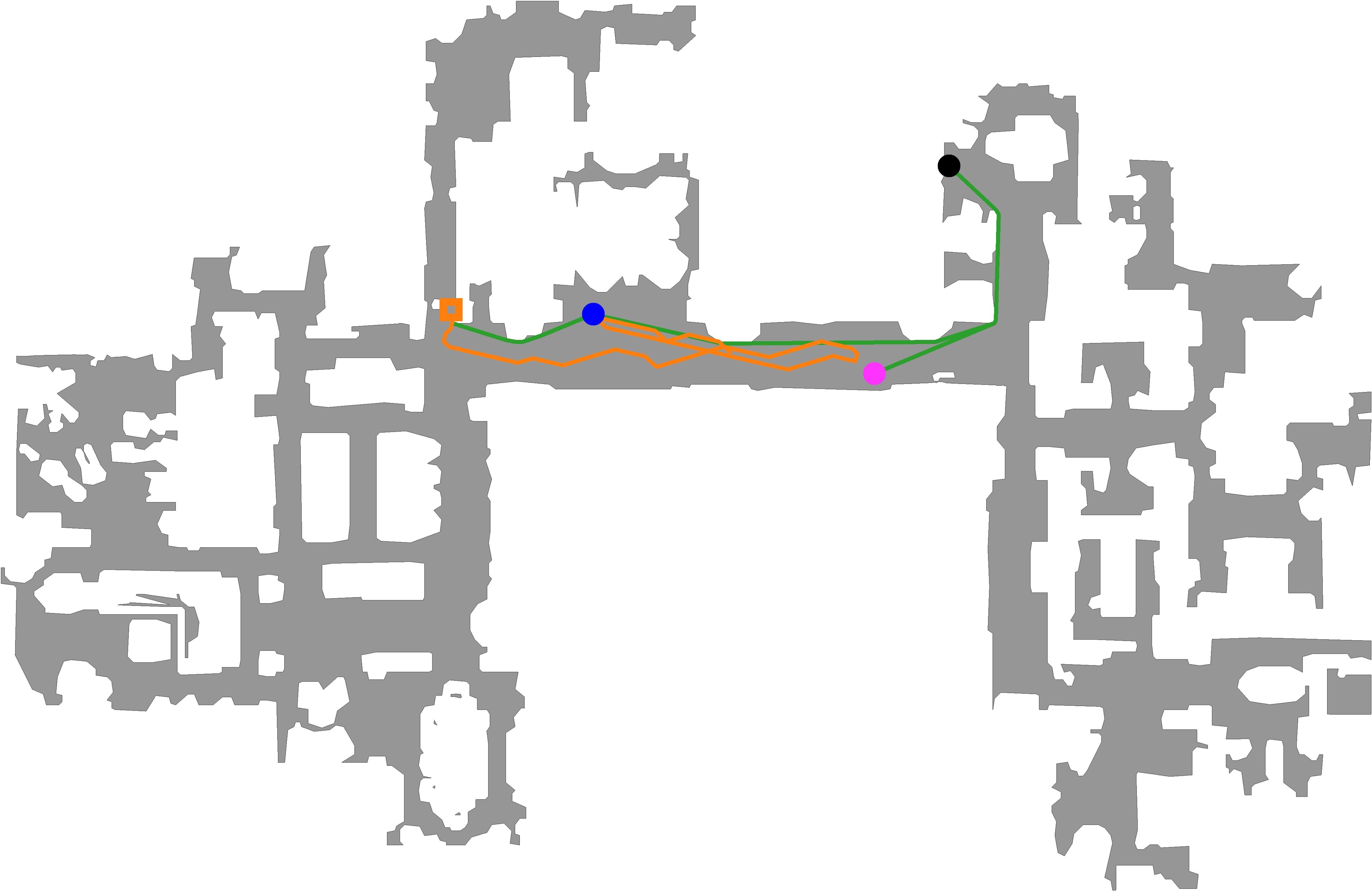}  \newline \vizc{0.33}{0.07}     &   \includegraphics[trim={35cm 45cm 32cm 10cm},clip,width=1\linewidth]{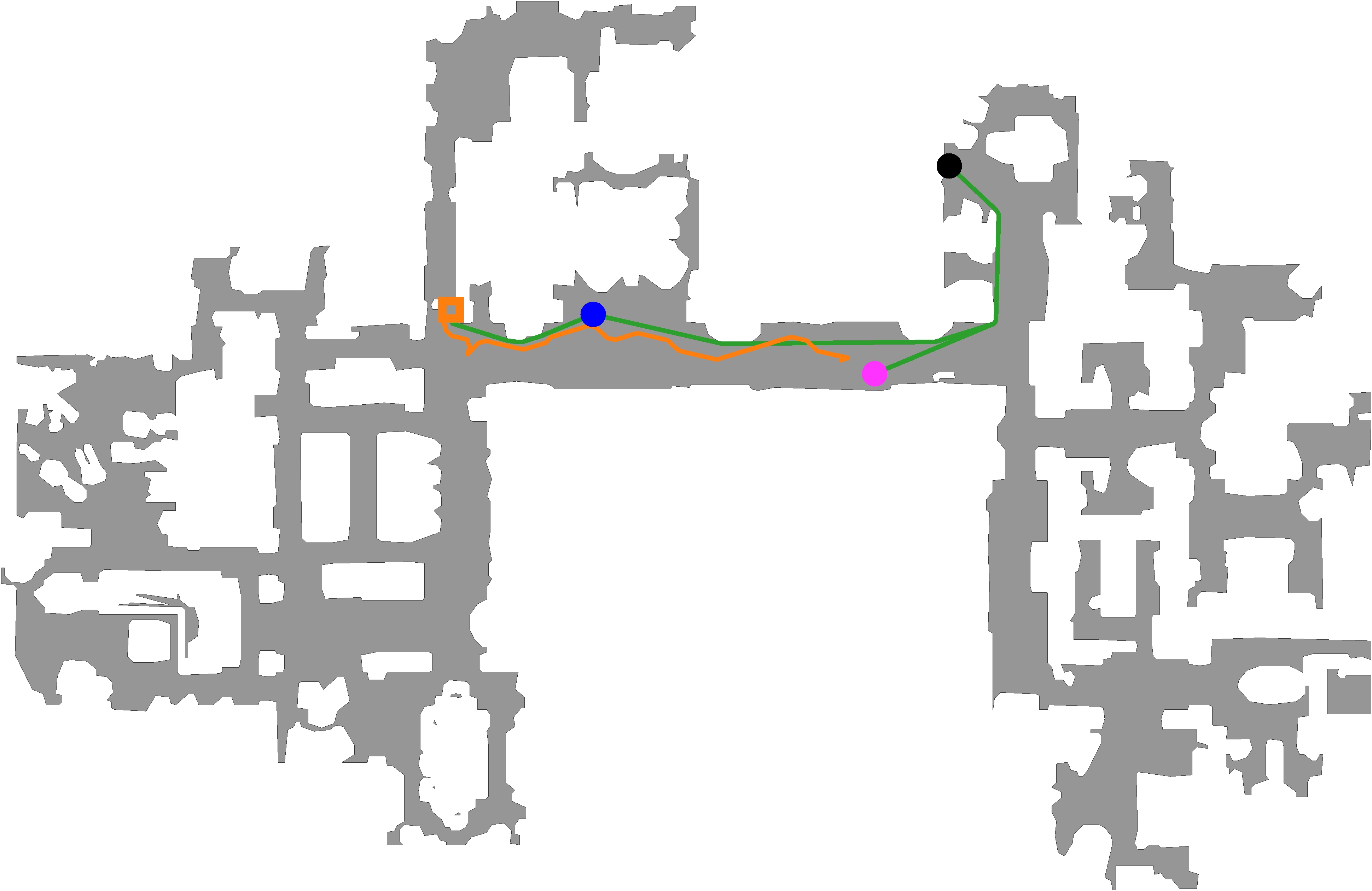}  \newline \vizc{0}{0}    \\ \\
         
     \multicolumn{5}{c}{Goal order: 1\goalb{white}{black}, 2\goal{black}, 3\goal{blue}} \\
         \includegraphics[trim={10cm 32cm 70cm 0cm},clip,width=1\linewidth]{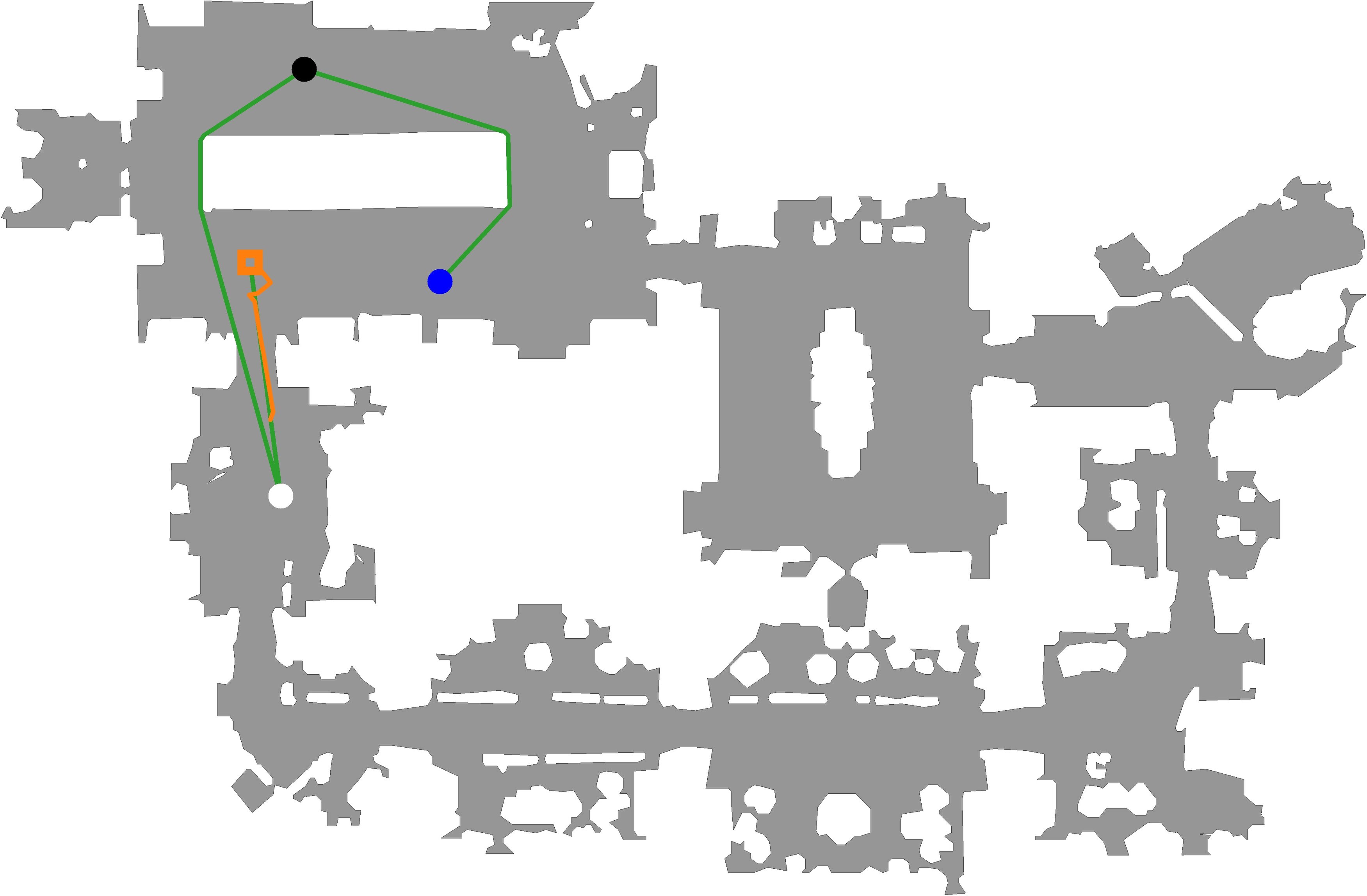} \newline \vizc{0}{0}    &    \includegraphics[trim={10cm 32cm 70cm 0cm},clip,width=1\linewidth]{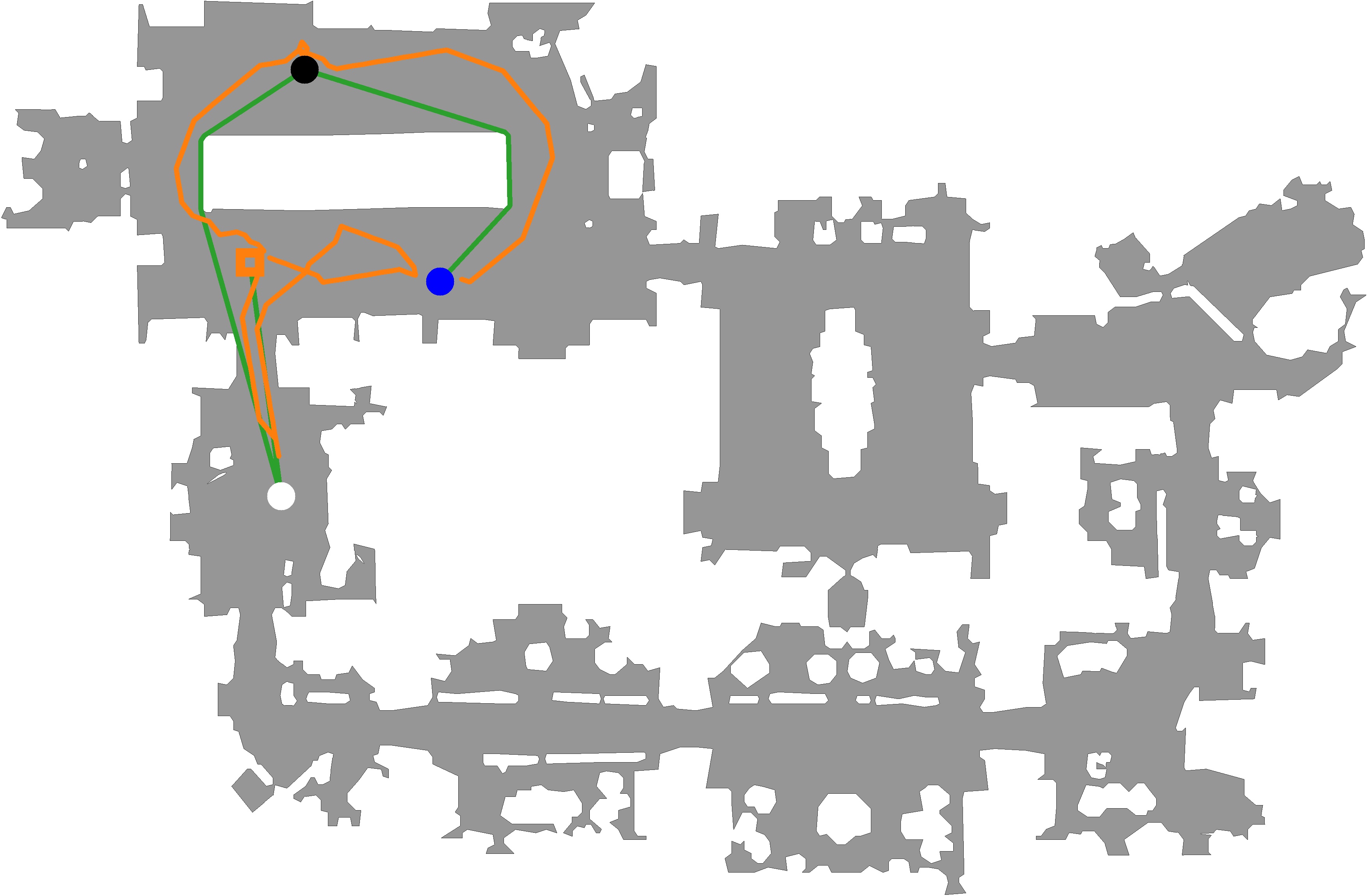} \newline \vizc{1}{0.71}          & \includegraphics[trim={10cm 32cm 70cm 0cm},clip,width=1\linewidth]{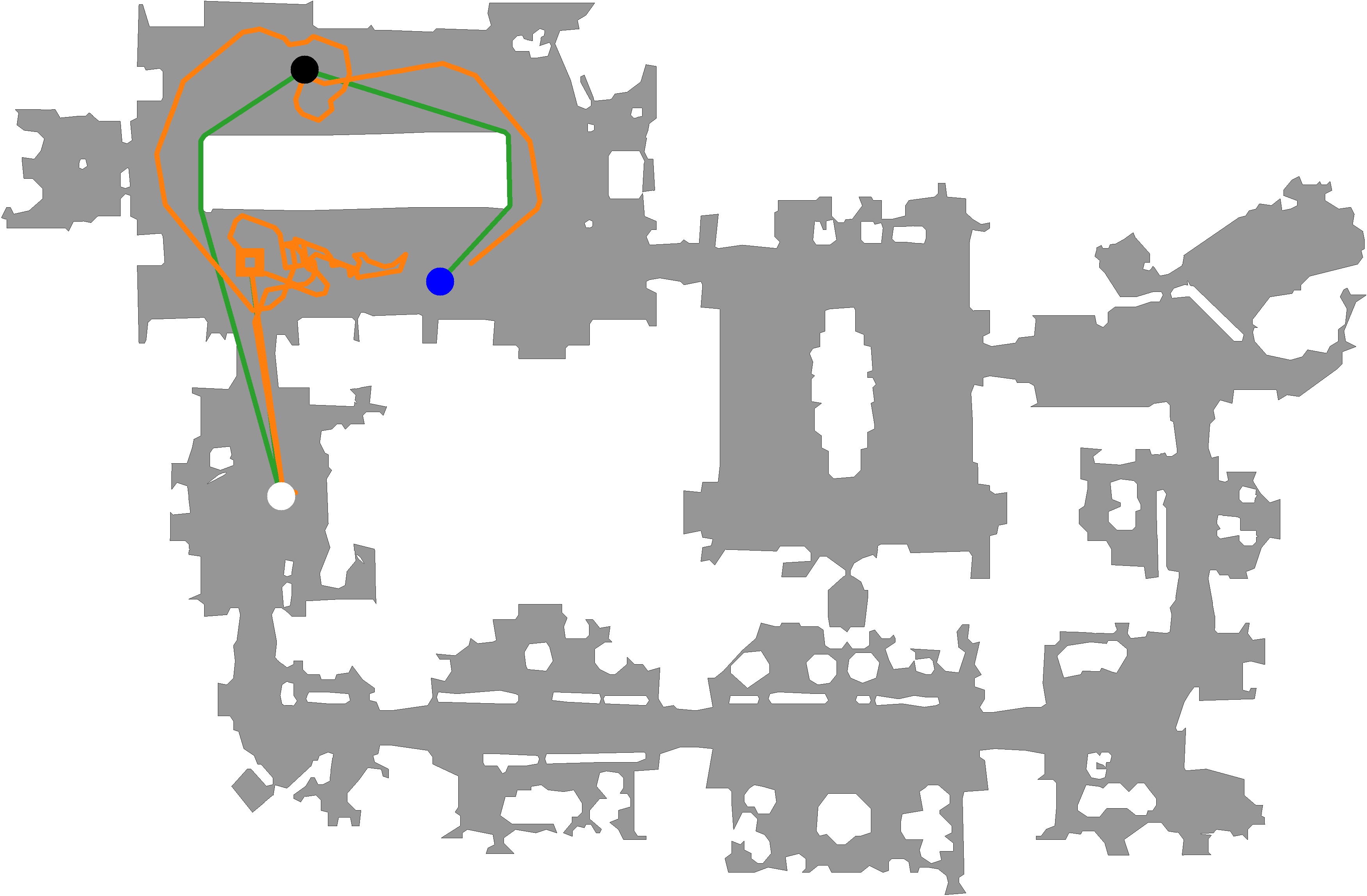}  \newline \vizc{1}{0.49}          &   \includegraphics[trim={10cm 32cm 70cm 0cm},clip,width=1\linewidth]{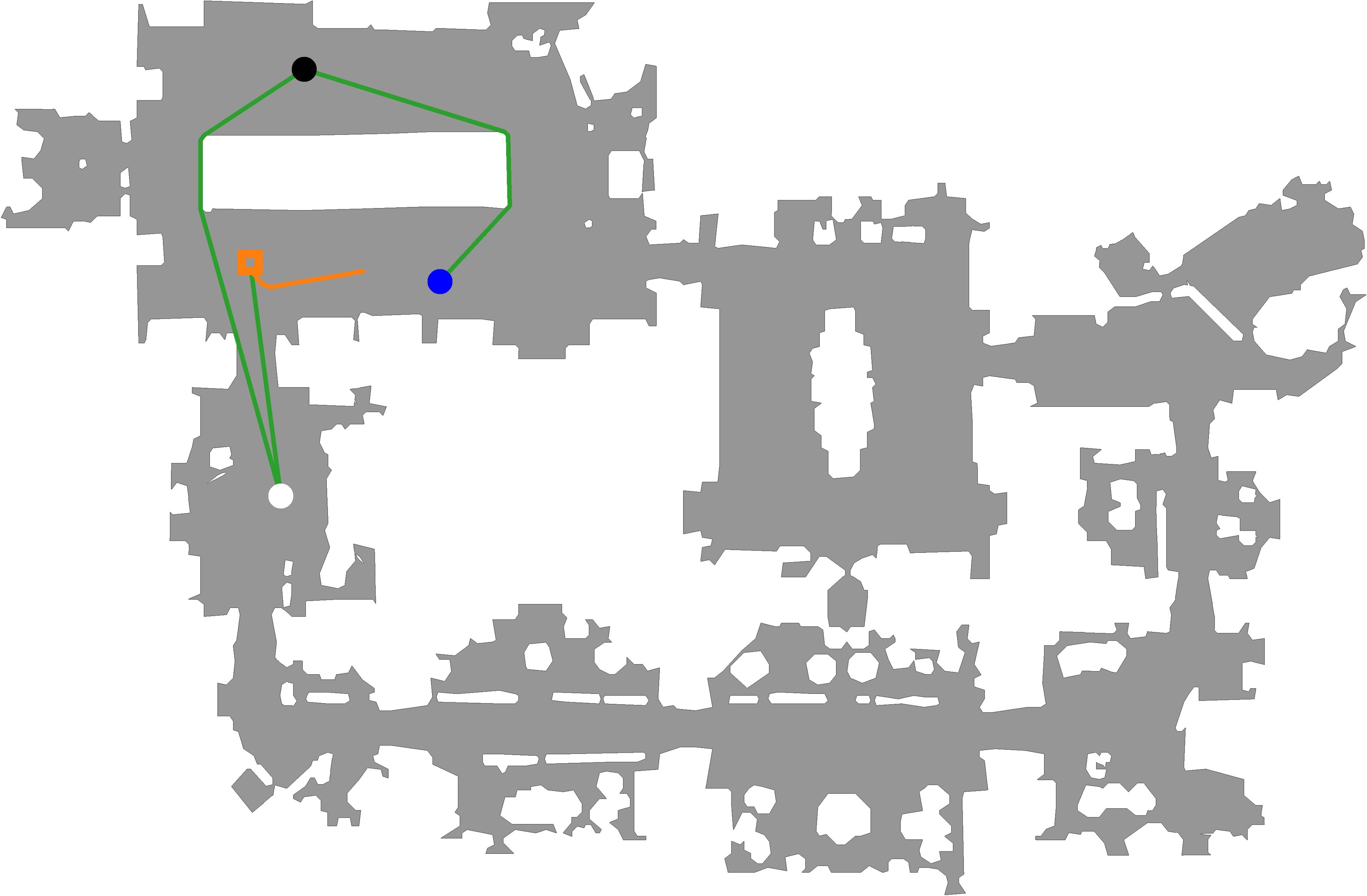} \newline \vizc{0}{0}    &   \includegraphics[trim={10cm 32cm 70cm 0cm},clip,width=1\linewidth]{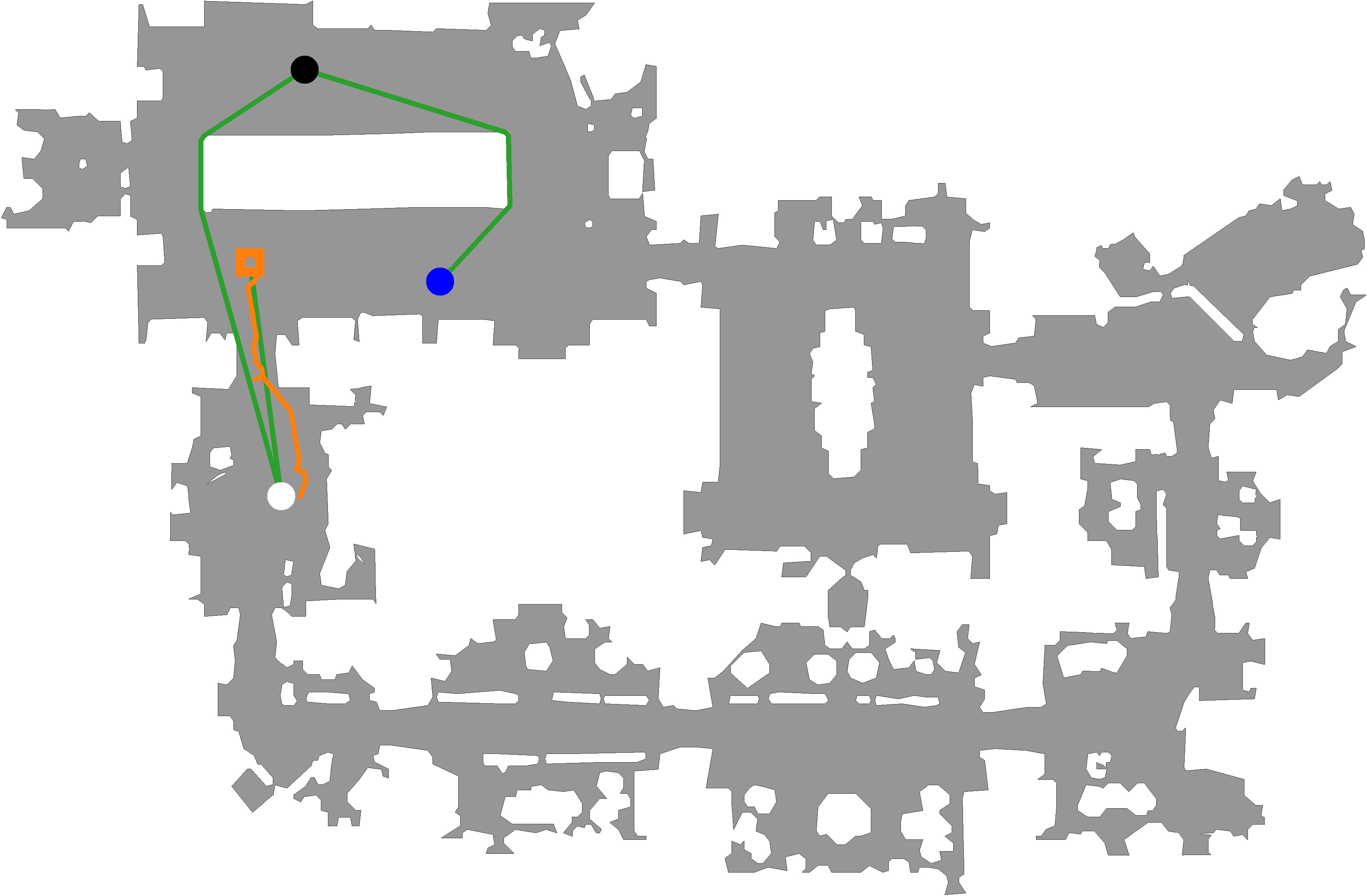} \newline \vizc{0.33}{0.28}      \\ \\
         
    \multicolumn{5}{c}{Goal order: 1\goal{red}, 2\goal{black}, 3\goal{green}} \\
        \includegraphics[width=1\linewidth]{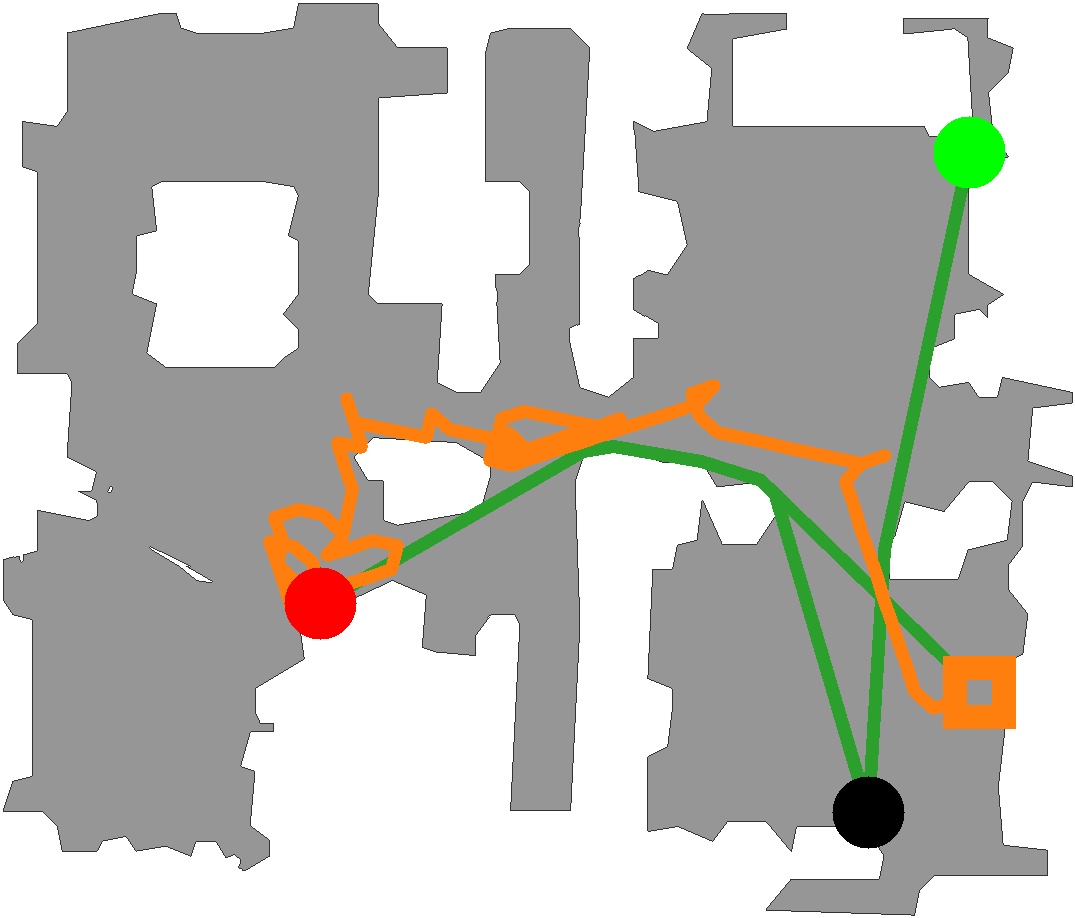} \newline \vizc{0.33}{0.13}    &    \includegraphics[width=1\linewidth]{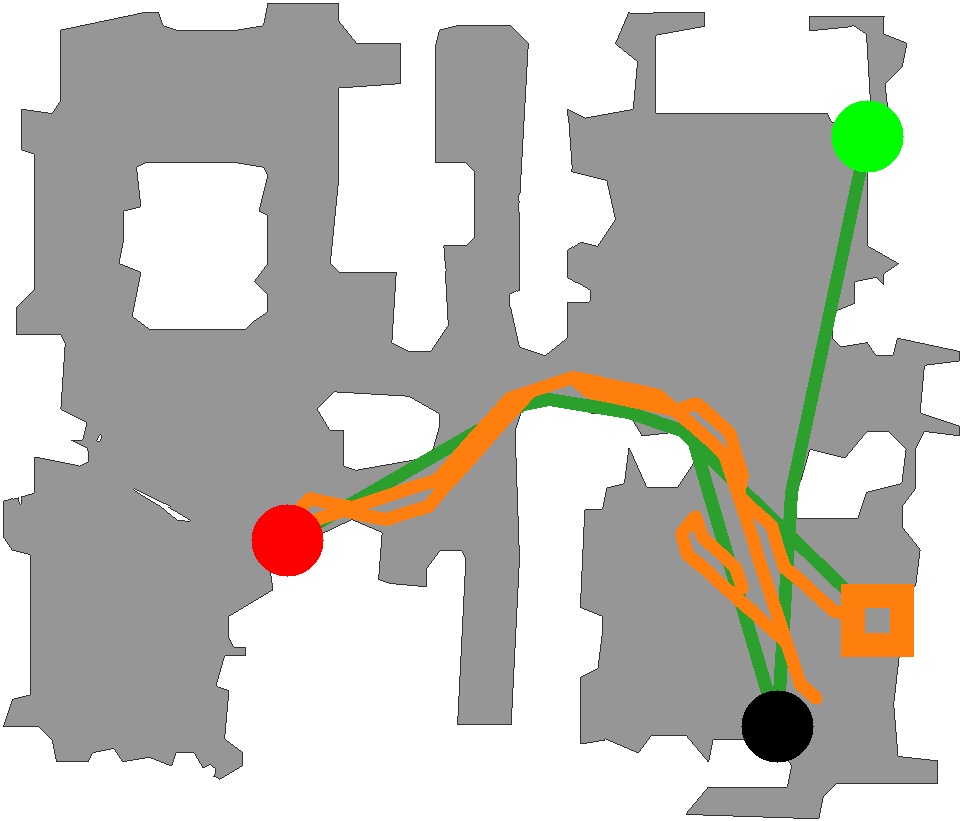} \newline \vizc{0.33}{0.30}           & \includegraphics[width=1\linewidth]{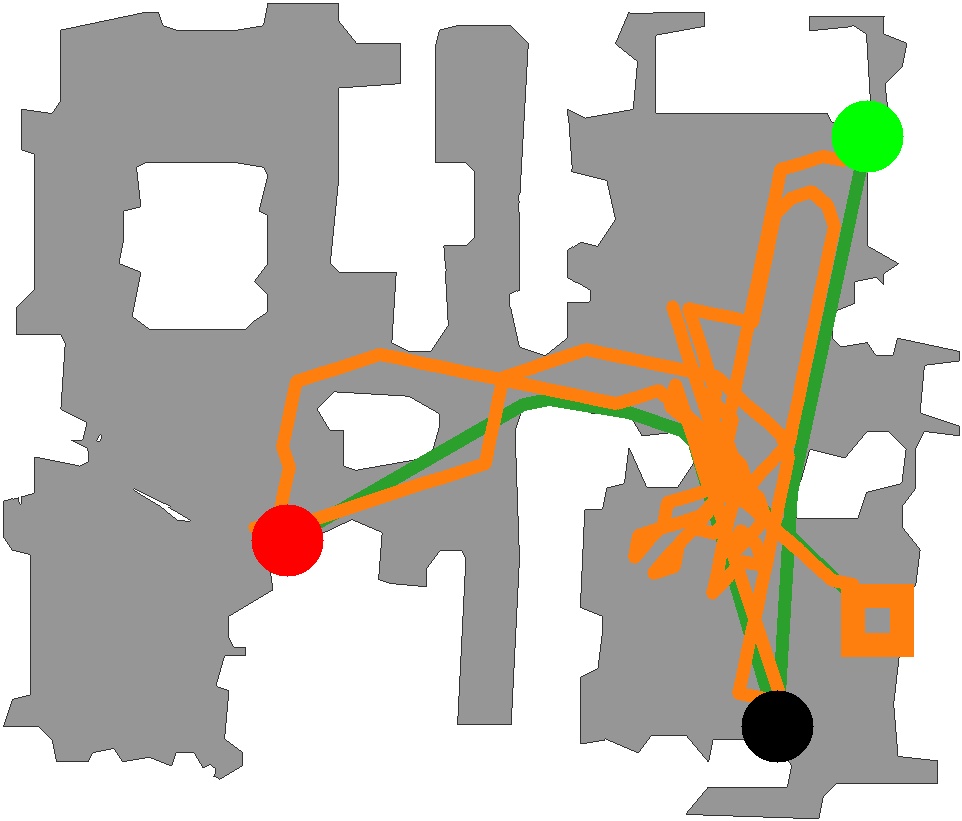} \newline \vizc{1}{0.44}         &   \includegraphics[width=1\linewidth]{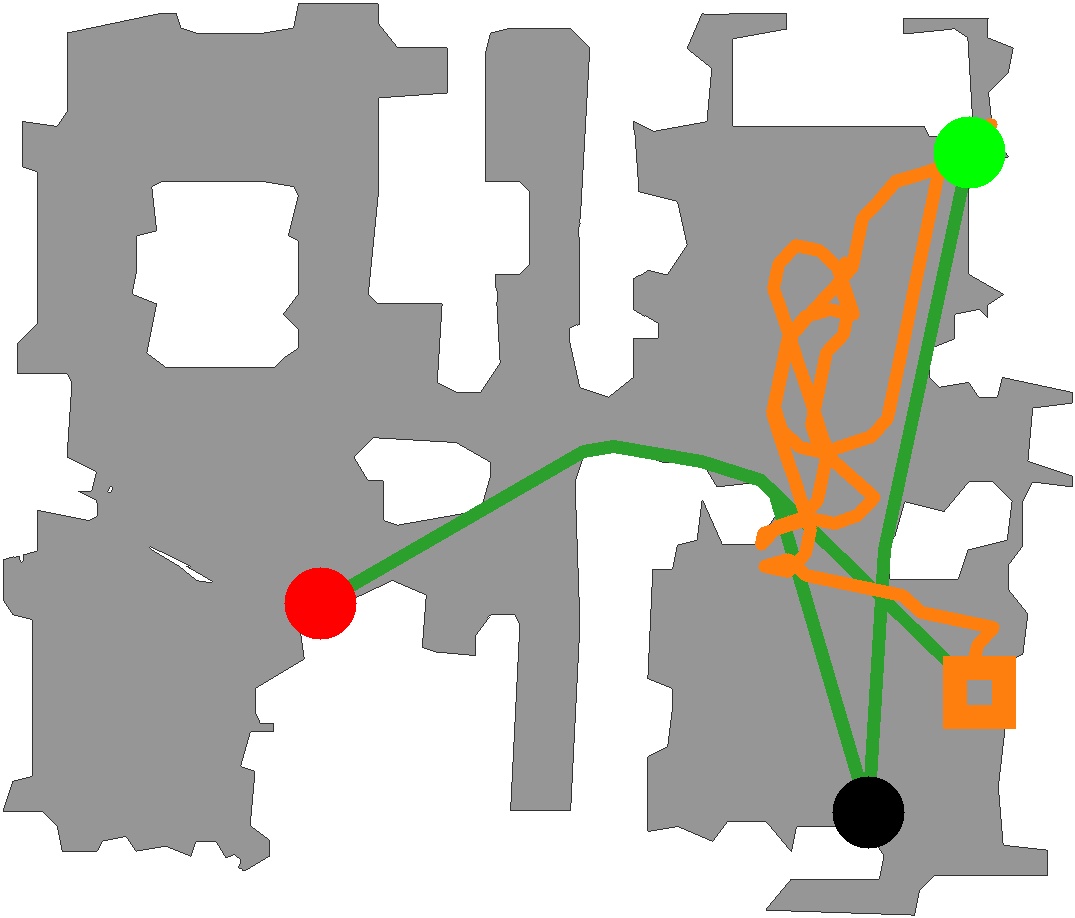} \newline \vizc{0}{0}    &   \includegraphics[width=1\linewidth]{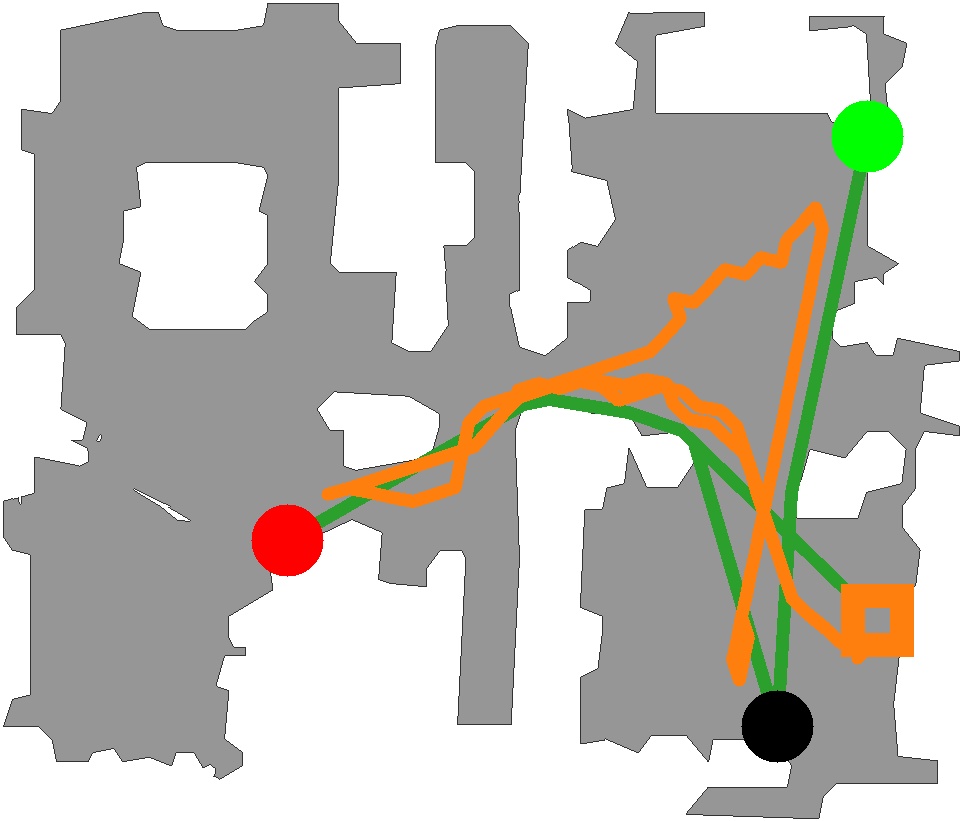}  \newline \vizc{0.66}{0.51}    \\ \\

    \multicolumn{5}{c}{Goal order: 1\goal{red}, 2\goal{magentagoal}, 3\goal{black}} \\
        \includegraphics[width=1\linewidth]{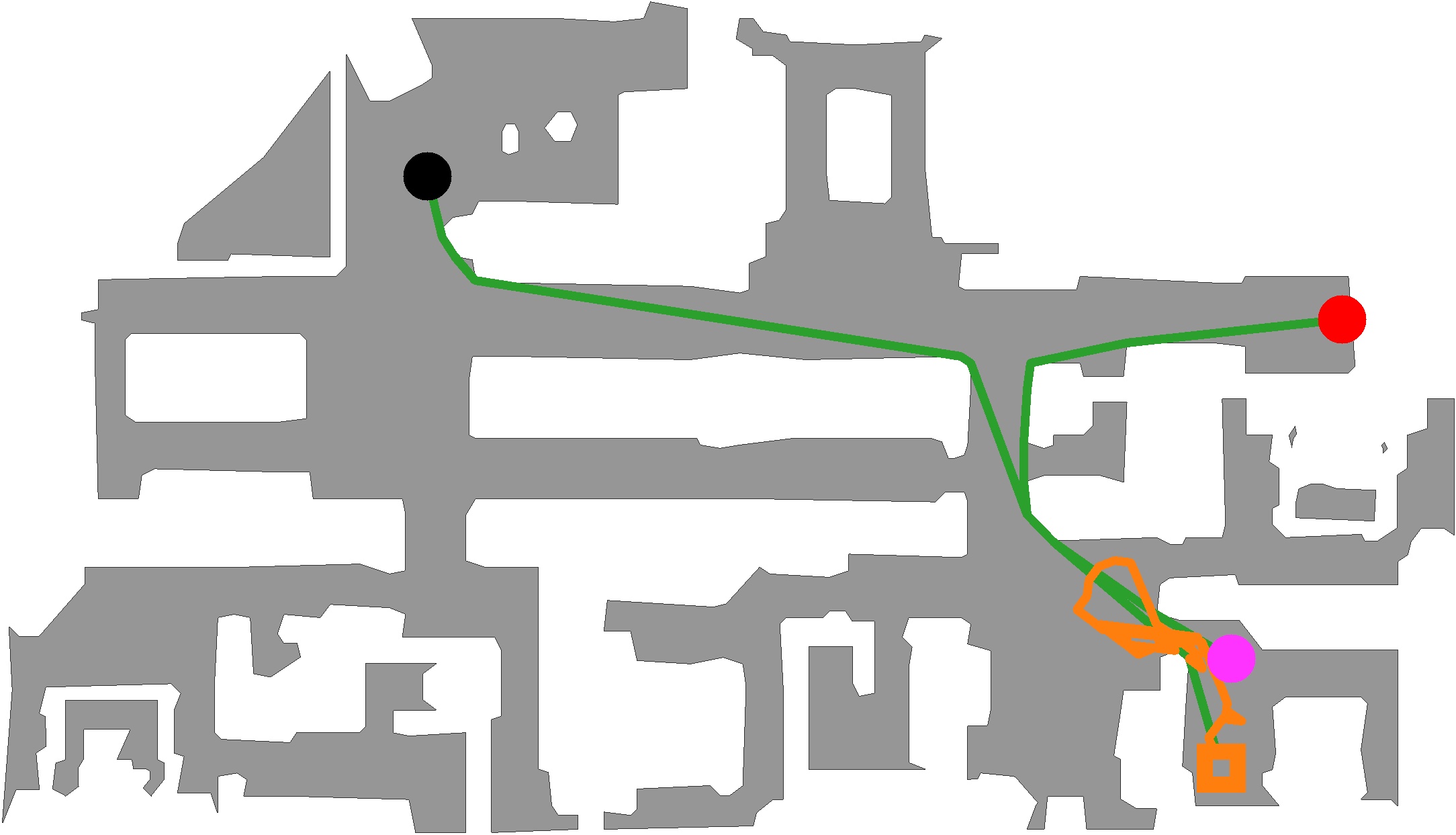} \newline \vizc{0}{0}    &    \includegraphics[width=1\linewidth]{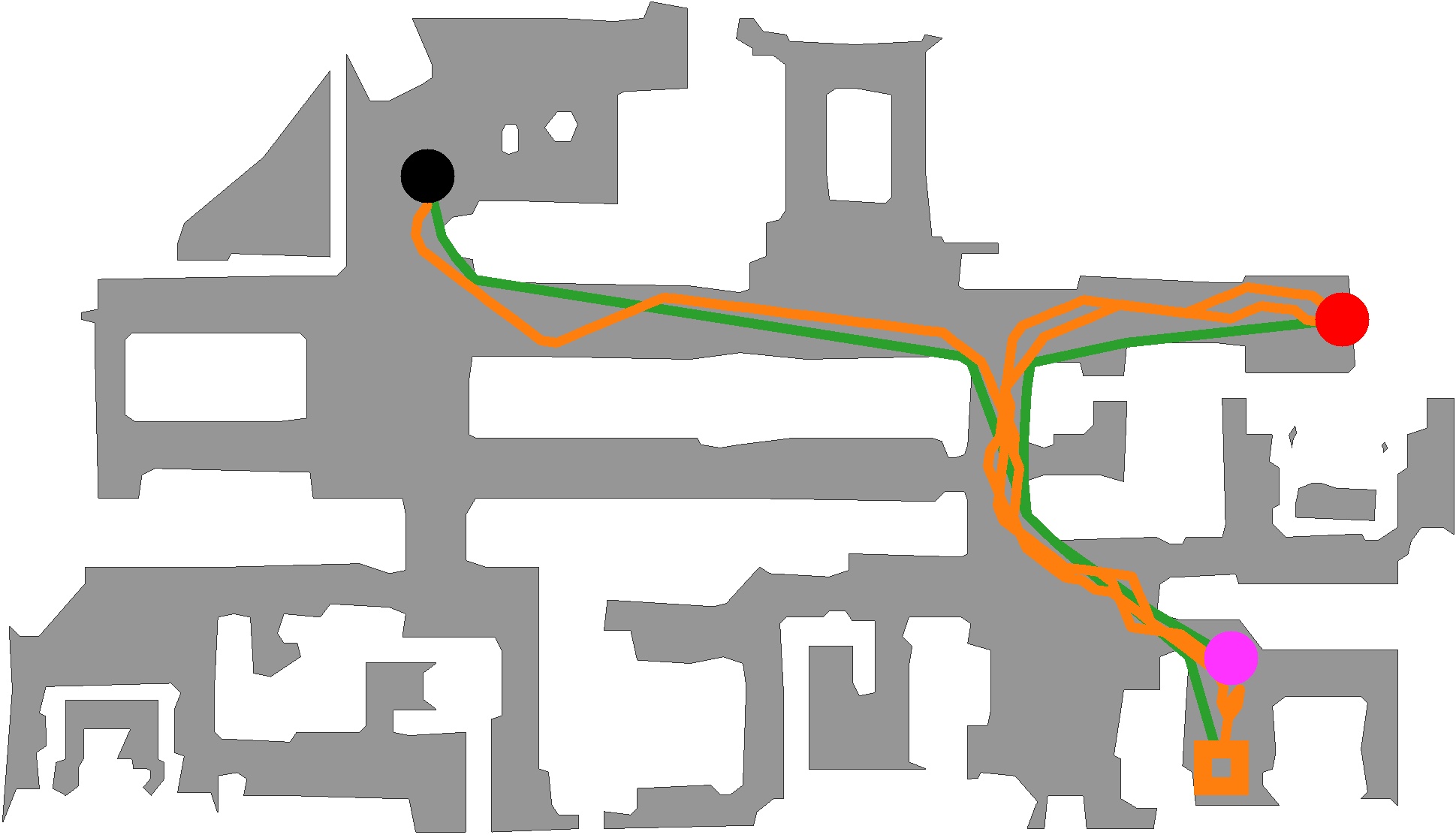} \newline \vizc{1}{0.91}           & \includegraphics[width=1\linewidth]{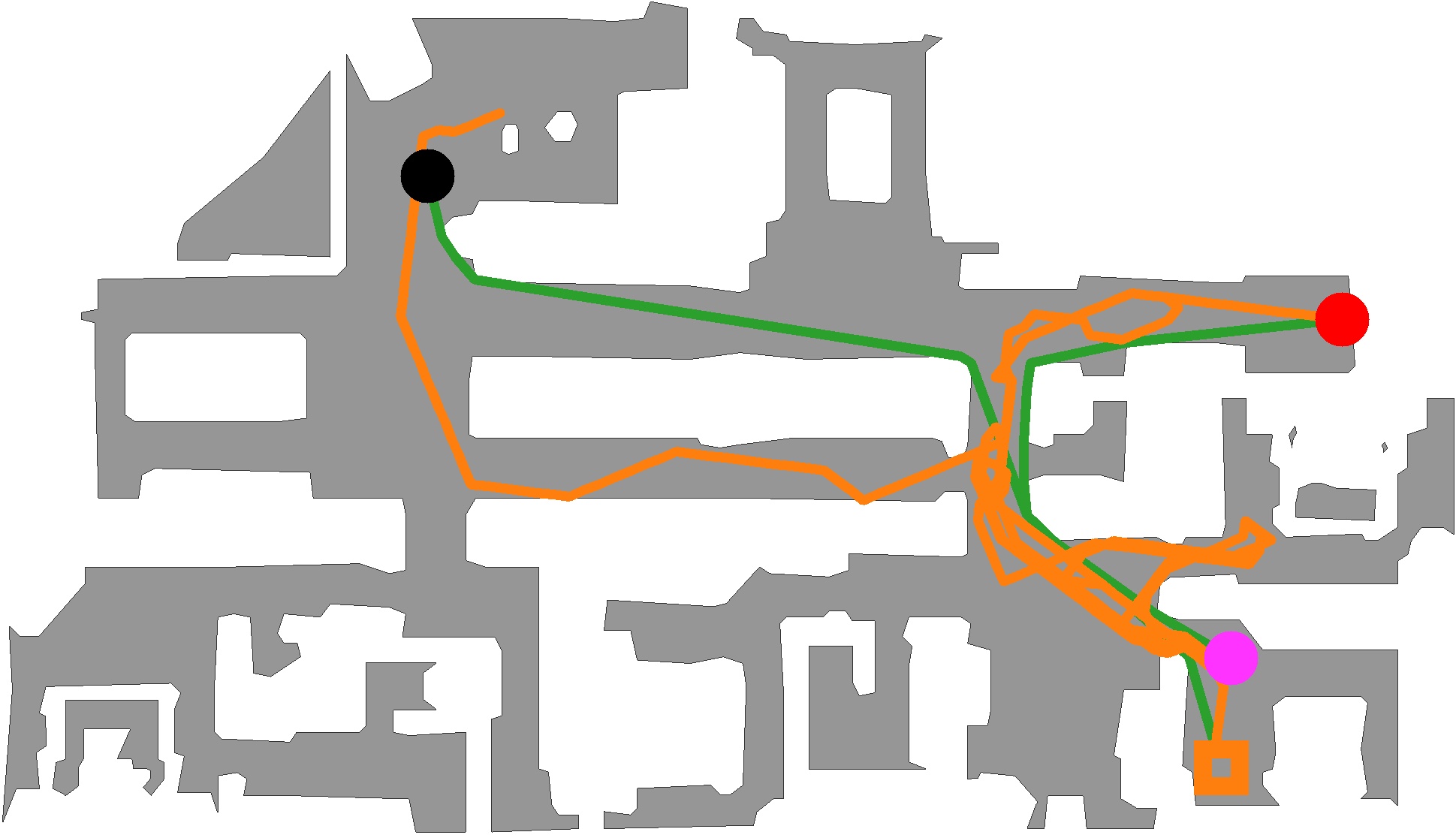} \newline \vizc{1}{0.53}         &   \includegraphics[width=1\linewidth]{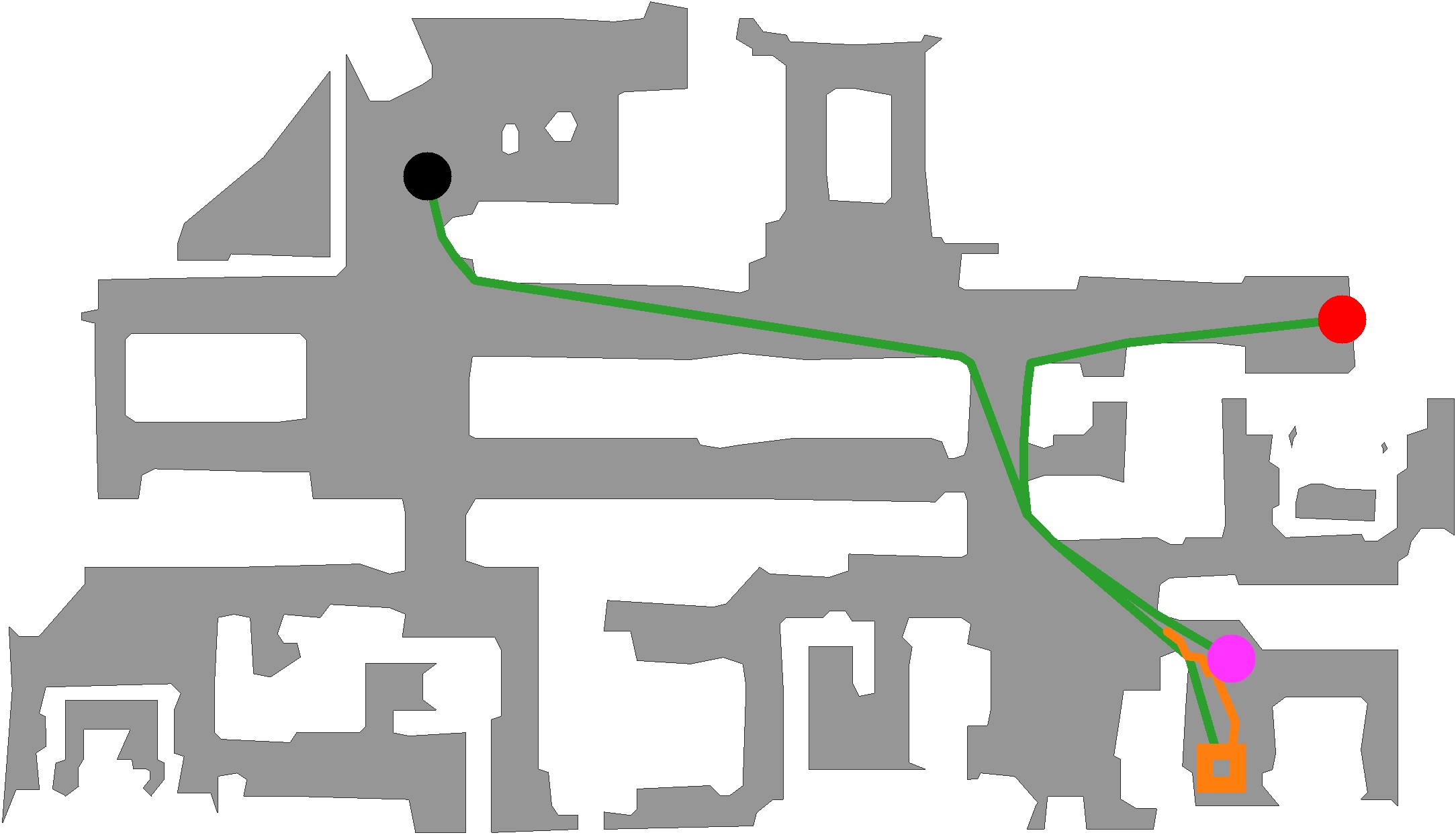} \newline \vizc{0}{0}    &   \includegraphics[width=1\linewidth]{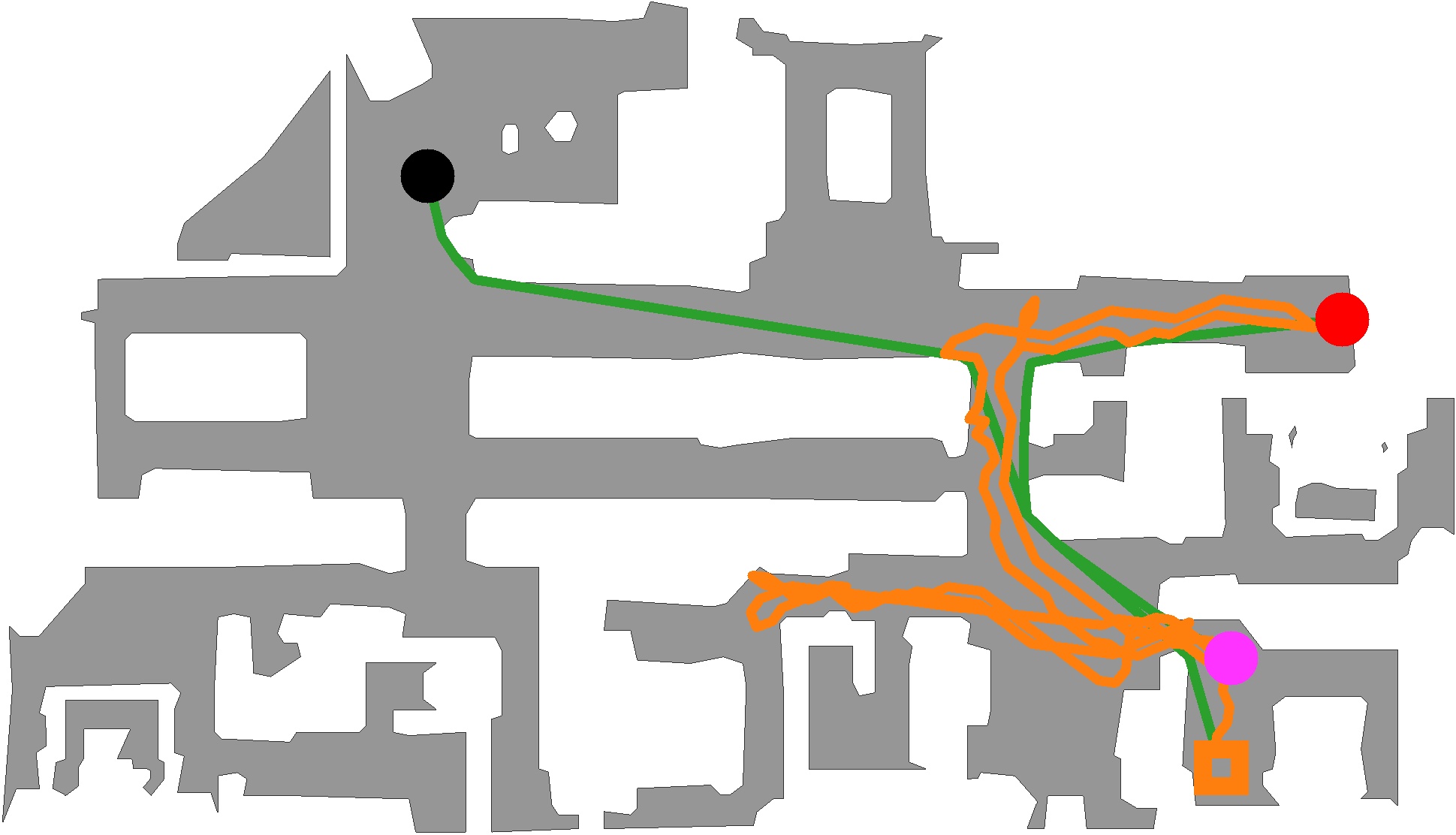}  \newline \vizc{0.66}{0.53} \\ \\
        
    \multicolumn{5}{c}{Goal order: 1\goal{red}, 2\goal{yellow}, 3\goalb{white}{black}} \\
        \includegraphics[width=1\linewidth]{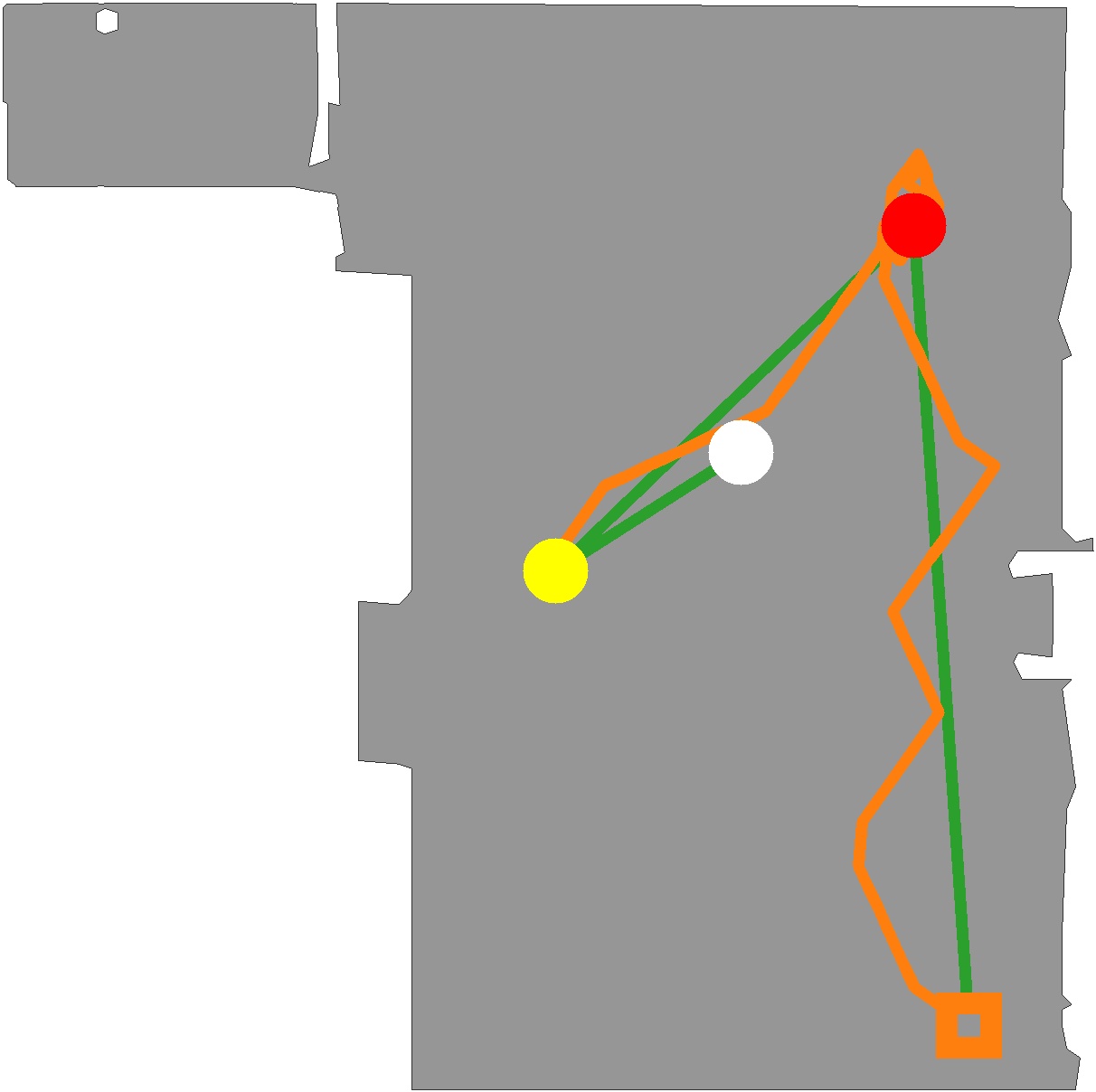} \newline \vizc{0.66}{0.48}    &    \includegraphics[width=1\linewidth]{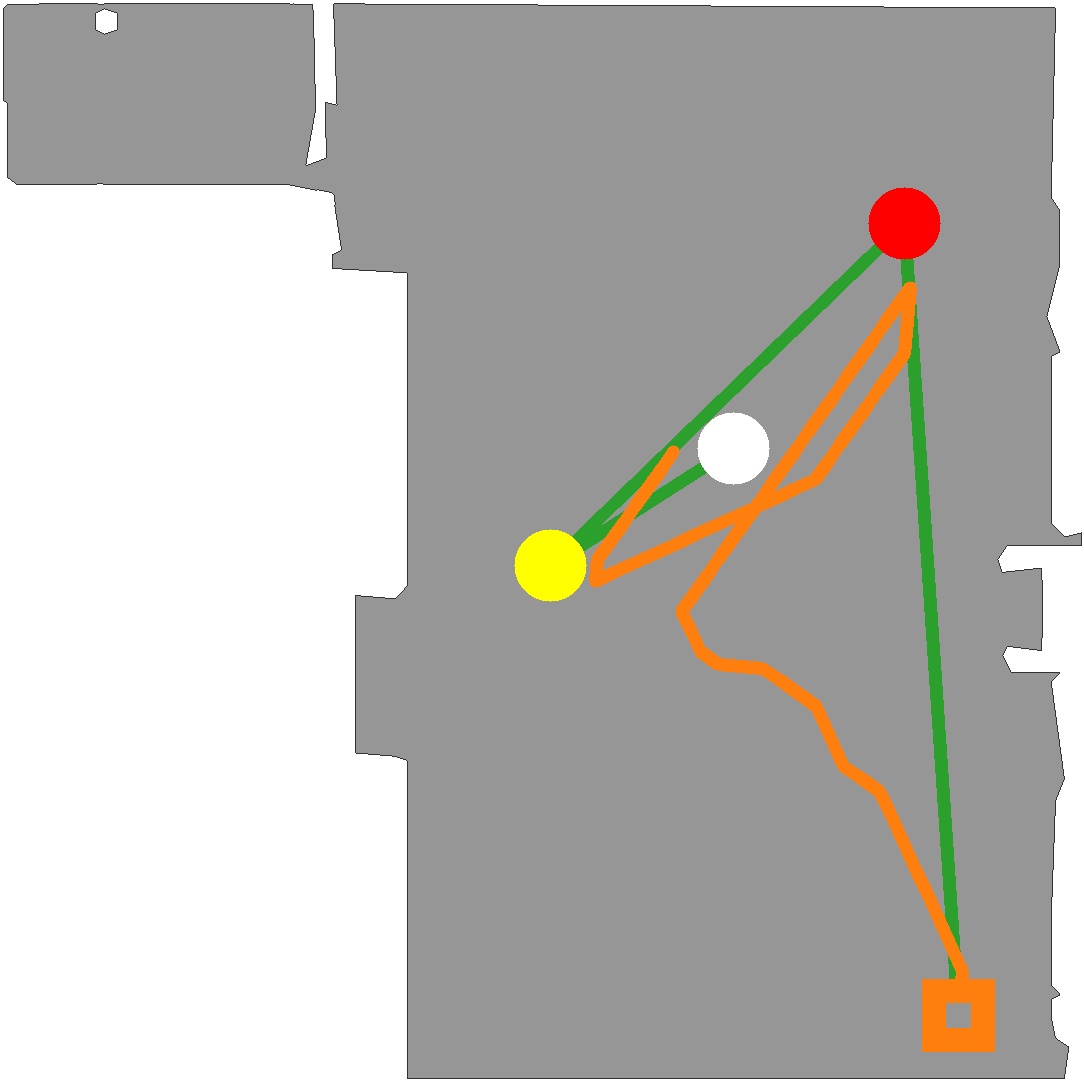} \newline \vizc{1}{0.97}           & \includegraphics[width=1\linewidth]{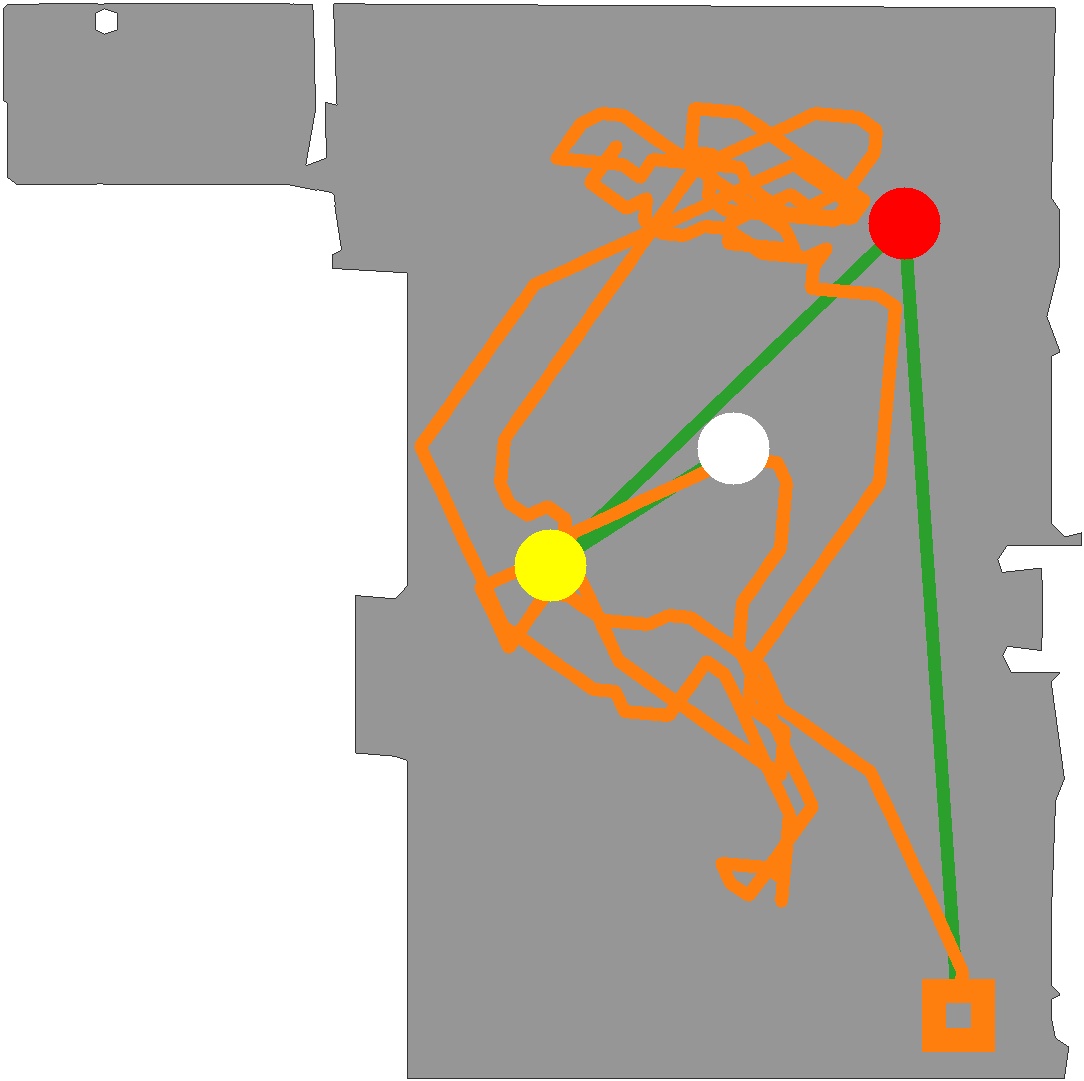} \newline \vizc{0.66}{0.17}         &   \includegraphics[width=1\linewidth]{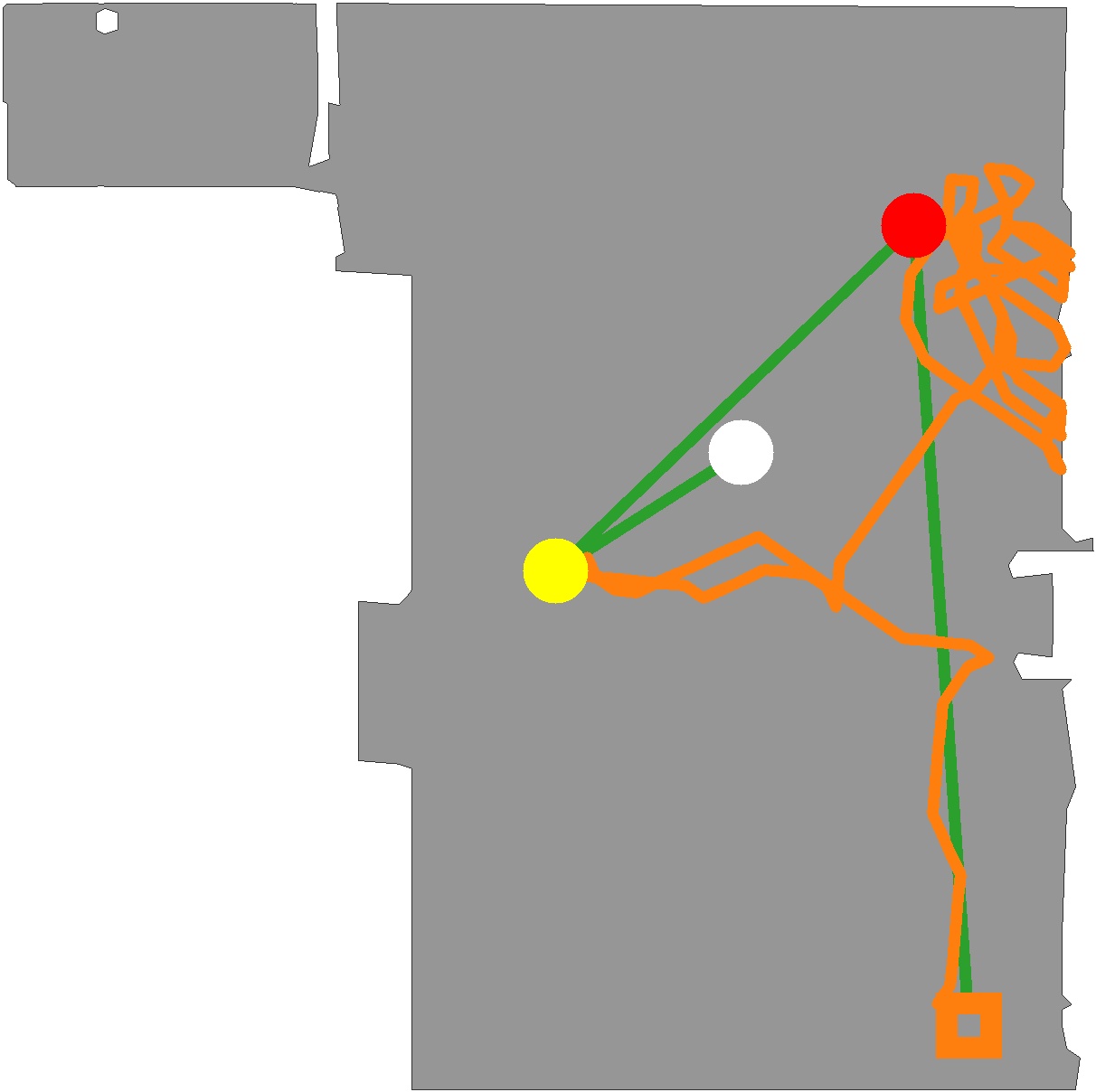} \newline \vizc{0}{0}    &   \includegraphics[width=1\linewidth]{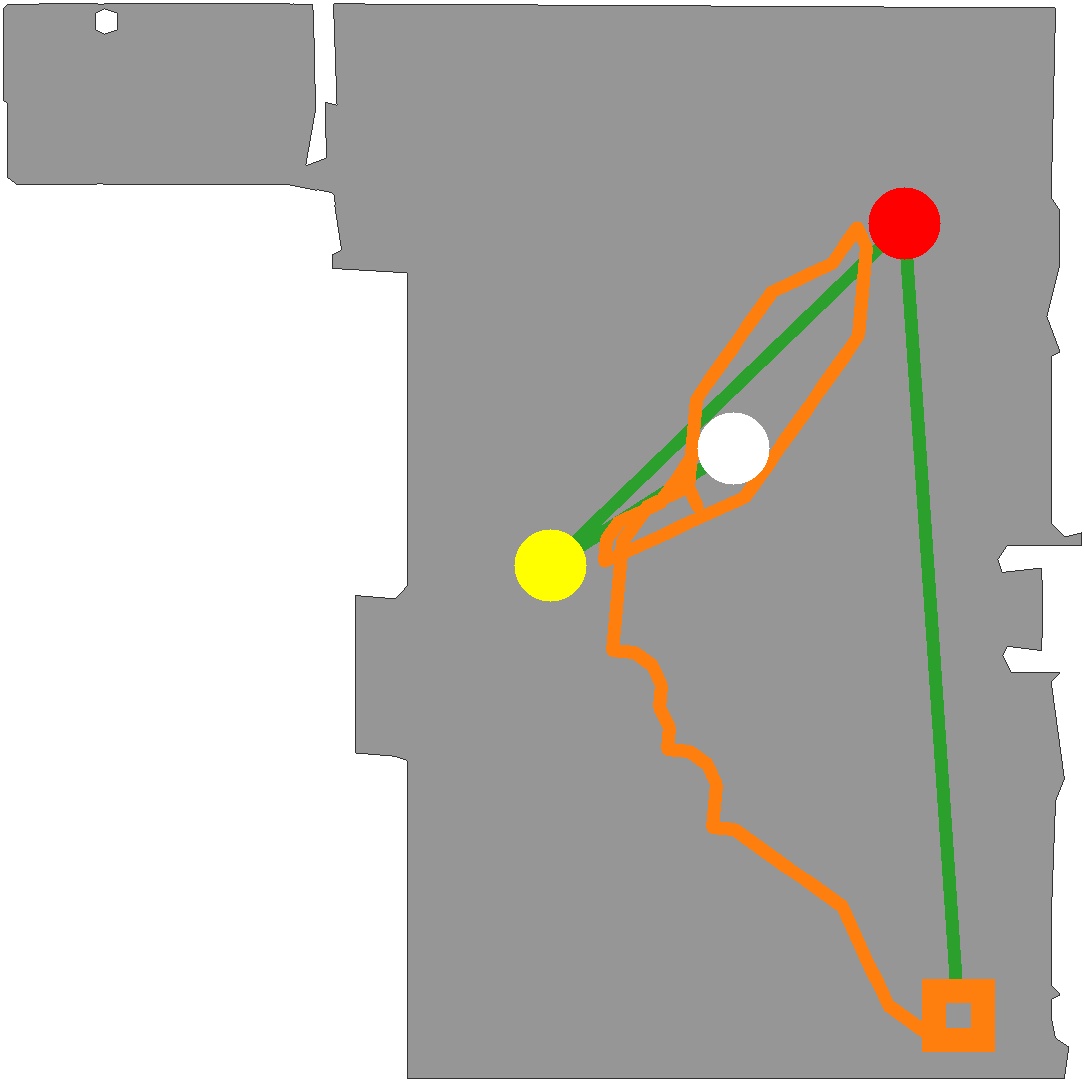}  \newline \vizc{1}{0.84}    \\ \\

\end{tabular}
}

\caption{Additional example episodes for different agents.
{\color{orange}Agent path} and {\color{cadmiumgreen}shortest path} in orange and green, with the start shown by {\startsq} (orange square).
}
\label{fig:more-episode-maps}
\end{figure}

\xhdr{Additional episode visualizations.}
\Cref{fig:more-episode-maps} shows additional visualizations of test set episodes.
We see that the episodes span a range of environments, with fairly complex paths between goals that frequently require some degree of backtracking.
As we saw in the analysis of agent performance when goals are previously seen vs not, such episodes requiring backtracking help us to benchmark the ability of the agents to store and use information on previously seen goals.

\clearpage
\newpage

\begin{table}
\ra{1.3}
\centering
\caption{Test set performance for ablations of the \OracleMap model. \DyMap is dynamically updated to only indicate the current goal object instead of all goals. \OnlyMap shows results with only map information and no RGBD sensors. The \NoObject is the same as \OracleMap except the goal objects are not inserted into the environment so the agent must navigate to the target locations based on the map information alone.
}
\label{tab:onlymap}
\resizebox{0.7\linewidth}{!}{
\begin{tabular}{@{}l rrr @{\hspace{7mm}} rrr @{}}
\toprule
& \multicolumn{3}{c}{\Success (\%)} & \multicolumn{3}{c}{\SPL (\%)}\\
\cmidrule(lr){2-4} \cmidrule(lr){5-7}
& \mon{1} & \mon{2} & \mon{3} & \mon{1} & \mon{2} & \mon{3} \\
\midrule
\OracleMap & $94$ & $74$ & $48$ & $77$ & $59$ & $38$ \\
\midrule
\DyMap & $94$ & $85$ & $81$ & $74$ & $71$ & $66$ \\
\OnlyMap & $54$ & $9$ & $0$ & $42$ & $7$ & $0$ \\
\NoObject & $91$ & $72$ & $40$ & $75$ & $57$ & $33$ \\
\bottomrule
\vspace{-7mm}
\end{tabular}
}
\end{table}

\xhdr{Oracle map ablations.}
We perform a series of ablation experiments to investigate how information provided in the oracle map and its connection with other sensory input during the task can determine agent performance.
The results of these experiments are in \Cref{tab:onlymap} and are summarized below.

\begin{compactitem}
  \item \emph{Dynamically updated oracle map.}
  Instead of storing the embedding of all target objects in the oracle map at the beginning of the episode, we store the embedding of only the current target object and dynamically update the map when the agent finds a target object.
  This helps simplify training since the agent does not have to distinguish between the embeddings of various objects stored in the map (only the current object embedding is stored).
  This effectively breaks down an \mon{m} task into $m$ ObjectNav (\mon{1}) tasks.
  As expected, the \Success rate in an \mon{m} task is approximately $s^m$, where $s$ is the \Success rate of \mon{1}.
  Note that this is significantly better than the \Success rate in the original setup where all the object embeddings are stored in the oracle map at the start of the episode.
  \item \emph{Map-only baseline.}
  We experiment with agents that do not have RGBD sensors and must only use the oracle map to navigate to their target locations in an \mon{m} task (\OnlyMap).
  Although the agent has access to the locations of all the target objects in the environment (in the form of embeddings stored in the map), it fails to perform as well as the agent using RGBD sensors.
  The \Success rate drops rapidly from \mon{1} to \mon{2} to \mon{3}.
  This could be due to two reasons: 1) the agent is unable to recognize the target object and call \found when it is near the object, or 2) the agent is unable to navigate (avoid obstacles, walk through hallways etc.) effectively through the environment.
  Further experiments and qualitative analysis of rollouts from validation episodes reveal that the decreased performance is primarily due to the inability of the agent to navigate effectively through the scenes.
  \item \emph{Hidden goal objects.}
  In this variant (\NoObject), the agent is provided an oracle map and has access to RGBD sensory information but the objects are not actually inserted in the environment.
  The agent must therefore infer the location of the target objects from the oracle map.
  The RGBD sensors are expected to help learn basic navigation skills like obstacle avoidance.
  As is evident from \Cref{tab:onlymap}, the model performs just as well as the one where objects are inserted in the environment.
  This also shows that the fall in performance in the \OnlyMap baseline can largely be attributed to a lack of basic navigation skills rather than an inability to call \found action at the right time.
  \item \emph{Varying map resolution.}
  We experimented with different map resolutions and sizes of oracle maps during validation.
  Final results are given with the resolution settings that we found to perform best.
  The resolution of the map refers to the physical length of a single grid cell in the environment.
  The size of the map refers to the the area of the global map that is cropped (and rotated) for generating the egocentric map $m_t$.
  Similarly to \citet{chen2018learning,chaplot2020learning}, we experimented with a variant where we stack both high and low resolution maps together.
  Here, the map consists of $4$ channels, $2$ each for the high and low resolution maps.
  The two maps correspond to different area sizes in the environment and therefore provide the agent with more local or more global views of the scene.
  We did not find any improvements in agent performance over the base \OracleMap model.
\end{compactitem}

\textbf{Dynamically updating maps}.
Similar to ablations with dynamically updated oracle maps, we experiment with variants of other models in which only the embedding of the current target object is stored in the object channel of the map.
This is not possible with \ProjNeuralMap since we do not store object embeddings.
In the \texttt{DynamicOracleEgoMap}, if the current goal has been discovered its embedding is stored in the map.
If not, the object channel of the map does not store anything.
In the \texttt{DynamicObjRecogMap}, if the current goal has been discovered (and correctly identified through object classification), its embedding is stored in the map.
Otherwise, the object channel does not store any information.
\Cref{tab:dynamic-models} summarizes the performance metrics for these variants. 
We see that these variants do not perform better than the base models in which the embeddings of all target objects discovered so far are stored in the map.

\begin{table}
\ra{1.3}
\centering
\caption{Test set performance for ablations of the \OracleEgoMap model. \texttt{DynamicOracleEgoMap} is dynamically updated to only indicate the current goal object (if it has been seen yet) instead of all goals.  \texttt{DynamicObjRecogMap} is dynamically updated to only indicate the current goal object (it is has been seen and identified).
}
\label{tab:dynamic-models}
\resizebox{0.6\linewidth}{!}{
\begin{tabular}{@{}l rrr @{\hspace{7mm}} rrr @{}}
\toprule
& \multicolumn{3}{c}{\Success (\%)} & \multicolumn{3}{c}{\SPL (\%)}\\
\cmidrule(lr){2-4} \cmidrule(lr){5-7}
& \mon{1} & \mon{2} & \mon{3} & \mon{1} & \mon{2} & \mon{3} \\
\midrule
\texttt{DynamicOracleEgoMap} & $83$ & $62$ & $41$ & $65$ & $46$ & $25$ \\
\texttt{DynamicObjRecogMap} & $79$ & $52$ & $20$ & $56$ & $35$ & $16$ \\
\bottomrule
\end{tabular}
}
\end{table}

\textbf{Maximum episode length determination.}
The maximum allowed episode length for all experiments is $2@500$.
This limit is large enough that it does not affect the success rate in an episode.
To ensure that this threshold does not influence our results, we evaluated all agents with the time limit set proportional to $m$ ($2@500$ for \mon{3}, $\frac{2}{3}\times 2@500$ for \mon{2} and $\frac{1}{3}\times 2@500$ for \mon{1}).
This proportional thresholding has negligible impact on the performance metrics.
Further, the mean and median episode lengths are $276.2$ and $151$ steps respectively for \mon{3} experiments, which shows that most episodes terminate by calling a wrong \found action rather than by reaching the maximum time limit.

\xhdr{Varying target object shapes.}
We vary the shapes of the target objects, by choosing from $8$ different objects: cone, cube, cylinder, frustum, joined inverted frustums, sphere, tetrahedron and torus.
All these objects have a horizontal length/diameter of $0.4m$.
They have the same red color and are all placed on the top of a thin cylindrical support.
\Cref{tab:color_vs_shape} compares scenarios with these varying shape targets to the old scenarios where shape was constant and color was varying instead.
\mon{1} performs better in this new setting but \mon{2} and \mon{3} perform worse.
It is perhaps easier for the agent to identify goals if they are close to camera height as opposed to identifying goals that extend across the height of the agent.
So this new setting favors the agents in identifying goal objects, but it is more difficult to differentiate between different target objects.

\begin{table}
\ra{1.3}
\centering
\caption{Test set performance of \NoMap baseline when varying color and shapes.}
\label{tab:color_vs_shape}
\resizebox{\linewidth}{!}{
\begin{tabular}{@{}l rrr rrr @{\hspace{7mm}} rrr rrr @{}}
\toprule
 & \multicolumn{3}{c}{\Success (\%)} & \multicolumn{3}{c}{\Progress (\%)} & \multicolumn{3}{c}{\SPL (\%)} & \multicolumn{3}{c}{\PPL (\%)}\\ \cmidrule(lr){2-4} \cmidrule(lr{7mm}){5-7} \cmidrule(l{0mm}r){8-10} \cmidrule(lr){11-13}
 & \mon{1} & \mon{2} & \mon{3} & \mon{1} & \mon{2} & \mon{3} & \mon{1} & \mon{2} & \mon{3} & \mon{1} & \mon{2} & \mon{3}\\
\midrule
\NoMap (color) & $62$ & $24$ & $10$ & $62$ & $39$ & $24$ & $35$ & $13$ & $4$ & $35$ & $21$ & $14$\\
\NoMap (shape) & $73$ & $1$ & $0$ & $73$ & $9$ & $5$ & $55$ & $1$ & $0$ & $55$ & $6$ & $4$\\

\bottomrule

\end{tabular}
}
\end{table}
\begin{table}[t]
\ra{1.3}
\centering
\caption{
Test set performance of \OracleMap on the \mon{3} task while allowing a fixed number of wrong \found actions during the episode. First column (\found budget) lists the number of wrong \found actions allowed while the other columns list the corresponding performance metrics.  
}
\label{tab:found-budget}
\resizebox{0.6\linewidth}{!}{
\begin{tabular}{@{}l c c c c  @{}}
\toprule
\found budget& \Progress (\%) & \Success (\%) & \PPL (\%) & \SPL (\%)\\
\midrule
0  &  $62$ & $48$ & $49$ & $38$   \\
1  &  $76$ & $66$ & $58$ & $50$   \\
2  &  $79$ & $70$ & $58$ & $50$   \\
3  &  $82$ & $74$ & $59$ & $54$   \\
5  &  $85$ & $78$ & $60$ & $55$   \\
10  &  $87$ & $82$ & $60$ & $55$   \\
15  &  $88$ & $82$ & $60$ & $56$   \\
20  &  $90$ & $85$ & $60$ & $57$   \\
50  &  $90$ & $85$ & $60$ & $57$   \\
Oracle Found  &  $93$ & $89$ & $65$ & $62$   \\
\bottomrule
\end{tabular}
}
\end{table}

\xhdr{Effect of having a \textit{hard} \found action.}
An episode of \mon{m} can terminate either by calling a wrong \found action or by reaching the time limit ($2@500$ for all experiments in this paper).
Calling a \found action when not within the threshold distance of the current target object leads to a `wrong' \found action.
To understand the effect of having a \textit{hard} \found action, where a single wrong \found terminates the episode immediately, we allow the agent to call a fixed number of wrong \found actions during the episode.
We found that allowing even a single wrong \found action leads to a significant increase in performance metrics.
This suggests that many episodes terminate due to calling \found action at the wrong time and fixing this inadequacy could improve \mon{m} performance significantly.
\Cref{tab:found-budget} summarizes the results of performance metrics against the number of wrong \found actions allowed in \OracleMap model on the \mon{3} task.
Note that these evaluations were performed on agents trained in the usual way, where a single wrong \found action terminates the episode.
The last row of the table (Oracle Found) quotes the performance metrics when an `oracle' issues a \found action when the agent is within the threshold distance of the current target object.
For this, the action is sampled from the three action probabilities corresponding to \{\textsc{forward}, \textsc{turn-left}, \textsc{turn-right}\} and \found action is called automatically when the agent is near the current target goal.

\end{document}